\documentclass[10pt,journal,compsoc,letterpaper]{IEEEtran}
\fontsize{9.5pt}{11.5pt}

\usepackage{algorithmic}
\usepackage{amsfonts}
\usepackage{amsmath}
\usepackage{amssymb}
\usepackage{caption}
\usepackage[nocompress]{cite}
\usepackage{dblfloatfix}
\usepackage{graphicx}
\usepackage[hidelinks,breaklinks=true]{hyperref}
\usepackage{ifthen}
\usepackage{sansmath}
\usepackage[caption=false,font=footnotesize,labelfont=sf,textfont={sf,sansmath}]{subfig}
\usepackage[flushleft]{threeparttable}
\usepackage{upgreek}
\usepackage{fancyhdr}
\usepackage{eso-pic}
\usepackage{kantlipsum}
\usepackage{url}
\usepackage[bottom]{footmisc}
\DeclareCaptionFont{sansmath}{\sansmath}
\newcommand{\secref}[1]{Section \ref{#1}}
\newcommand{\secreftwo}[2]{Sections \ref{#1} and \ref{#2}}
\newcommand{\eqnref}[1]{Eq. (\ref{#1})}
\newcommand{\eqnreftwo}[2]{Eqs. (\ref{#1}) and (\ref{#2})}

\newcommand{\figref}[1]{Fig. \ref{#1}}
\newcommand{\figreftwo}[2]{Figs. \ref{#1} and \ref{#2}}

\newcommand{\tabref}[1]{Table \ref{#1}}

\newcommand{\R}{\mathbb{R}}

\newcommand{\mli}[1]{\mathit{#1}}
\providecommand{\e}[1]{\ensuremath{\times 10^{#1}}}

\AddToShipoutPicture{%
	\AtPageUpperLeft{%
		\setlength\unitlength{1in}%
		\hspace*{\dimexpr0.5\paperwidth\relax}
		\makebox(0.29,-0.635)[r]{\scriptsize \sf{IEEE TRANSACTIONS ON PATTERN ANALYSIS AND MACHINE INTELLIGENCE, 2019}}%
}}
\begin{document}


	\title{Unsupervised Learning of a Hierarchical Spiking Neural Network for Optical Flow Estimation: \\From Events to Global Motion Perception}
	
	
	\author{
		Federico Paredes-Vall\'es \href{https://orcid.org/0000-0002-9478-7195}{\includegraphics[scale=0.08]{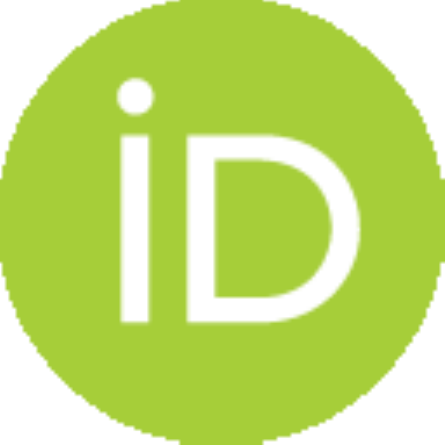}},
		Kirk~Y.~W.~Scheper \href{https://orcid.org/0000-0003-2770-5556}{\includegraphics[scale=0.08]{orcid_128x128.pdf}},
		and~Guido~C.~H.~E.~de~Croon \href{https://orcid.org/0000-0001-8265-1496}{\includegraphics[scale=0.08]{orcid_128x128.pdf}},~\IEEEmembership{Member,~IEEE}
		
		\IEEEcompsocitemizethanks{\IEEEcompsocthanksitem The authors are with the Department of Control and Simulation (Micro Air Vehicle
			Laboratory), Faculty of Aerospace Engineering, Delft University of Technology, Kluyverweg 1, 2629 HS Delft, The Netherlands.\protect\\
			E-mail: $\{$\href{mailto:f.paredesvalles@tudelft.nl}{f.paredesvalles}, \href{mailto:k.y.w.scheper@tudelft.nl}{k.y.w.scheper}, \href{mailto:g.c.h.e.decroon@tudelft.nl}{g.c.h.e.decroon$\}$@tudelft.nl}}%
	}

	
	\IEEEtitleabstractindextext{
		\begin{abstract}
			The combination of spiking neural networks and event-based vision sensors holds the potential of highly efficient and high-bandwidth optical flow estimation. This paper presents the first hierarchical spiking architecture in which motion (direction and speed) selectivity emerges in an unsupervised fashion from the raw stimuli generated with an event-based camera. A novel adaptive neuron model and stable spike-timing-dependent plasticity formulation are at the core of this neural network governing its spike-based processing and learning, respectively. After convergence, the neural architecture exhibits the main properties of biological visual motion systems, namely feature extraction and local and global motion perception. Convolutional layers with input synapses characterized by single and multiple transmission delays are employed for feature and local motion perception, respectively; while global motion selectivity emerges in a final fully-connected layer. The proposed solution is validated using synthetic and real event sequences. Along with this paper, we provide the \textit{cuSNN} library, a framework that enables GPU-accelerated simulations of large-scale spiking neural networks. Source code and samples are available at \url{https://github.com/tudelft/cuSNN}.
		\end{abstract}
		
		}
	
	
	\maketitle
	\IEEEdisplaynontitleabstractindextext
	\IEEEpeerreviewmaketitle
	
	
	\IEEEraisesectionheading{\section{Introduction}\label{sec:2}}
	
	\IEEEPARstart{W}{henever} an animal endowed with a visual system navigates through an environment, turns its gaze, or simply observes a moving object from a resting state, motion patterns are perceivable at the retina level as spatiotemporal variations of brightness \cite{borst2015common}. These patterns of apparent motion, formally referred to as \emph{optical flow} \cite{gibson1950perception}, are a crucial source of information for these animals to estimate their ego-motion and to have a better understanding of the visual scene. A great example of the efficacy of these cues in nature is in flying insects \cite{borst2010fly, borst2015common}, which are believed to heavily rely on these visual cues to perform high-speed maneuvers such as horizontal translation or landing \cite{srinivasan1996honeybee}.
	
	Considering their size and weight limitations, insects are a clear indicator of the efficiency, robustness, and low latency of the optical flow estimation conducted by biological systems. The ability to reliably mimic this procedure would have a significant impact on the field of micro-robotics due to the limited computational capacity of their onboard processors. As an example, Micro Air Vehicles (MAVs), such as the DelFly Explorer \cite{de2014autonomous} or the DelFly Nimble\cite{karasek2018tailless}, could benefit from a bio-inspired visual motion estimation for high-speed autonomous navigation in cluttered environments.
	
	Biological visual systems receive their input from photoreceptors in the retina. These light-sensitive neurons absorb and convert incoming light into electrical signals which serve as input to the so-called ganglion cells. The activity of these neurons consists of temporal sequences of discrete \emph{spikes} (voltage pulses) that are sent to large networks of interconnected cells for motion estimation, among other tasks. Since it is spike-driven, these biological architectures are characterized by a sparse, asynchronous, and massively parallelized computation. Further, they are seen to adapt their topology, i.e. connectivity pattern, in response to visual experience \cite{kirkwood1994hebbian, katz1996synaptic}. This adaptation, or \emph{learning} mechanism, allows them to operate robustly in different environments under a wide range of lighting conditions.
	
	In contrast, the working principle of the majority of cameras used for artificial visual perception is categorized as \emph{frame-based}: data is obtained by measuring the brightness levels of a pixel array at fixed time intervals. Although convenient for some computer vision applications, these sensors are inefficient for the task of motion estimation as the frame rate is independent of the dynamics of the visual scene. Additionally, due to the limited temporal resolution of these sensors, rapidly moving objects may introduce motion blur, limiting the accuracy of optical flow estimation.
	
	However, not all artificial systems rely on conventional frame-based cameras for visual motion estimation. Inspired by biological retinas, several \emph{event-based} vision sensors have recently been presented \cite{lichtsteiner2008128, posch2011qvga, brandli2014240, sees1}. Similar to ganglion cells, each of the elements of the pixel array reacts asynchronously to brightness changes in its corresponding receptive field by generating \emph{events}. A microsecond temporal resolution, latencies in this order of magnitude, a wide dynamic range, and a low power consumption make these sensors an ideal choice for visual perception.
	
	Regardless of the vision sensor, the estimation of optical flow by artificial systems is normally performed algorithmically, with solutions that are built on simplifying assumptions that make this problem tractable \cite{benosman2014event, fortun2015optical}. In spite of this, the recent progress in parallel computing hardware has enabled artificial motion perception to be addressed from a more bio-inspired perspective: Artificial Neural Networks (ANNs). Similar to biological architectures, ANNs consist of large sets of artificial neurons whose interconnections can be optimized for the task at hand. However, despite the high accuracy reported with both frame- \cite{ilg2016flownet} and event-based sensors \cite{zhu2018ev, ye2018unsupervised}, there is still a fundamental difference: the underlying communication protocol in ANNs relies on synchronous packages of floating-point numbers, rather than on trains of asynchronous discrete spikes. As a consequence, these architectures are often computationally expensive.
	
	Taking further inspiration from nature, Spiking Neural Networks (SNNs) have been proposed as a new generation of ANNs \cite{maass1997networks}. As the name suggests, the computation carried out by these bio-realistic neural models is asynchronous and spike-based, which makes them a suitable processing framework for the sparse data generated by event-based sensors \cite{orchard2014bioinspired}. Moreover, SNNs can benefit from an efficient real-time implementation in \emph{neuromorphic hardware}, such as IBM's TrueNorth chip \cite{merolla2014million} or Intel's Loihi processor \cite{davies2018loihi}. Despite these promising characteristics, the spiking nature of these networks limits the application of the successful gradient-based optimization algorithms normally employed in ANNs. Instead, learning in SNNs is dominated by Spike-Timing-Dependent Plasticity (STDP) \cite{caporale2008spike}, a biologically plausible protocol that adapts the strength of a connection between two neurons based on their correlated activity. STDP has been successfully applied to relatively simple image classification tasks \cite{masquelier2007unsupervised, diehl2015unsupervised, kheradpisheh2018stdp, tavanaei2017multi, shrestha2017stable}. However, until now, no study has discussed the use of this learning rule for the estimation of event-based optical flow.
	
	\begin{figure}[t]
		\centering
		\includegraphics[width=0.425\textwidth]{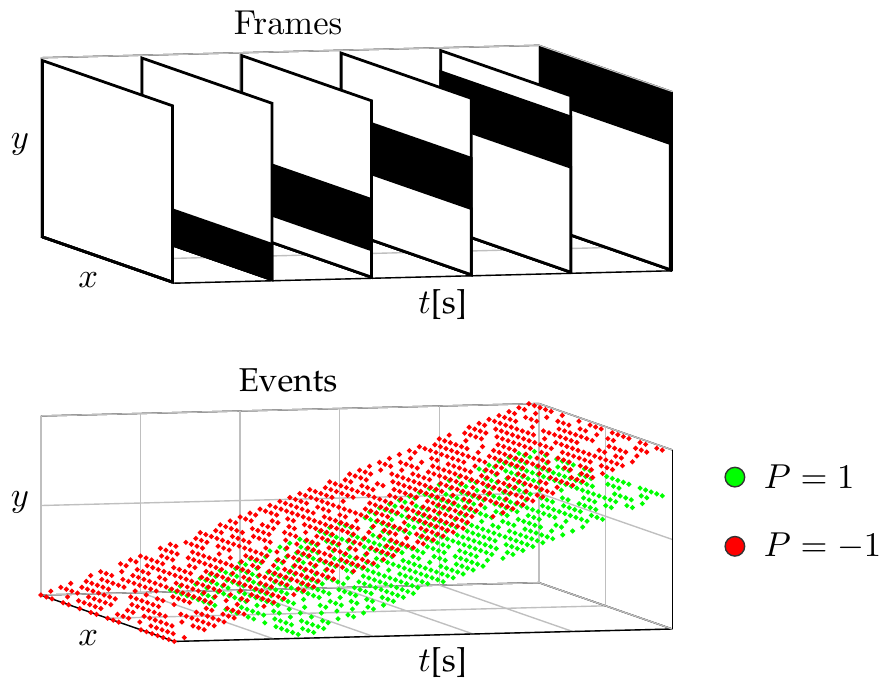}
		\caption{Comparison of the output of frame- and event-based vision sensors under the stimulus of a black horizontal bar moving upward over a homogeneous white background. While frames are basically two-dimensional snapshots of the visual scene, events are spatiotemporal sparse points tracking the leading and trailing edges of the bar.}
		\label{figrel:1}
	\end{figure}
	
	This paper contains \emph{three main contributions}. First, a novel adaptive mechanism for the Leaky Integrate-and-Fire (LIF) spiking neuron model \cite{stein1965theoretical} is introduced. This adaptation extends the applicability of this model to the rapidly varying input statistics of a moving event-based vision sensor. Second, a novel, inherently-stable STDP implementation is proposed. With this learning rule, the strength of neural connections naturally converges to an equilibrium distribution without the need for the ad-hoc mechanisms used by most of the existing formulations. Third, the proposed neuron model and STDP rule are combined in a hierarchical SNN architecture that, after learning, resembles the main functionalities of biological visual systems: feature extraction and local and global motion perception. To the best of the authors' knowledge, this paper shows, for the first time, that neural selectivity to the local and global motion of input stimuli can emerge from visual experience in a biologically plausible unsupervised fashion.
	
	The rest of the paper is structured as follows. \secref{sec:3} provides background information concerning event-based vision, SNNs, and optical flow estimation. The foundations of the spike-based processing and learning of the proposed SNN are detailed in \secreftwo{sec:4}{sec:5}, respectively. Thereafter, the network architecture is described in \secref{sec:6}, and empirically evaluated in \secref{sec:7}.
	
	
	\section{Background Information}\label{sec:3}
	\subsection{Event-Based Vision Sensors}\label{rel:1}
	Inspired by biological retinas, each of the pixels of an event-based vision sensor reacts asynchronously to local changes in brightness by generating discrete temporal events. Specifically, the generation of an event is triggered whenever the logarithmic change of the image intensity $I(x, y, t)$ exceeds a predefined threshold $C$ such that $\bigl|\Delta \text{ log}\big(I(x, y, t)\big)\bigr| > C$ \cite{lichtsteiner2008128}. This variation is computed with respect to a reference brightness level set by the last occurring event at that pixel.
	
	Each event encodes information about its timestamp $t$, its corresponding $(x, y)$ location in the pixel array, and the polarity $\smash{P\in\{-1,1\}}$ of the intensity change. This communication protocol is referred to as Address-Event Representation (AER), and any camera that makes use of it is categorized as Dynamic Vision Sensor (DVS). A visual comparison of the output of frame- and event-based sensors under the same stimulus is illustrated in \figref{figrel:1}.
	
	\subsection{Spiking Neural Networks}\label{rel:3}
	\paragraph*{\textbf{Models of spiking neurons}}
	
	In biological networks, neural communication consists in the exchange of voltage pulses \cite{maass1997networks}. For the reproduction of this asynchronous and spike-based mechanism in SNNs, multiple models of spiking neurons have been presented at various levels of abstraction. Biophysical formulations lead to accurate representations of neural dynamics \cite{hodgkin1952quantitative}, however, their complexity limits their use in large-scale networks. Alternatively, phenomenological formulations offer a compromise between computational load and biological realism. The most used models are the aforementioned LIF \cite{stein1965theoretical}, the Izhikevich \cite{izhikevich2003simple}, and the Spike Response Model \cite{kistler1997reduction}.
	
\begin{figure}[tb]
		\centering
		\includegraphics[width=0.47\textwidth]{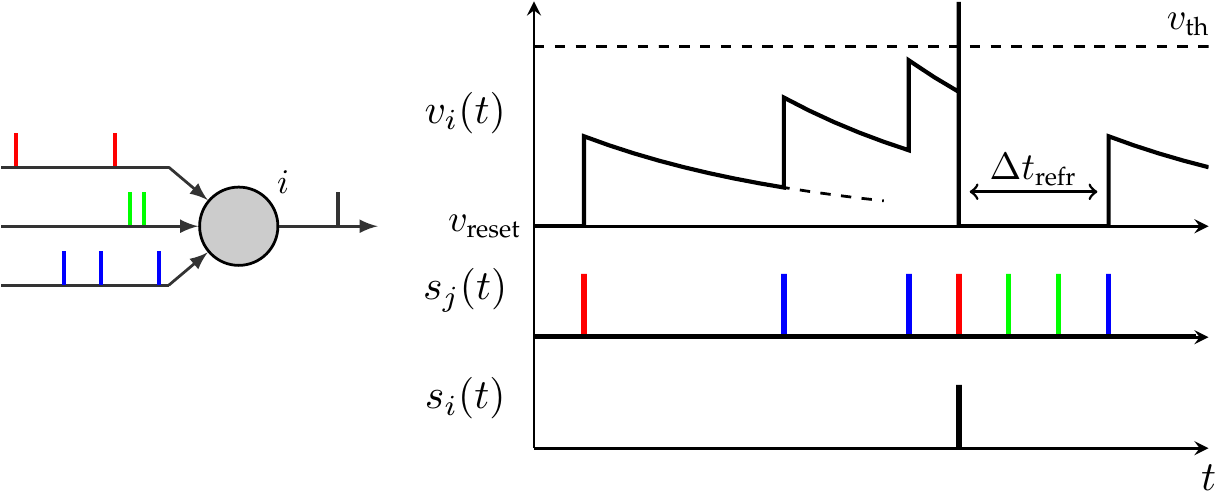}
		\caption{A model of a LIF neuron. The graphic (right) shows the temporal course of the membrane potential $v_{i}(t)$ of the $i$th neuron (left), driven by a sample presynaptic spike train $s_{j}(t)$ from three input synapses. Spikes are depicted as vertical bars at the time at which they are received (if presynaptic) or emitted (if postsynaptic). Here, the reset $v_{\text{reset}}$ and resting $v_{\text{rest}}$ potentials are equal in magnitude.}
		\label{figrel:3}
	\end{figure}

	From a conceptual perspective, the majority of these models share some fundamental principles and definitions. The junction of two neurons is called \emph{synapse}; and relative to these cells, the transmitting neuron is labeled as \emph{presynaptic}, while the receiving as \emph{postsynaptic}. Each spiking neuron, as processing unit, is characterized by an internal state variable, known as \emph{membrane potential} $v_{i}(t)$, which temporally integrates presynaptic spikes over time. If the arrival of a spike leads to an increase (decrease) in $v_{i}(t)$, then the spike is said to have an \emph{excitatory} (\emph{inhibitory}) effect on the neuron. $v_{i}(t)$ decays to a resting potential $v_{\text{rest}}$ in case no input is received. Lastly, a postsynaptic spike is triggered whenever $v_{i}(t)$ crosses the \emph{firing threshold} $v_{\text{th}}$. Afterwards, the neuron resets its membrane potential to $v_{\text{reset}}$, and enters in a \emph{refractory period} $\Delta t_{\text{refr}}$ during which new incoming spikes have negligible effect on $v_{i}(t)$. \figref{figrel:3} illustrates these concepts for the case of a LIF neuron \cite{stein1965theoretical}.
	\paragraph*{\textbf{Synaptic plasticity}}
	
	Defined as the ability to modify the \emph{efficacy} (weight) of neural connections, \emph{synaptic plasticity} is the basic mechanism underlying learning in biological networks \cite{baudry1998synaptic}. These architectures are seen to rely on different learning paradigms depending on their duty \cite{doya1999computations}. For instance, information encoding in biological visual systems is established in an unsupervised fashion, while reinforcement and supervised learning are employed for tasks such as decision making and motor control. Accordingly, various forms of synaptic plasticity have been proposed for SNNs.
	
	In the context of SNNs, unsupervised learning is generally referred to as Hebbian learning, since plasticity rules from this paradigm are based on Hebb's postulate: ``\emph{cells that fire together, wire together}'' \cite{hebb1952organisation}. In essence, these methods adapt the efficacy of a connection based on the correlated activity of pre- and postsynaptic cells. Among others, the biologically plausible STDP protocol is, by far, the most popular Hebbian rule for SNNs \cite{caporale2008spike}. With STDP, the repeated arrival of presynaptic spikes to a neuron shortly before it fires leads to synaptic strengthening, also known as Long-Term Potentiation (LTP); whereas if the arrival occurs shortly after the postsynaptic spike, synapses are weakened through Long-Term Depression (LTD). Therefore, the change of efficacy $\Delta W$ is normally expressed as a function of the relative timing between these two events. STDP formulations exclusively dependent on this parameter are referred to as \emph{additive} rules \cite{gerstner2002spiking}, are inherently unstable, and require the use of constraints for the weights, thus resulting in bimodal distributions \cite{caporale2008spike}. On the other hand, \emph{multiplicative} STDP rules incorporate the current efficacy value in the computation of $\Delta W$. The formulations proposed in \cite{masquelier2007unsupervised, diehl2015unsupervised, kheradpisheh2018stdp} incorporate the weights in a proportional fashion, and represent the current sate-of-the-art in pattern recognition with SNNs. However, they still lead to bimodal distributions. Contrarily, \cite{tavanaei2017multi, shrestha2017stable} claim that, by incorporating the weight dependency in an inversely proportional manner, stable unimodal distributions are obtained. Nevertheless, their stability results from a complex temporal LTP-LTD balance, and it is not theoretically guaranteed.
	
	Several lines of research can be distinguished regarding the use of supervised learning in SNNs, with the most promising based on the well-known error backpropagation algorithm \cite{rumelhart1988learning}. Firstly, numerous adaptations to the discontinuous dynamics of SNNs have recently been proposed for learning temporally precise spike patterns \cite{lee2016training, wu2018spatio, taherkhani2018supervised, shrestha2018slayer}. Alternatively, due to the popularity of this method in ANNs, SNNs commonly rely on transferring optimization results from their non-spiking counterparts \cite{perez2013mapping, zambrano2017efficient, rueckauer2017conversion}. In both cases, high accuracy levels are reported in image classification tasks, but still far below those obtained with conventional ANNs.
	
	With respect to reinforcement learning in SNNs, various models have been presented, the majority of which consist in the modulation of STDP with a reward function \cite{florian2007reinforcement, izhikevich2007solving}. However, applications of this paradigm are mainly focused on neuroscience research \cite{rombouts2012neurally, rombouts2012biologically}, besides several goal-directed navigation problems \cite{friedrich2016goal, bing2018end} and the digit-recognition application from \cite{mozafari2018combining}.
	
	\subsection{Event-based Optical Flow Estimation}\label{rel:2}
	
	The recent introduction of the DVS and other retinomorphic vision sensors has lead to the development of several novel approaches to event-based optical flow estimation. Depending on their working principle, these solutions are divided into algorithmic and neural methods.
	
	Gradient-, plane-fitting-, frequency-, and correlation-based approaches set the basis of the algorithmic state-of-the-art. These techniques compute sparse optical flow estimates for each newly detected event (or group of events) based on its spatiotemporal polarity-specific neighborhood. Firstly, adaptations of the gradient-based Lucas-Kanade algorithm \cite{Lucas1981a} were presented in \cite{benosman2012asynchronous, brosch2015event}. Secondly, the methods proposed in \cite{benosman2014event, aung2018event, hordijk2017vertical} extract optical flow by computing the gradients of a local plane fitted to a spatiotemporal surface of events. Thirdly, multiple adaptations of the bio-inspired frequency-based methods have been introduced \cite{tschechne2014bio, barranco2015bio, brosch2015event}, which allow the implementation in neuromorphic hardware \cite{brosch2016event}. Lastly, the recent correlation-based approaches presented in \cite{zhu2017event, gallego2017accurate, mitrokhin2018event, gallego2018unifying} employ convex optimization algorithms to associate groups of events over time, and report the highest algorithmic accuracy to date. Part of this category is also the block-matching method recently proposed in \cite{liu2018abmof}, which employs conventional search techniques to find the best matching group of events in previous temporal slices of the input.
	
	The estimation of event-based optical flow with neural models is dominated by SNNs. However, there are a couple of ANN-based approaches worth remarking. In \cite{zhu2018ev}, a self-supervised learning scheme was employed to train a convolutional ANN to estimate dense image flow. The input to the network consists of the per-pixel last timestamp and count of events over a specific time window. Using the average timestamp instead, in \cite{ye2018unsupervised}, the authors presented the first neural model to approach the full structure-from-motion problem using event-based input. In \cite{ye2018unsupervised}, two ANNs are employed for depth and dense optical flow estimation. Regarding the latter task, accuracy levels considerably higher than those from \cite{zhu2018ev} are reported.
	
	Though the main goal of \cite{lagorce2015spatiotemporal} is for predicting future input activations, this work presented the first neural architecture capable of learning spatiotemporal features from raw event data. For this purpose, multiple recurrent ANNs were employed in combination with a single layer of spiking neurons. A self-supervised learning scheme, based on a recursive least-squares algorithm, was proposed for training the ANNs to capture the spatiotemporal information. Note that, for compatibility reasons, this approach requires the conversion of the event data into analog signals.
	
	With respect to pure SNN-based approaches, in \cite{giulioni2016event, haessig2018spiking}, the authors propose an architecture in which motion selectivity results from the synaptic connections of a bursting neuron to two neighboring photoreceptors, one excitatory and the other inhibitory. If the edge is detected first by the excitatory cell, spikes are emitted at a fixed rate until the inhibitory pulse is received. Otherwise, the neuron remains inactive. Optical flow is consequently encoded in the burst length and in the relative orientation of the photoreceptors.
	
	In contrast, the SNNs presented in \cite{richter2014bio, orchard2013spiking} extract motion information through synaptic delays and spiking neurons acting as coincidence detectors. A simple spike-based adaptation of the Reichardt model \cite{reichardt1961autocorrelation} is introduced in \cite{richter2014bio} to show the potential of this approach. This idea is explored in more detail in \cite{orchard2013spiking}, in which the authors propose the convolution of event sequences with a bank of spatiotemporally-oriented filters, each of which is comprised of non-plastic synapses with equal efficacies, but with delays tuned to capture a particular direction and speed. Similarly to frequency-based methods \cite{adelson1985spatiotemporal}, these filters compute a confidence measure, encoded in the neural activity, rather than the optical flow components. Additionally, this solution employs a second spike-based pooling layer for mitigating the effect of the aperture problem \cite{ullman1979interpretation}.
	
	Whether, and how, direction and speed selectivity emerge in biological networks from visual experience still remains an open question. Some initial work by \cite{shon2004motion, wenisch2005spontaneously, adams2015computational} shows that robust local direction selectivity arises in neural maps through STDP if, apart from presynaptic feedforward connections, neurons receive spikes from cells in their spatial neighborhood through plastic synapses with distance-dependent transmission delays. However, no study has assessed the speed selectivity of these cells, which is crucial for optical flow estimation.
	
	
	\vspace{-1.5pt}
	\section{Adaptive Spiking Neuron Model}\label{sec:4}\label{found:1}
	
\begin{figure}[t]
		\centering
		\includegraphics[width=0.45\textwidth]{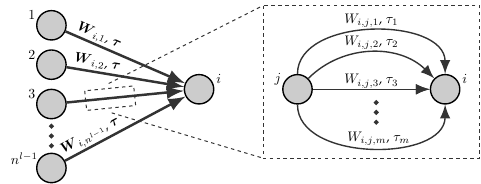}
		\caption{Schematic of the feedforward connectivity between neurons from two adjacent layers (left). These connections can be considered as being multisynaptic (right), each one having its own efficacy, transmission delay, and trace.}
		\label{figlif:1}
	\end{figure}

	Let $\smash{j = 1, 2, \ldots, n^{l-1}}$ denote a group of presynaptic neurons, from layer $\smash{l-1}$, fully connected in a feedforward fashion to a set of postsynaptic cells $\smash{i = 1, 2, \ldots, n^{l}}$, from layer $l$. As depicted in \figref{figlif:1}, these neural connections can be considered as \emph{multisynaptic}, i.e. the link between two cells is not restricted to a single synapse, but several can coexist. In this work, the number of multisynaptic connections $m$ is layer-specific, and each synapse has its own transmission delay as given by $\smash{\boldsymbol{\tau}\in\R^{m}}$. In addition to this delay vector, layer connectivity is also characterized by a weight matrix $\smash{\boldsymbol{W}\in \R^{n^{l}\times n^{l-1}\times m}}$, which determines the synaptic efficacy of the connections.
	
	Apart from $\smash{\boldsymbol{W}}$ and $\smash{\boldsymbol{\tau}}$, each synapse keeps track of an additional parameter that captures the recent history of spikes transmitted. Referred to as the \emph{presynaptic trace} \cite{morrison2007spike}, and defined as $\smash{\boldsymbol{X}\in\R^{n^{l}\times n^{l-1}\times m}}$, its dynamics is given by:
	\begin{equation}\label{eqlif:1}
	\begin{aligned}
	\lambda_{X}\frac{dX_{i,j,d}(t)}{dt} = -X_{i,j,d}(t) + \alpha s_{j}^{l-1}(t-\tau_{d})
	\end{aligned}
	\end{equation}
	\noindent{where $\smash{\smash{\lambda_{X}}}$ is the time constant of the system, $\smash{\alpha}$ is a scaling factor, and $\smash{\boldsymbol{s}^{l}(t)\in\R^{n^{l}}}$ denotes the (binary) record of neural activity, or \emph{spike train}, of cells from layer $l$. Note that $\smash{d = 1, 2, \ldots, m}$ serves to refer both to connections within a multisynaptic group and their corresponding delays.}
	
	From \eqnref{eqlif:1}, whenever a spike arrives at a postsynaptic neuron $i$ via a synapse with transmission delay $\smash{\tau_{d}}$, the corresponding presynaptic trace $\smash{X_{i,j,d}(t)}$ increases by a factor of $\smash{\alpha}$. In case no spike is received, the trace decays exponentially towards zero according to $\smash{\lambda_{X}}$.
	
	\begin{figure*}[t]
		\centering
		\includegraphics[width=\textwidth]{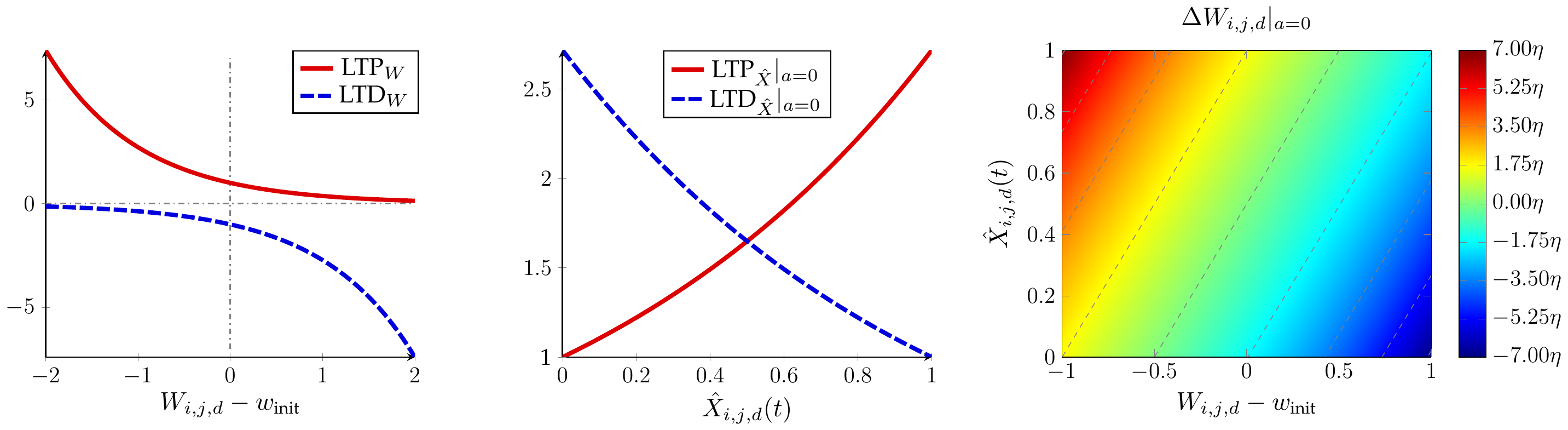}
		\caption{Illustration of the novel multiplicative STDP rule proposed in this work. The weight update (right) results from the linear combination of the non-exclusive LTP and LTD processes. These, in turn, are characterized by symmetrical dependencies on the synaptic weights (left) and normalized presynaptic traces (center).  Note that, in the schematic of the weight update (right), the weight axis is limited to the $[-1, 1]$ range only for the purpose of a better visualization of the equilibrium weights for $a = 0$.}
		\label{figstdp:1}
	\end{figure*}
	
	The LIF model \cite{stein1965theoretical} is the most widely used spiking neuron model in literature. This is due to its assumption that in SNNs, information is not encoded in the spike amplitude, but rather in the firing time. Consequently, neural activity is reduced to discrete and binary temporal events, thus ensuring computational tractability. The neuron model used in this paper is a modified LIF formulation, defined as:
	\begin{align}
	\lambda_{v}\frac{dv_{i}(t)}{dt} &= -\big(v_{i}(t) - v_{\text{rest}}\big) + i_{i}(t)\label{eqlif:2}\\ 
	i_{i}(t) &= \sum_{j=1}^{n^{l-1}} \sum_{d=1}^{m} \big(W_{i,j,d} s_{j}^{l-1}(t-\tau_{d}) - X_{i,j,d}(t)\big)\label{eqlif:3}
	\end{align}
	\noindent{where $\smash{\lambda_{v}}$ denotes the time constant of the membrane potential, and $\smash{\boldsymbol{i}(t)}$ is the so-called \emph{forcing function} of the system.}
	
	From \eqnreftwo{eqlif:2}{eqlif:3}, the membrane potential $v_{i}(t)$ of a neuron evolves over time by integrating scaled presynaptic spikes from its input synapses, similarly to the conventional LIF model \cite{stein1965theoretical}. Whenever $v_{i}(t)$ reaches (or surpasses) the firing threshold $v_{\text{th}}$, a postsynaptic spike is generated, i.e. $\smash{s_{i}^{l}(t) = 1}$, and $v_{i}(t)$ is reset to $v_{\text{reset}}$. In addition, the neuron enters in a refractory period $\Delta t_{\text{refr}}$ during which presynaptic spikes have no effect on $v_{i}(t)$ to ensure the temporal separation of postsynaptic pulses. In case no spike is fired at time $t$, this is reflected in the neuron's spike train as $\smash{s_{i}^{l}(t) = 0}$.
	
	Unlike traditional LIF \cite{stein1965theoretical}, the forcing function $\boldsymbol{i}(t)$ of our neuron model includes an additional term, further referred to as the \emph{homeostasis} parameter, which is inspired by the internal regulatory mechanisms of biological organisms \cite{abercrombie2017dictionary}. This is used to adapt the neural response to the varying input statistics---in particular, to the per-pixel firing rate---using the presynaptic trace $\boldsymbol{X}$ as an \emph{excitability} indicator. Inferring from \eqnref{eqlif:3}, this parameter acts, in essence, as an inhibitory penalty in the update rule of $\boldsymbol{v}(t)$. A postsynaptic neuron connected to a group of highly-active presynaptic cells is said to have low excitability due to its relatively high $\boldsymbol{X}$. For this neuron to fire, it needs to receive a large number of presynaptic spikes shortly separated in time. Conversely, the same cell connected to poorly-active neurons is highly excitable; and thus, the firing threshold $v_{\text{th}}$ can still be reached despite the considerably larger time difference between input spikes. Note that, to get the desired neural adaptation, the scaling factor $\alpha$, from \eqnref{eqlif:1}, needs to be selected in accordance with the neural parameters, mainly $v_{\text{th}}$ and the range of possible $\boldsymbol{W}$ values. The effect of this parameter on the neural response is explored in \cite{neuroFede}. 
	When dealing with an event-based camera as source of input spikes, the firing rate of the sensor is not only correlated to the appearance of features from the visual scene, but also to their optical flow and the sensitivity settings of the camera. Slow apparent motion leads to successive events being more distant in time than those captured from fast motion. Consequently, if these events are to be processed with a network of spiking neurons, a homeostasis mechanism is required to ensure that similar features are detected regardless of the encoding spike rate.
	
	Other approaches to homeostasis have been presented, such as threshold balancing \cite{bohte2012efficient} or weight scaling \cite{van2000stable}. However, these methods use postsynaptic spikes to adjust the homeostatic inhibition through an adaptive mechanism. With such neural feedback, there is a delay in adjusting the excitability of the neurons. These approaches are therefore less suitable for the rapidly varying statistics of the data generated by a moving event-based vision sensor.
	
	
	\vspace{-2pt}
	\section{Stable STDP Learning Rule}\label{sec:5}\label{found:2}
	
	In this work, we propose a novel multiplicative STDP implementation that, contrary to the state-of-the-art of this learning protocol, is inherently stable by combining the weight-dependent exponential rule from \cite{shrestha2017stable} with presynaptic trace information. Hereafter, we will simply refer to it as STDP.
	
	Whenever a neuron $i$ fires a spike, the efficacy of its presynaptic connections is updated as follows:
	
	\begin{equation}\label{eqstdp:1}
	\Delta W_{i,j,d} = \eta(\text{LTP}+\text{LTD})
	\end{equation}
	\begin{equation}\label{eqstdp:2}
	\begin{split}
	\text{LTP} &= \text{LTP}_{W} \cdot \text{LTP}_{\hat{X}},\\
	\text{LTP}_{W} &= e^{-(W_{i,j,d}-w_{\text{init}})},\\
	\text{LTP}_{\hat{X}} &= e^{\hat{X}_{i,j,d}(t)}-a,
	\end{split}
	\quad
	\begin{split}
	\text{LTD} &= \text{LTD}_{W} \cdot \text{LTD}_{\hat{X}}\\
	\text{LTD}_{W} &= -e^{(W_{i,j,d}-w_{\text{init}})}\\
	\text{LTD}_{\hat{X}} &= e^{(1-\hat{X}_{i,j,d}(t))}-a
	\end{split}
	\end{equation}
	
	\noindent{where $\eta$ is the learning rate of the rule, $w_{\text{init}}$ refers to the initialization weight of all synapses at the beginning of the learning process, and $\hat{\boldsymbol{X}}_{i}\in[0,1]$ denotes the presynaptic traces of neuron $i$ normalized to the current maximum at the moment of firing. Further, for stability, $\eta > 0$ and $a < 1$ regardless of the value of $w_{\text{init}}$ (see Appendix A).}
	
	From \eqnreftwo{eqstdp:1}{eqstdp:2}, the weight update $\Delta \boldsymbol{W}_{i}$ results from the linear combination of the output of two non-mutually exclusive processes: LTP, for strengthening, and LTD, for weakening synaptic connections. Both of these processes are dependent on the weights ($\text{LTP}_{W}$, $\text{LTD}_{W}$) and normalized traces ($\text{LTP}_{\hat{X}}$, $\text{LTD}_{\hat{X}}$) of the synapses under analysis. On the one hand, the weight dependency of our learning rule takes inspiration from the STDP formulation presented in \cite{shrestha2017stable}. $\text{LTP}_{W}$ and $\text{LTD}_{W}$ are inversely proportional to $\boldsymbol{W}_{i}$ in an exponential fashion, and are centered around $w_{\text{init}}$ (see \figref{figstdp:1}, left). Consequently, the effect of $\text{LTP}_{W}$ decreases (increases) the larger (smaller) a synaptic weight is in comparison to $w_{\text{init}}$. The opposite relation holds true for $\text{LTD}_{W}$. On the other hand, rather than relying on the precise spike timing \cite{shrestha2017stable}, our rule employs normalized presynaptic trace information as a measure of the relevance of a particular connection to the postsynaptic spike triggering the update. The higher (lower) the value of $\hat{X}_{i,j,d}(t)$, the larger (smaller) the effect of $\text{LTP}_{\hat{X}}$, and vice versa for $\text{LTD}_{\hat{X}}$ (see \figref{figstdp:1}, center).
	
	With this formulation, a weight is established for each value of $\hat{X}_{i,j,d}(t)$ through a stable equilibrium of LTP-LTD contributions on $\Delta \boldsymbol{W}_{i}$ (see \figref{figstdp:1}, right). The parameter $a$ has control over this non-linear mapping through the steepness of $\text{LTP}_{\hat{X}}$ and $\text{LTD}_{\hat{X}}$ in $\hat{\boldsymbol{X}}_{i}\in[0,1]$. 
	The higher (lower) the value of $a$---below the stability limit---, the wider (narrower) the distribution of synaptic weights after convergence. 
	As such, no additional mechanism is required for preventing weights from vanishing or exploding. Synapses characterized by weights that are higher (lower) than their corresponding equilibrium state are consistently depressed (potentiated) until synapse-specific stability is achieved. 
	
	To track the convergence of the learning process, we propose the use of the following mean square error criterion, where $\hat{\boldsymbol{W}}_{i}\in[0,1]$ denotes the presynaptic weights of neuron $i$ after an update, normalized to the current maximum:
	\begin{equation}\label{eqstdp:3}
	\begin{aligned}
	\mathcal{L}_{i} = \frac{1}{n^{l-1}m}\sum_{j=1}^{n^{l-1}}\sum_{d=1}^{m}\big(\hat{X}_{i,j,d}(t) - \hat{W}_{i,j,d}\big)^2
	\end{aligned}
	\end{equation}
	
	As the learning progresses, the moving average of $\mathcal{L}_{i}$ converges to a (close-to-zero) equilibrium state. In this work, we stop synaptic plasticity using a fixed threshold on this parameter, denoted by $\mathcal{L}_{\text{th}}$.
	
	\subsection{Local inter-lateral competition}\label{found:2.1}
	
	For neurons to learn distinct features from the input data through STDP, this learning rule needs to be combined with what is known as a Winner-Take-All (WTA) mechanism \cite{thorpe1990spike}. This form of competition implies that, when a neuron fires a spike and updates its presynaptic weights according to \eqnreftwo{eqstdp:1}{eqstdp:2}, the rest of postsynaptic cells (from the same layer) locally connected to the same input neurons get inhibited. As a result, these cells are prevented from triggering STDP while the neuron that fired first, i.e. the \emph{winner}, remains in the refractory period.
	
	Instead of relying on non-plastic synapses transmitting inhibitory spikes with a certain delay, our implementation assumes that the internal dynamics of these neurons are intercorrelated. Whenever the winner resets its membrane potential and enters in the refractory period, neurons affected by the WTA mechanism do the same immediately afterwards. In case multiple neurons fire simultaneously, the cell with the highest membrane potential has preference for triggering the weight update. Further, the postsynaptic spikes from the other firing neurons are not considered. To ensure coherence between the training and inference phases of our proposed SNN, layers trained with STDP maintain the WTA mechanism after the learning process.
	
	
	\section{Spiking Neural Network Architecture\\ for Motion Perception}\label{sec:6}
\begin{figure}[tb]
		\centering
		\includegraphics[width=0.475\textwidth]{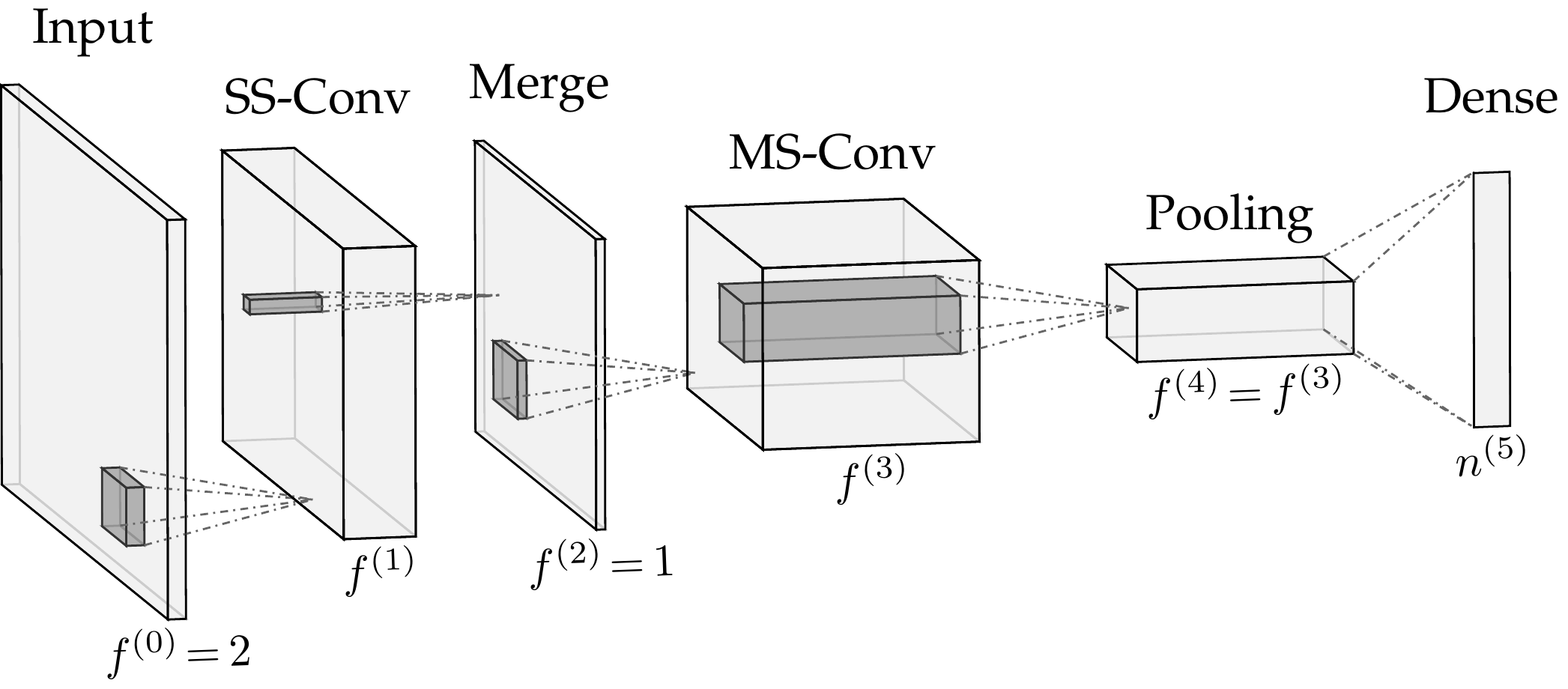}
		\caption{Overview of the feedforward SNN architecture.}
		\label{figarch:1}
	\end{figure}

	To extract a robust measure of motion from the raw camera input, we propose the multi-layer SNN illustrated in \figref{figarch:1}. This section highlights the unique goal of each of the layers comprising this architecture, together with the variations of the proposed neuron model and learning rule that are required depending on their connectivity scheme.
	\subsection{Input Layer}\label{arch:1}
	Being the first stage of the network, the Input layer encodes the event-based sensor data in a compatible format for the rest of the architecture. This layer can be understood as to be comprised of spiking neurons with no internal dynamics, whose neural activity is determined by event arrival. Neurons are arranged in two-dimensional \emph{neural maps}, one per polarity, resembling the grid-like topology of the vision sensor. Depending on the spatial resolution of these maps, each neuron is assigned with the polarity-specific events of one or multiple pixels with no overlap.
	\subsection{SS-Conv Layer: Feature Extraction}\label{arch:2}
	
	The goal of the single-synaptic convolutional layer, or SS-Conv, is to extract visual features from the input, and by doing so, to filter out the input events that may otherwise corrupt the learning process, and hence the performance, of subsequent layers in the architecture.
	
	Neurons in this layer are retinotopically arranged in $\smash{k = 1,2,\ldots,f^{\raisebox{1.5pt}{$\scriptscriptstyle (1)$}}}$ two-dimensional maps. Each of these neurons receives spikes from presynaptic cells within a specific spatial receptive field, of size $r$, in all maps of the previous layer. This sparse connectivity is characterized by a set of excitatory synaptic weights, formally referred to as a \emph{convolutional kernel} $\smash{\boldsymbol{W}_{k}\in\R^{r\times f^{\raisebox{-0.5pt}{$\scriptscriptstyle (0)$}}}}$, that is equal for all neurons belonging to the same map. This form of \emph{weight sharing} ensures that, within a map, neurons are selective to the same feature but at different spatial locations.
	
	Let the input connectivity of neuron $i$ from the map $k$ be characterized by the aforementioned convolutional kernel $\smash{\boldsymbol{W}_{k}}$, the presynaptic trace $\smash{\boldsymbol{X}_{i}\in\R^{r\times f^{(0)}}}$, and the spike train $\smash{s_{i,k}^{\raisebox{1.5pt}{$\scriptscriptstyle (0)$}}(t)}$. Further, let $\boldsymbol{N}_{i,k}$ refer to the map-specific direct neural neighborhood of the cell, including itself. Then, considering neural connections as single-synaptic with transmission delay $\tau$, the forcing function driving the internal dynamics of neurons in this layer is defined as follows:
	\begin{equation}\label{eqconv2d:1}
	\begin{aligned}
	i_{i,k}(t) = &\sum_{j=1}^{r}\sum_{\mli{ch}=1}^{f^{(0)}} W_{j,\mli{ch},k} s_{j,\mli{ch}}^{\raisebox{1.5pt}{$\scriptscriptstyle (0)$}}(t-\tau) \\
	&- \max_{\forall b \in \boldsymbol{N}_{i,k}} \sum_{j=1}^{r} \sum_{\mli{ch}=1}^{f^{(0)}} X_{b,j,\mli{ch}}(t)
	\end{aligned}
	\end{equation}
	
	Apart from the sparse connectivity, the only difference between this expression and the fully-connected formulation, i.e. \eqnref{eqlif:3}, is in the homeostasis parameter. When arranged retinotopically, the neurons' dynamics do not only depend on their own presynaptic trace $\boldsymbol{X}_{i}$, but also on the synaptic traces characterizing their direct spatial neural neighborhood $\boldsymbol{N}_{i,k}$. By using the maximum trace, neurons are prevented from specializing to the leading edge of moving visual features, rather than to the features themselves (see Appendix D.1).
	
	An augmentation of the proposed STDP rule is also required to handle the fact that multiple updates can be generated simultaneously in different spatial locations of the same map. Since these neurons share convolutional kernel, $\Delta \boldsymbol{W}_{k}$ is computed through synapse-specific averages of the local contributions. Additionally, due to the high overlap of presynaptic receptive fields, the WTA inhibitory mechanism described in \secref{found:2.1} is expanded to cells within a small neighborhood of the firing neurons, regardless of the neural map. Note that, after learning, only the neuron-specific competition is maintained.
	\subsection{Merge Layer: Feature Aggregation}\label{arch:2.5}
	Due to the aperture problem \cite{ullman1979interpretation}, the different types of local motion that can be perceived at this stage of the architecture are exclusively dependent on the spatial configuration of input features, i.e. their appearance, and not on their polarity. Consequently, the $\smash{f^{(1)}}$ neural maps of the SS-Conv layer can be merged into a single combined map without losing useful information for motion perception. The Merge layer is used for this purpose. Compared to when local motion is to be perceived directly from the SS-Conv output, this operation results in a decrease of both the number of convolutional kernels required in the subsequent layer, and the amount of per-kernel trainable parameters.
	
	Similarly to SS-Conv, the Merge layer is convolutional and single-synaptic. The internal dynamics of its neurons is driven by \eqnref{eqconv2d:1} (with $l=2$ in this case), but without the need for $\boldsymbol{N}_{i,k}$ since presynaptic connections are not plastic. Because of the latter, the application of the WTA mechanism is also neglected. Instead, this layer is characterized by a single $1\times 1$ convolutional kernel with unitary connections to each of the neural maps of the previous layer.
	\subsection{MS-Conv Layer: Local Motion Perception}\label{arch:3}
	MS-Conv is presented as a variation of the SS-Conv layer whose role is to provide local motion estimates of the features extracted in the previous layers, by means of velocity-selective neurons. Similarly to feature identification, this selectivity emerges from visual experience through STDP.
	
	For the purpose of local motion perception, we propose an augmentation of \eqnref{eqconv2d:1} based on the foundations of frequency-based optical flow methods \cite{adelson1985spatiotemporal} and bio-inspired motion detectors \cite{reichardt1961autocorrelation, barlow1965mechanism}. Firstly, motion is to be extracted as orientation in the spatiotemporal domain. Therefore, neural connections in the MS-Conv layer are considered multisynaptic with different constant transmission delays as given by $\boldsymbol{\tau}\in\R^{m}$. Secondly, since these delays (and the rest of neural parameters) are equal for all (spatiotemporal) convolutional kernels, inhibitory synapses are required to prevent the firing of erroneous postsynaptic spikes when the input trace only fits part of the excitatory component of the kernels. To account for this, each MS-Conv kernel is defined by a pair of excitatory and inhibitory plastic weight matrices, denoted by $\boldsymbol{W}_{k}^{\text{exc}}\in\R^{r\times m}$ and $\boldsymbol{W}_{k}^{\text{inh}}\in\R^{r\times m}$, respectively. According to these additions, the forcing function of cells in this layer is expressed as:
	\begin{equation}\label{eqmsconv:1}
	\begin{aligned}
	i_{i,k}(t) = &\sum_{j=1}^{r} \sum_{d=1}^{m} (W_{j,d,k}^{\text{exc}} + \beta W_{j,d,k}^{\text{inh}}) s_{j}^{\raisebox{1.5pt}{$\scriptscriptstyle (2)$}}(t-\tau_{d}) \\
	&- \max_{\forall b \in \boldsymbol{N}_{i,k}}\sum_{j=1}^{r} \sum_{d=1}^{m} X_{b,j,d}(t)
	\end{aligned}
	\end{equation}
	\noindent{where $\smash{\beta\in[0,1]}$ scales the impact of inhibitory synapses, and the presynaptic trace is defined as $\smash{\boldsymbol{X}_{i}\in\R^{r\times m}}$.}
	
	Due to the neural spatial disposition, the implementation of STDP in this layer is, in essence, identical to the one employed for SS-Conv. The only difference comes from the fact that, for inhibitory synapses, the weights are initialized at $0$, and $\smash{w_{\text{init}}^{\text{inh}}}$ is set to $\smash{-w_{\text{init}}^{\text{exc}}}$. This discrepancy between $w_{\text{init}}^{\text{inh}}$ and the initialization weight enables neurons in this layer to be reactive to different input features until specialization.
	\subsection{Pooling Layer: From Local to Global}\label{arch:4}
	As an intermediate stage between the MS-Conv and Dense layers, the Pooling layer is employed in the SNN architecture as a means to reduce the spatial dimensionality of the former, and hence to facilitate the learning process of the latter. The intuition of this layer is that, by pooling local motion estimates over large portions of the visual scene, a more accurate measure of the global motion in each of these regions can be obtained, thus mitigating the effect of the aperture problem \cite{ullman1979interpretation}.
	
	Similarly to the Merge layer, the Pooling layer is convolutional and single-synaptic, and its presynaptic connections are not plastic. This layer is characterized by the same number of neural maps as the MS-Conv, each one assigned with an excitatory kernel $\boldsymbol{W}_{k}$ that has unitary weights with its presynaptic counterpart and null with the rest. In addition, there is no overlap between receptive fields.
	\subsection{Dense Layer: Global Motion Perception}\label{arch:5}
	The Dense layer, as the final stage of the SNN architecture, is comprised of individual neurons fully connected to cells in the Pooling layer via single-synaptic plastic connections. Similarly to final regions of biological visual motion systems \cite{borst2010fly, borst2015common}, neurons in this layer develop selectivity to the global motion of the scene from visual experience through STDP.
	
	With respect to implementation details, synaptic plasticity is conducted as described in \secref{found:2}, and the forcing function of Dense neurons resembles \eqnref{eqlif:3}, but referring to the convolutional presynaptic layer to which these cells are connected. This expression is then defined as:
	\begin{equation}\label{eqdense:1}
	\begin{aligned}
	i_{i}(t) = \sum_{j=1}^{n^{(4)}} \sum_{\mli{ch}=1}^{f^{(4)}} \big(W_{i,j,\mli{ch}} s_{j,\mli{ch}}^{\raisebox{1.5pt}{$\scriptscriptstyle (4)$}}(t-\tau) - X_{i,j,\mli{ch}}(t)\big)
	\end{aligned}
	\end{equation}
	\noindent{where the weights and trace of input connections are defined as $\smash{\boldsymbol{W}_{i}\in\R^{n^{(4)}\times f^{(4)}}}$ and $\smash{\boldsymbol{X}_{i}\in\R^{n^{(4)}\times f^{(4)}}}$, respectively.}
	
	
	\setcounter{figure}{6}
	\begin{figure*}[!b]
		\centering
		\subfloat[$x$-\hspace{0.5pt}$\tau$ representation\label{figarch:7c}]{%
			\includegraphics[width=0.465\textwidth]{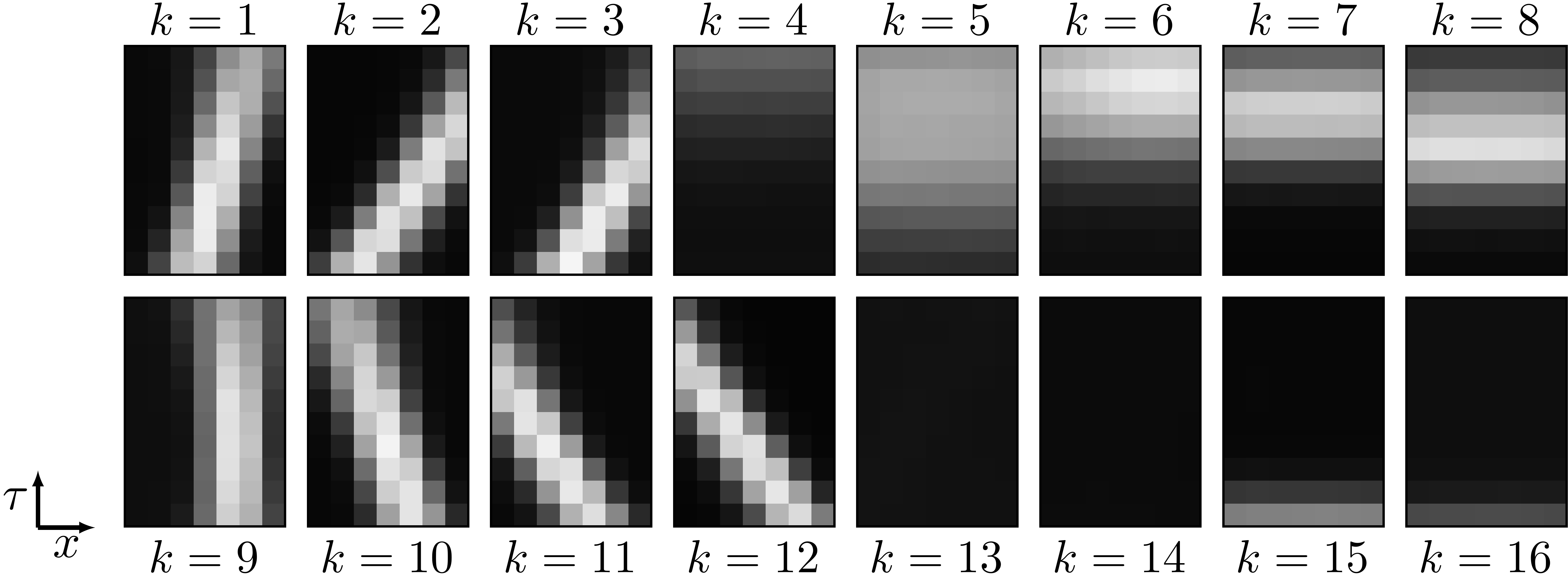}
		}
		\hspace{15pt}
		\subfloat[$y$-\hspace{0.5pt}$\tau$ representation\label{figarch:7d}]{%
			\includegraphics[width=0.465\textwidth]{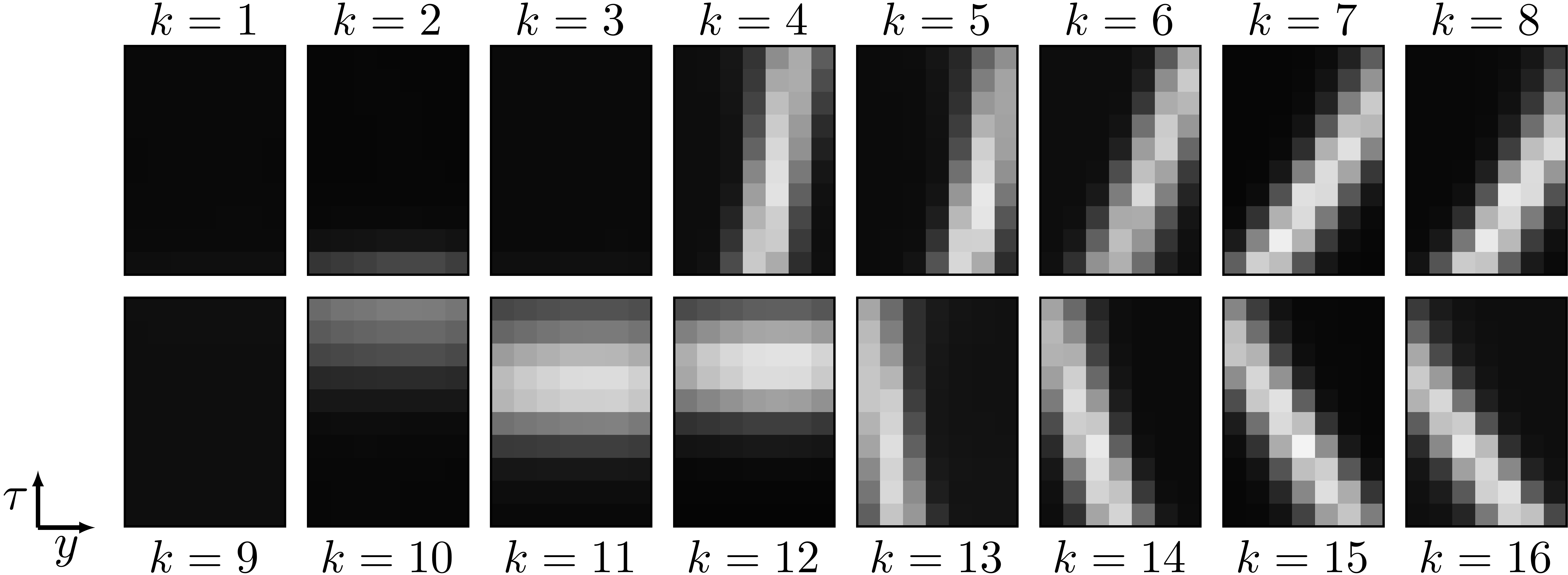}
		}
		\\\vspace{0pt}
		\subfloat[Pure horizontal motion\label{figarch:7a}]{%
			\includegraphics[width=0.485\textwidth]{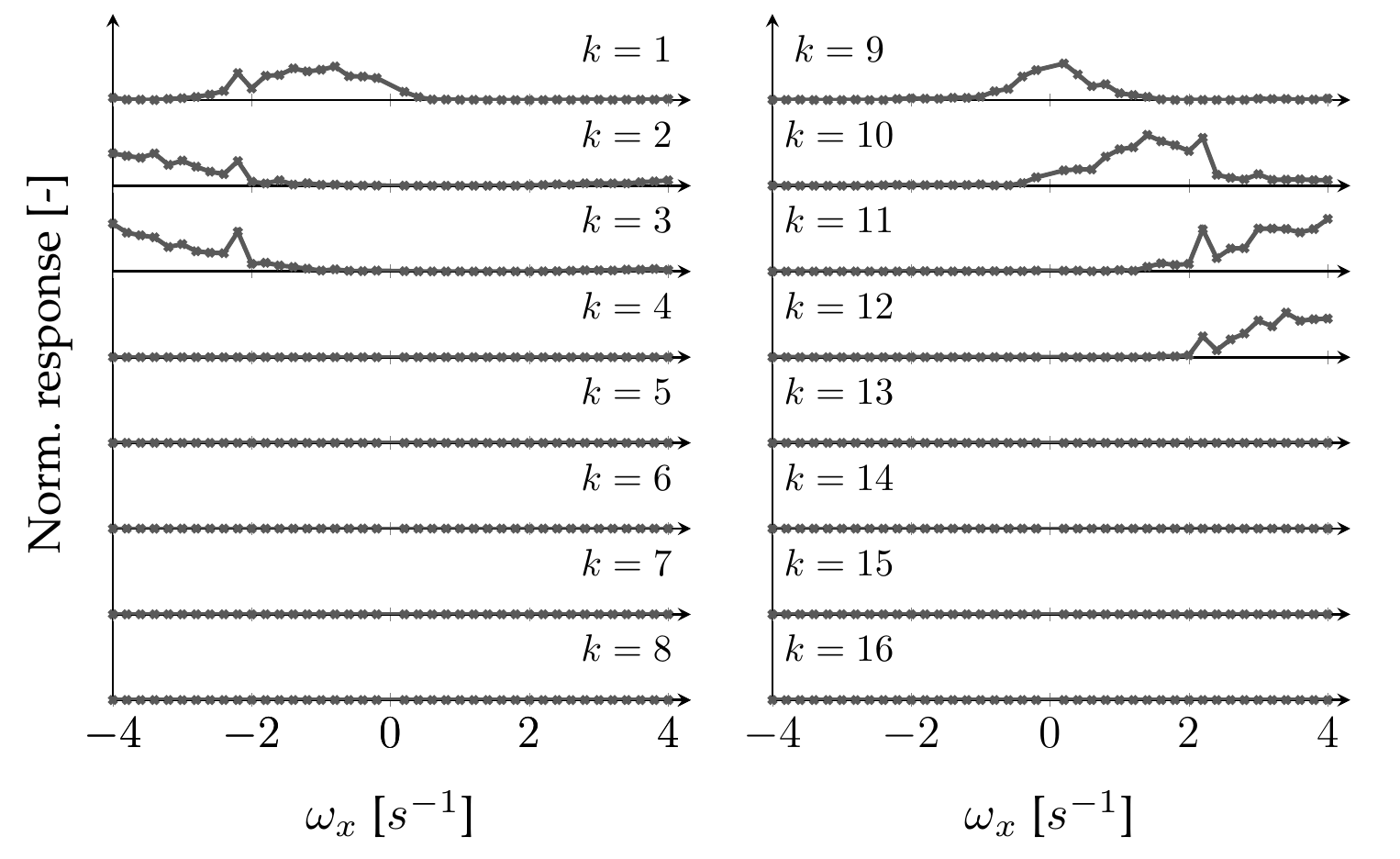}
		}
		\hspace{5pt}
		\subfloat[Pure vertical motion\label{figarch:7b}]{%
			\includegraphics[width=0.485\textwidth]{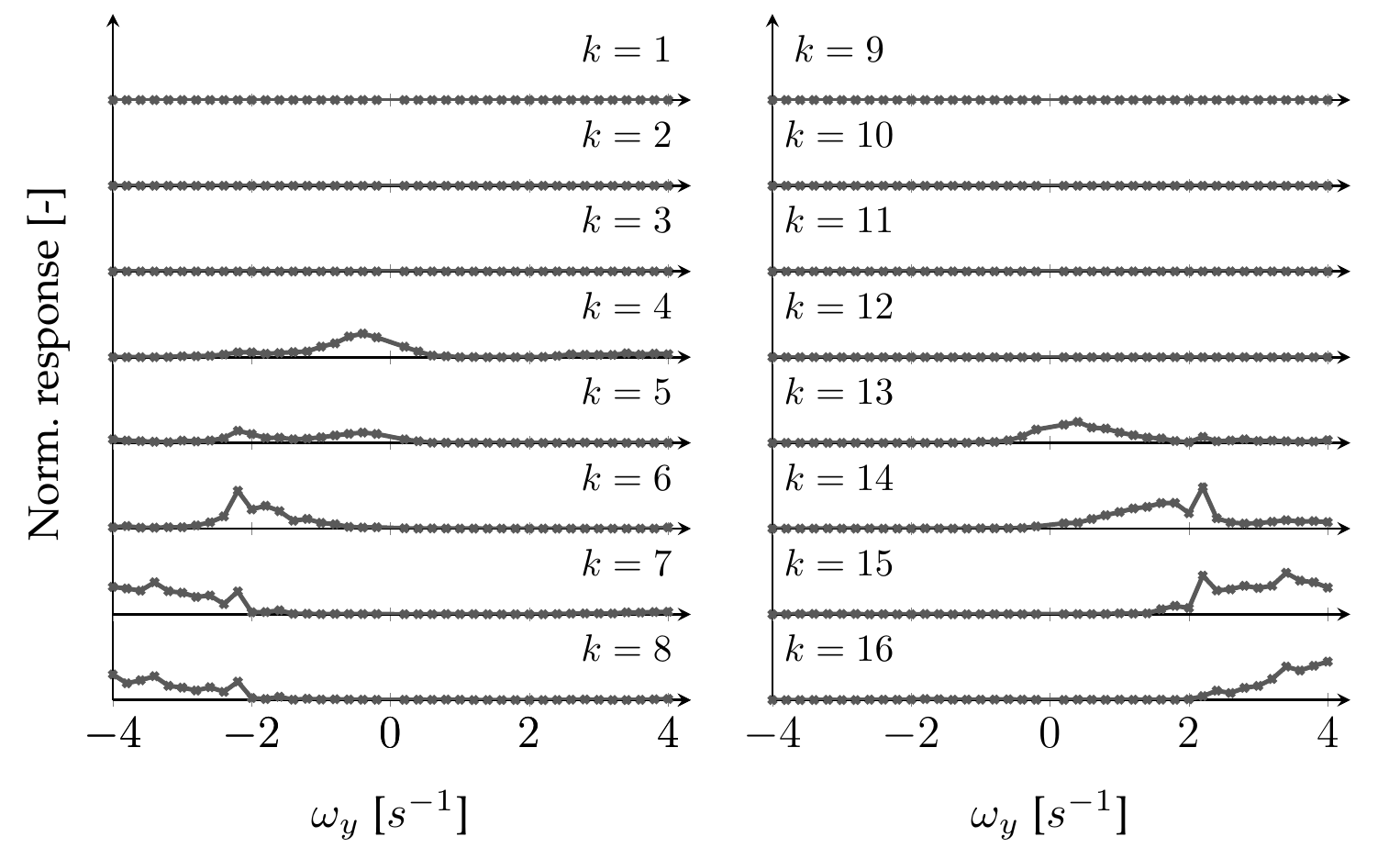}
		}
		\caption{Appearance (top) and neural response (bottom) of the sixteen spatiotemporal kernels learned from the checkerboard texture in the MS-Conv layer. Response plots are normalized by the maximum kernel response on the stimuli evaluated: 8.2763 spikes/ms by $k=11$ for $\omega_{x} = 4.0$ $s^{-1}$. Synaptic strength is encoded with brightness using the kernel formulation from \eqnref{eqmsconv:1}, i.e. $\smash{W^{\text{exc}} + \beta W^{\text{inh}}}$.}
		\label{figarch:7}
	\end{figure*}
	
	\setcounter{figure}{5}
	\begin{figure}[!b]
		\centering
		\includegraphics[width=0.275\textwidth]{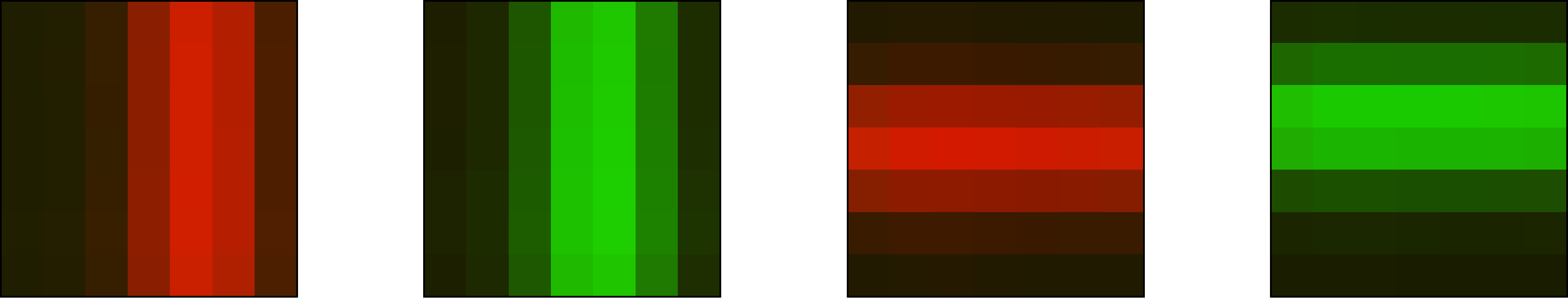}
		\caption{SS-Conv kernels learned from the checkerboard texture. Synaptic strength is encoded in color brightness: green for input neurons with positive (event) polarity, and red for negative.}
		\label{figbars:1}
	\end{figure}
	\setcounter{figure}{7}
	
	\section{Experimental Results}\label{sec:7}
	
	In this section, we evaluate the performance of our SNN on synthetic and real event sequences. Appendix B includes illustrations of the textures and natural scenes employed for generating these sequences, together with other implementation details, such as network parameters, sensor characteristics, training settings, and data augmentation mechanisms.
	
	\subsection{Synthetic Data Experiment}
	Firstly, we assess our motion-selective architecture on several noise-free sequences restricted to the pure vertical and horizontal image motion of a checkerboard pattern. This very structured texture and motion facilitate the understanding of the behavior and main properties of the network. Visual stimuli and ground truth were generated with the DVS simulator \cite{mueggler2017event}, and this analysis is based on the planar optical flow formulation from \cite{hordijk2017vertical} (see Appendix C).
	
	\begin{figure*}[!t]
		\centering
		\subfloat[Pure horizontal global motion\label{figarch:9a}]{%
			\includegraphics[width=0.495\textwidth]{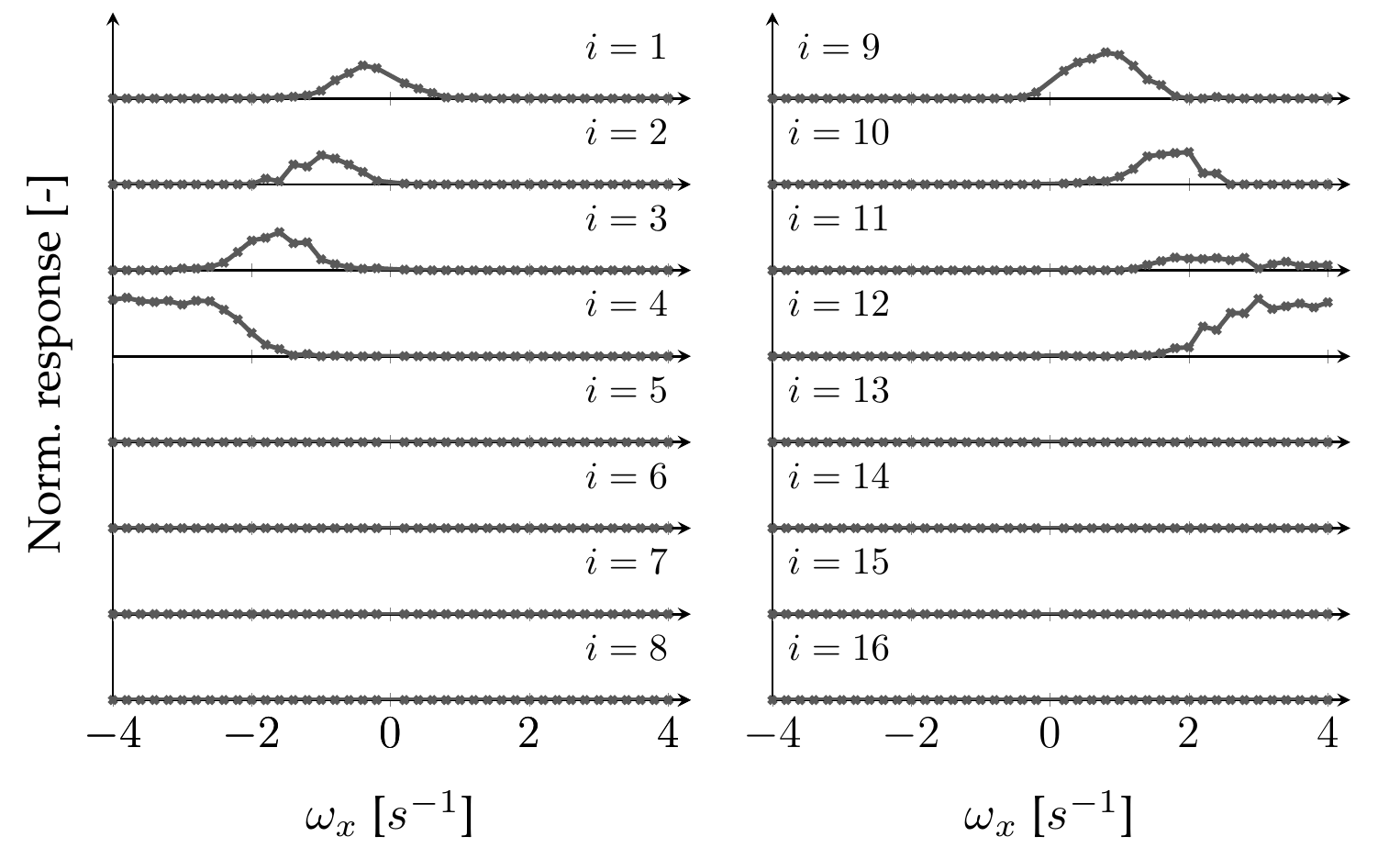}
		}
		\subfloat[Pure vertical global motion\label{figarch:9b}]{%
			\includegraphics[width=0.495\textwidth]{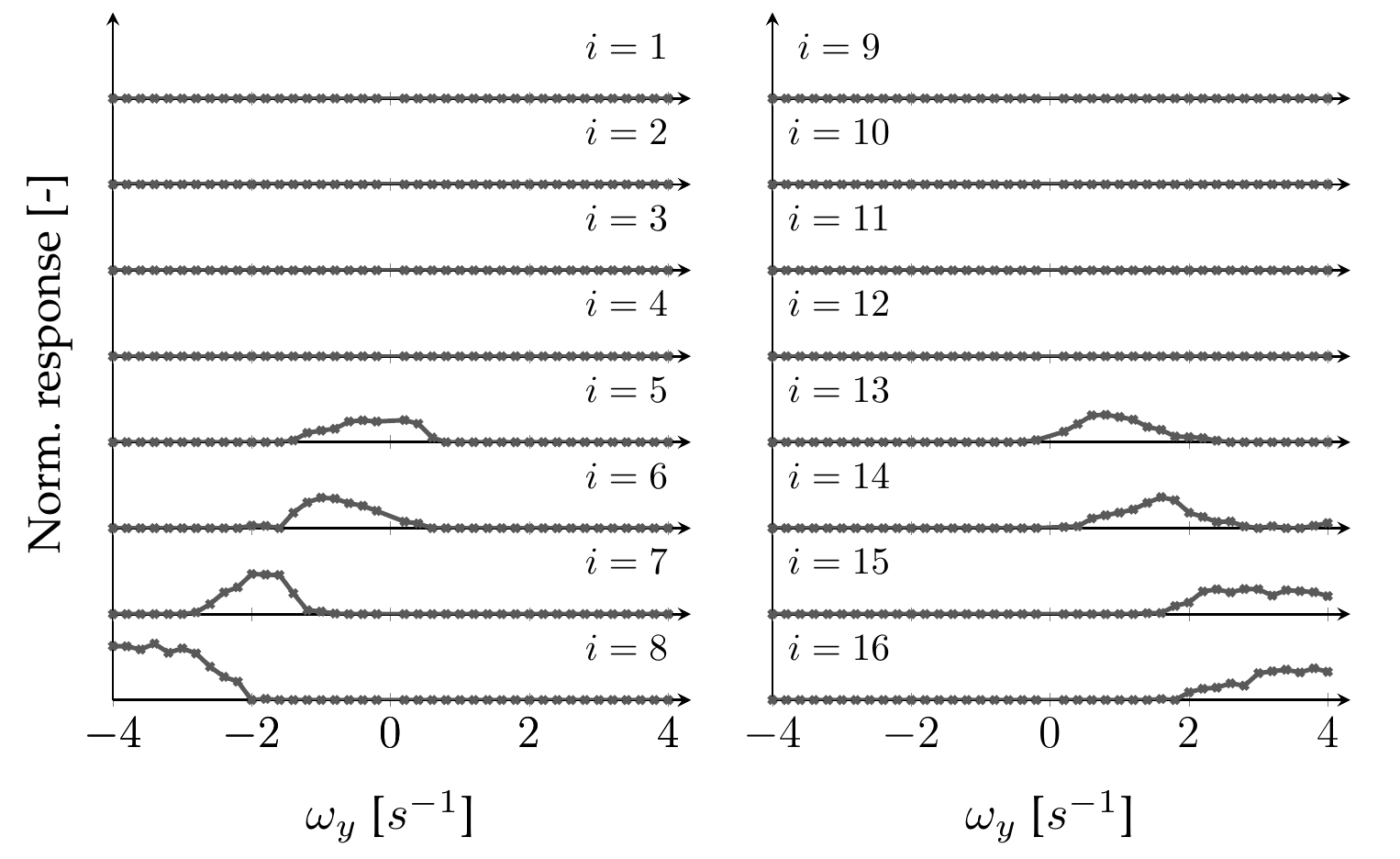}
		}
		\caption{Neural response of the sixteen individual neurons from the Dense layer trained in the checkerboard texture. Response plots are normalized by the maximum neural response on the stimuli evaluated: $0.3$ spikes/ms by $i=4$ for $\omega_{x} = -3.8$ $s^{-1}$.}
		\label{figarch:9}
	\end{figure*}
	
	Starting with the SS-Conv layer, \figref{figbars:1} shows the four convolutional kernels learned from these sequences. With this kernel scale, our learning rule leads to the successful identification of edges at the different spatial orientations present in the input data, and with the two combinations of event polarity. Using these kernels for feature extraction, and aggregating their spiking activity in the Merge layer, an MS-Conv layer consisting of sixteen spatiotemporal kernels was trained thereafter. \figref{figarch:7} shows the appearance of these kernels after convergence, and the response of their corresponding neural maps as a function of the ventral flow components ($\omega_x, \omega_y$).
	
	This figure confirms that, with the connectivity pattern of the MS-Conv layer, STDP leads to the successful identification of the spatiotemporally-oriented traces of input features, and hence their local motion. Out of the sixteen kernels trained, seven specialized to pure horizontal motion, and the remaining nine to pure vertical. Each direction of motion (up, down, left, right) was captured by at least four kernels, which, in turn, were selective to a particular stimulus speed. For instance, upward motion was identified by kernels $\smash{k=\{13,14,15,16\}}$, from slow to fast tuning speed. Therefore, kernels in this layer can be understood as local velocity-tuned filters that resemble those employed in frequency-based optical flow methods \cite{adelson1985spatiotemporal, tschechne2014bio, brosch2015event, orchard2013spiking}. However, instead of being manually designed, these filters emerge from visual experience in an unsupervised fashion. A three-dimensional illustration of two MS-Conv kernels can be found in Appendix D.2.
	
	In addition, remarkable is the fact that two of the (generally) four kernels that specialized to each of the aforementioned motion directions have overlapping neural responses despite the WTA mechanism described in \secref{found:2.1}. This is indicative of the relatively weak speed selectivity of MS-Conv neurons in comparison to their strong direction selectivity. Appendix D.3 confirms these results through an evaluation of both selectivities as a function of $\beta$.
	
	Lastly, selectivity to global motion emerges in neurons from a Dense layer trained as the final stage of the SNN, using the low-dimensional activity of the Pooling layer. \figref{figarch:9} shows the neural response (after convergence) of the sixteen cells in this layer as a function of ($\omega_x, \omega_y$). From this figure, it can be seen that neurons are successful at capturing the dominant global motion pattern from the spatial distribution of local motion estimates from previous layers. Out of the neurons trained, groups of four specialized to each motion direction, with different tuning speeds. Note that the velocity-selective properties of these neurons are exclusively dependent on those of the MS-Conv kernels. Appendix D.4 includes an evaluation of the temporal activity of these neurons in response to speed profiles that differ from the constant-speed sequences employed for learning.
	
	\subsection{Real Data Experiments}
	For the experiments with real data, we use samples from different sources. In a first evaluation, we employ the rotating-disk sequence from \cite{rueckauer2016evaluation}, which provides input events corresponding to a disk slowly turning at a constant speed. Furthermore, several unconstrained recordings of a roadmap pattern are used in a second experiment characterized by more unstructured and noisy visual stimuli. For this, we also use natural scene sequences from the Event Camera Dataset \cite{mueggler2017event} for validation. The DAVIS \cite{brandli2014240} and SEES1 \cite{sees1} are the DVS sensors with which this data was generated.
	
	\begin{figure}[!b]
		\centering
		\subfloat[Rotating disk\label{ssconvs:a}]{%
			\includegraphics[width=0.475\textwidth]{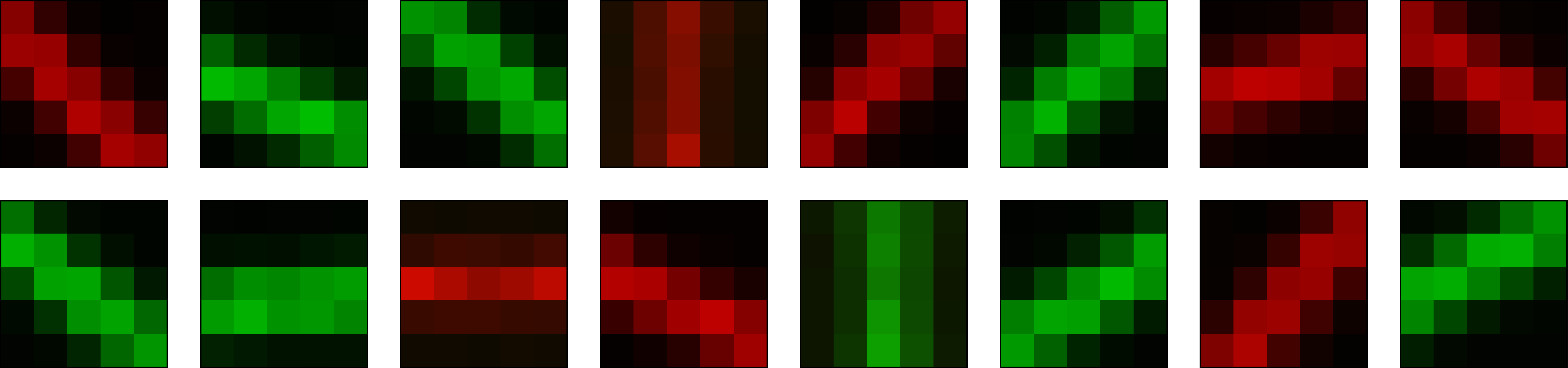}
		}
		\hspace{15pt}
		\subfloat[Roadmap\label{ssconvs:b}]{%
			\includegraphics[width=0.475\textwidth]{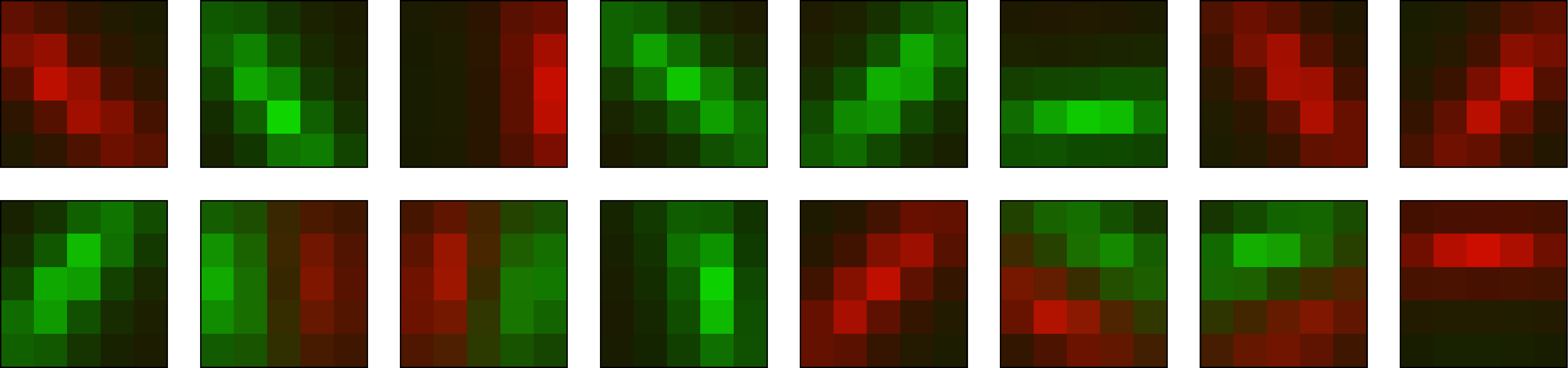}
		}
		\caption{SS-Conv kernels learned from real sequences. Synaptic strength is encoded in color brightness.}
		\label{ssconvs}
	\end{figure}
	
	\subsubsection{Rotating-Disk Sequence}
	\figref{ssconvs:a} shows the appearance of the SS-Conv kernels trained on the rotating-disk sequence. Similarly to the checkerboard case, neurons in this layer become selective to the most frequent input features, which are edges at different spatial orientations, and of different event polarity.
	
	\begin{figure}[!b]
		\vspace{-9pt}
		\centering
		\subfloat[Rotating disk\label{figcolor1}]{%
			\includegraphics[width=0.2\textwidth]{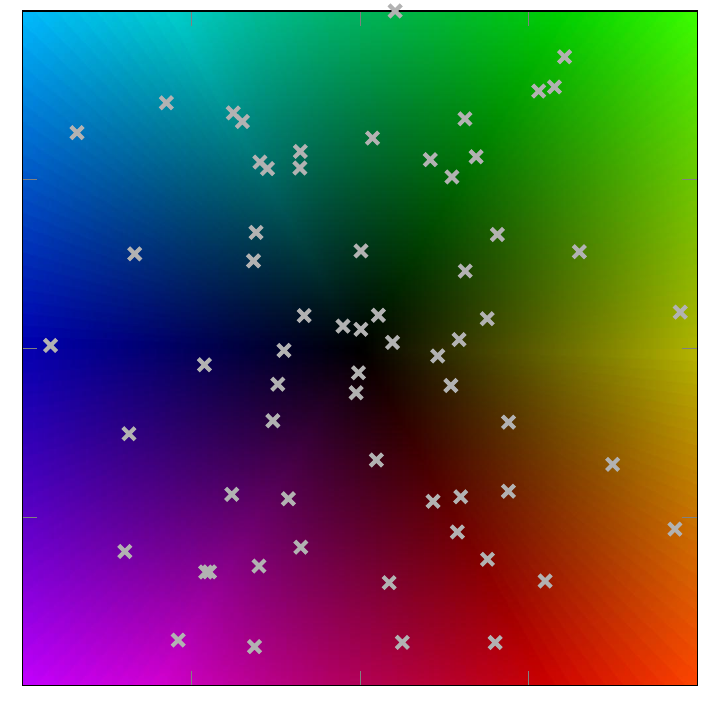}}
		\hspace{15pt}
		\subfloat[Roadmap\label{figcolor2}]{%
			\includegraphics[width=0.2\textwidth]{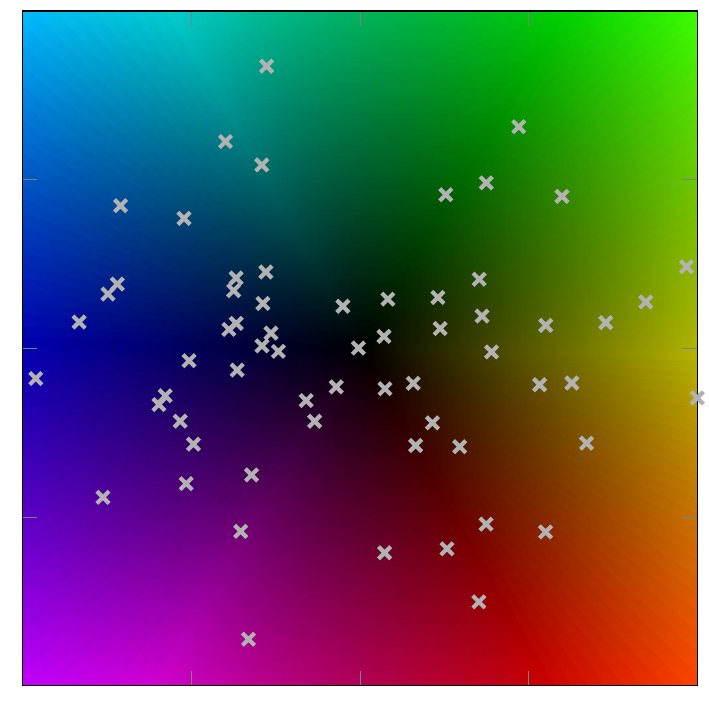}}
		\caption{MS-Conv kernels learned from real sequences in the (normalized) optical flow space, as in Appendices D.5 and D.6. Motion direction is encoded in color hue, and speed in color brightness. Each kernel is depicted as a cross.}
		\label{figcolor}
	\end{figure}
	
	\begin{figure}[!b]
		\centering
		\includegraphics[width=0.5\textwidth]{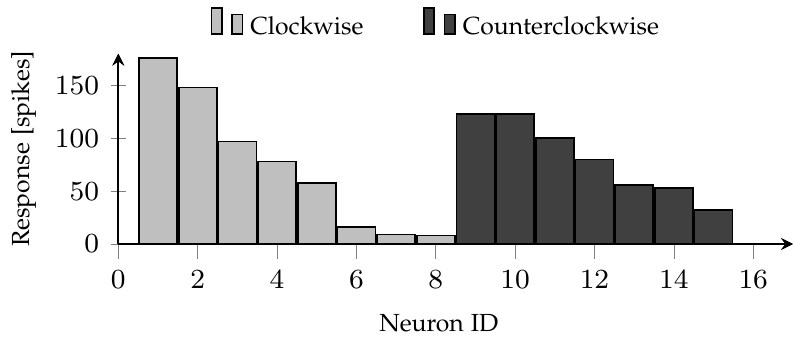}
		\caption{Neural activity of the Dense layer trained in the rotating-disk sequence, in response to the two global motion patterns in this recording.}
		\label{densedisk}
	\end{figure}
	
	\begin{figure*}[!t]
		\centering
		\includegraphics[width=\textwidth]{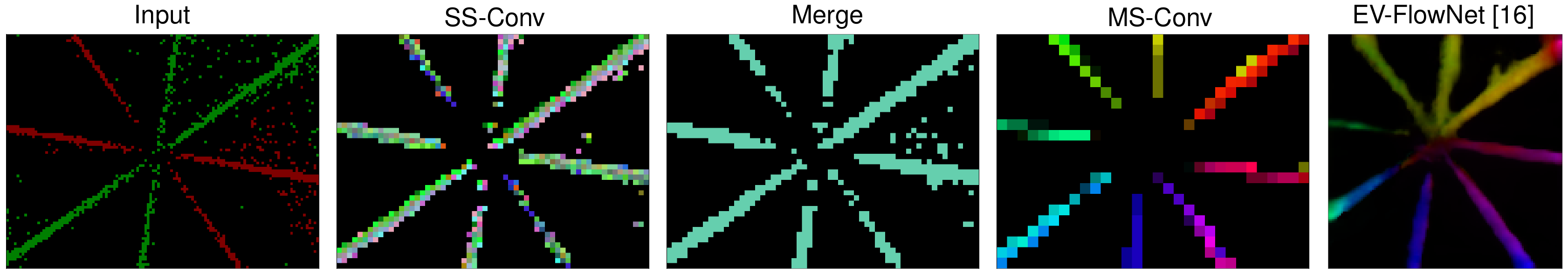}\\\vspace{2pt}
		\includegraphics[width=\textwidth]{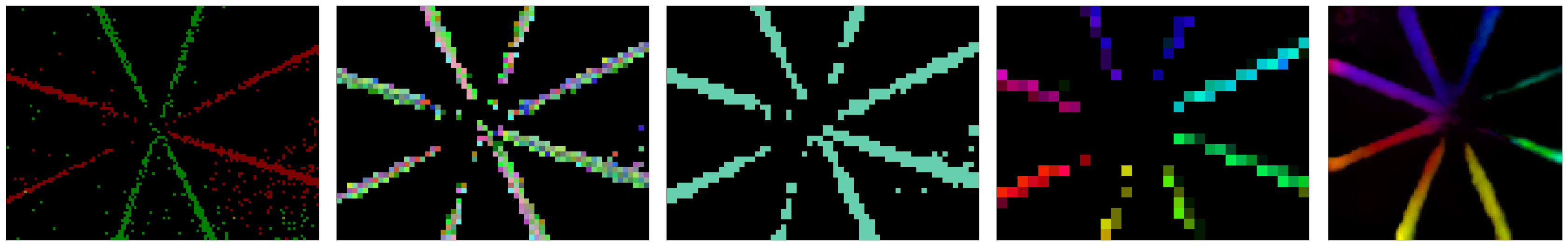}\\\vspace{2pt}
		\includegraphics[width=\textwidth]{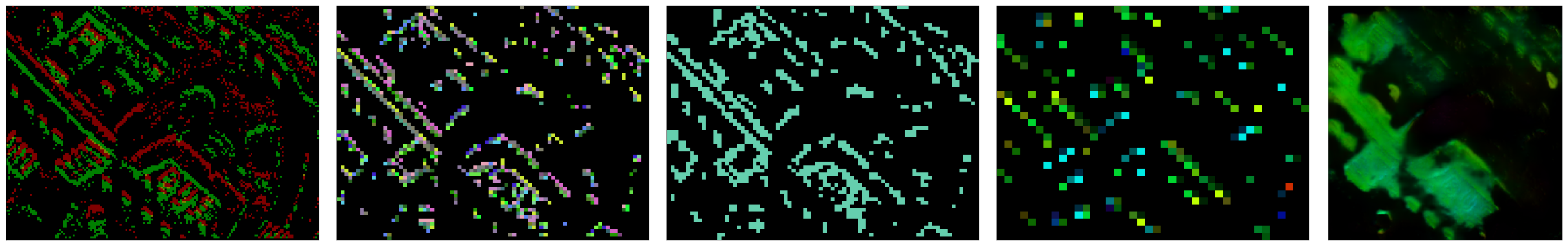}\\\vspace{2pt}
		\includegraphics[width=\textwidth]{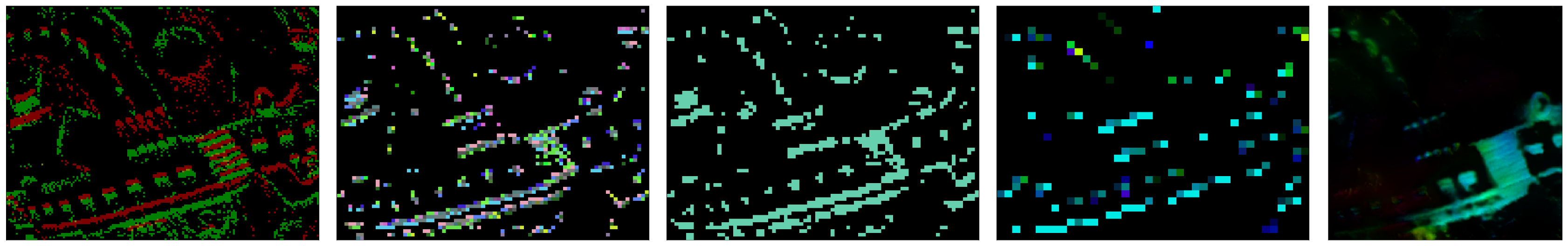}\\\vspace{2pt}
		\includegraphics[width=\textwidth]{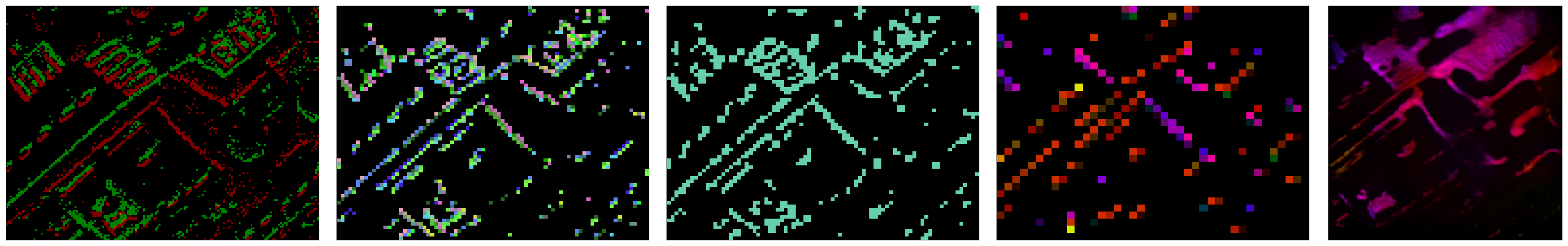}\\\vspace{2pt}
		\includegraphics[width=\textwidth]{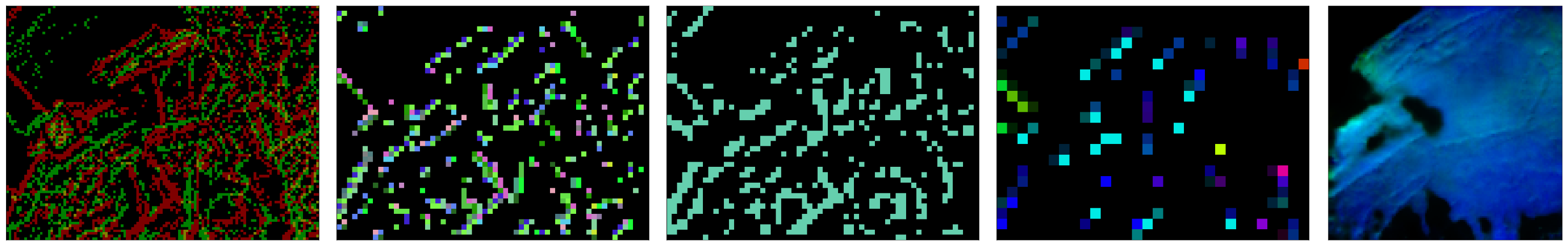}\\\vspace{2pt}
		\includegraphics[width=\textwidth]{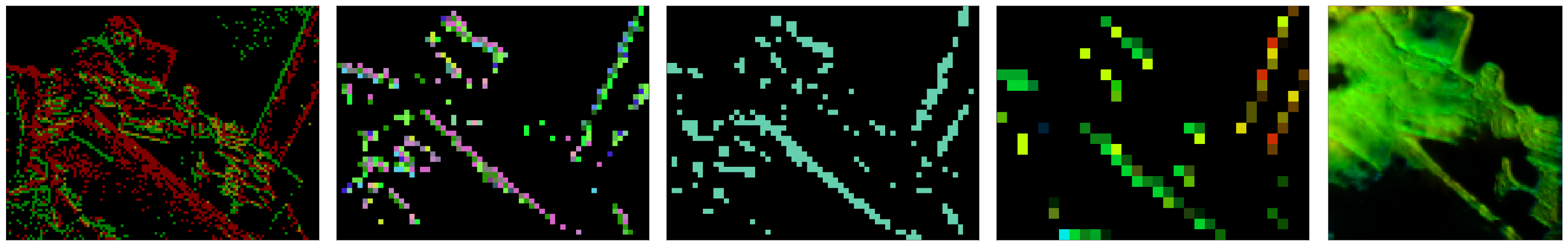}
		\caption{Qualitative results from the evaluation on real event sequences. From left to right, the first column corresponds to the input events, the following three to the spiking response of the SS-Conv, Merge, and MS-Conv layers, respectively; and the last column to the optical flow estimation of EV-FlowNet \cite{zhu2018ev}. A color is assigned to each of the kernels comprising the SS-Conv, Merge, and MS-Conv layers. MS-Conv color reference shown in \figref{figcolor}, and computed as in Appendices D.5 and D.6. SS-Conv and Merge color references not shown in this paper.}
		\label{rotating disk}
	\end{figure*}
	
	With respect to the MS-Conv layer of this architecture, \figref{figcolor1} shows its 64 kernels in the (normalized) optical flow space, according to the method explained in Appendices D.5 and D.6. From this figure, we observe that, through our STDP rule, these MS-Conv kernels learn to identify a wide variety of optical flow vectors, including diagonal motion at different speeds. The performance of this layer in local motion perception can be assessed from the qualitative results in \figref{rotating disk} (first two rows). Here, we compare the response of the network at this stage to the output of EV-FlowNet \cite{zhu2018ev}, which represents the state-of-the-art of conventional ANNs in event-based optical flow estimation. From these results, in both the clockwise and counterclockwise sequences, the response of the MS-Conv layer resembles that of EV-FlowNet, thus confirming the validity of our SNN in local motion perception. Additional qualitative results are provided in the supplementary video (see Appendix D.9).
	
	\begin{figure*}[!b]
		\vspace{-10pt}
		\centering
		\captionsetup[subfigure]{oneside,margin={0.5cm,0cm}}
		\subfloat[Horizontal global motion\label{denseroadmapa}]{\includegraphics[width=0.345\textwidth]{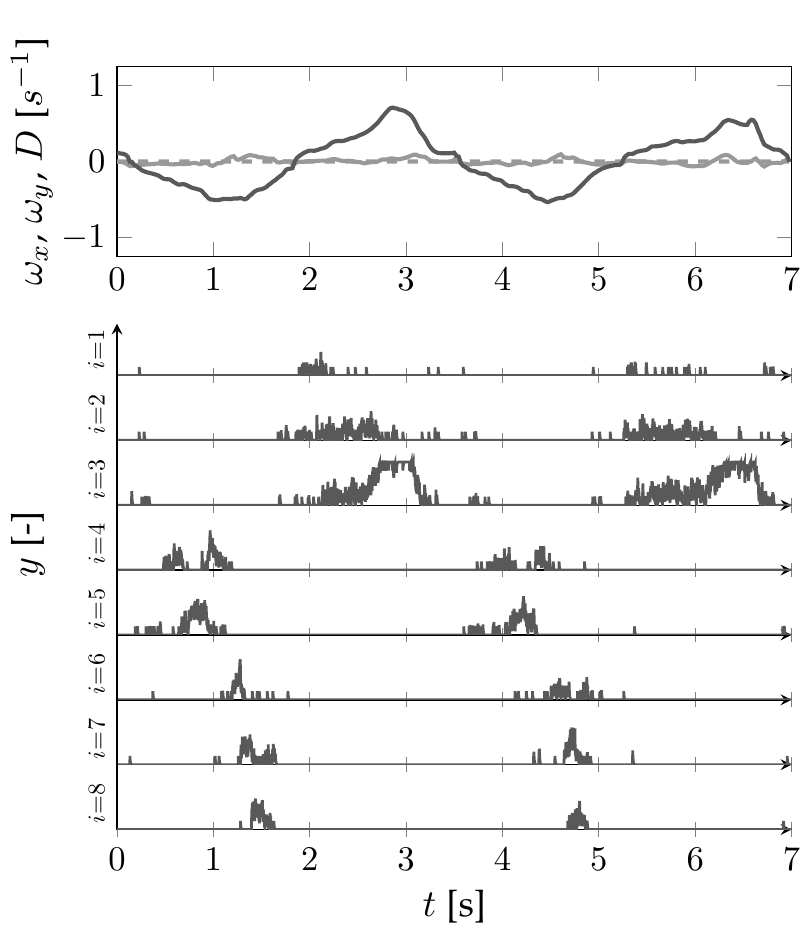}}
		\captionsetup[subfigure]{oneside,margin={0.2cm,0cm}}
		\subfloat[Vertical global motion]{\raisebox{-.001\height}{\includegraphics[width=0.324\textwidth]{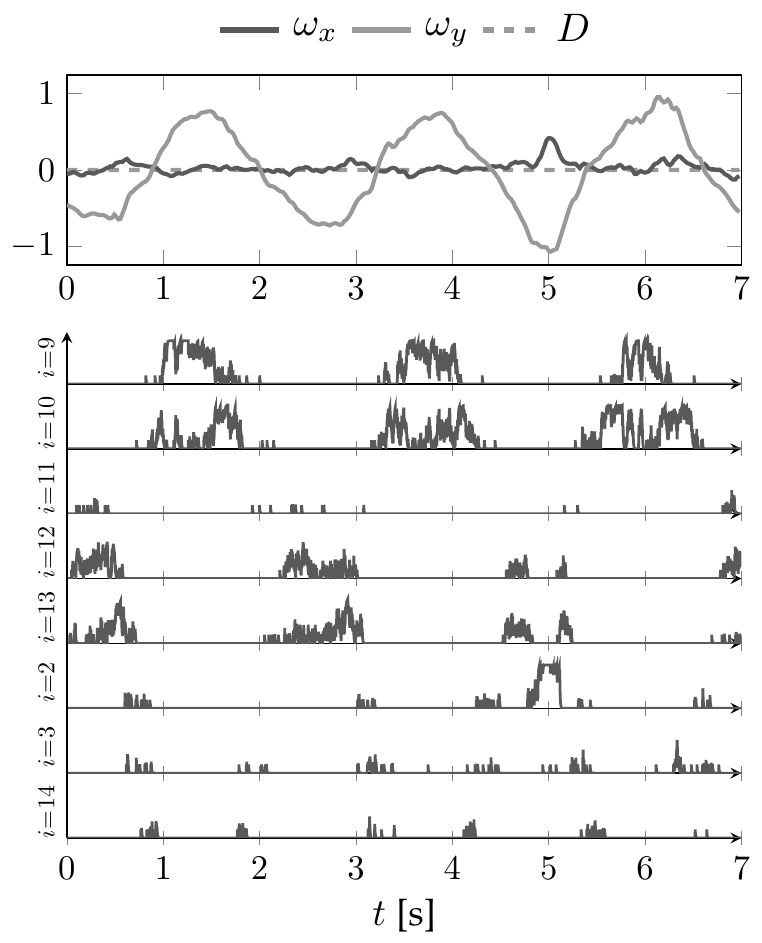}}}
		\subfloat[Diagonal global motion]{%
			\raisebox{-.001\height}{\includegraphics[width=0.325\textwidth]{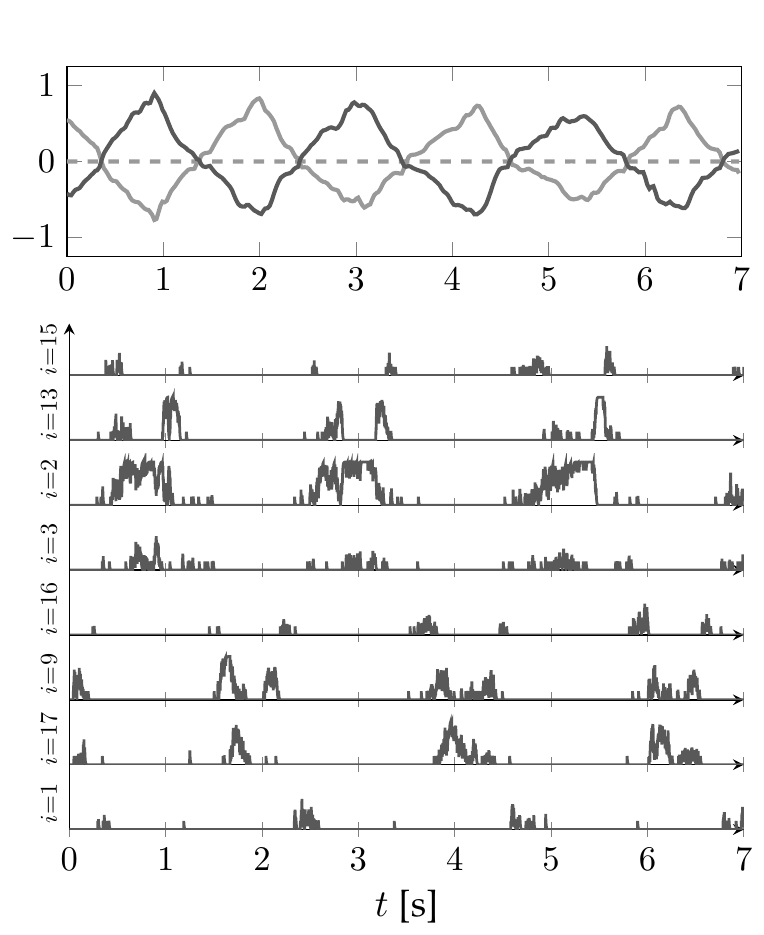}}}
		\caption{Temporal course of the postsynaptic trace (as in Appendix D.4) of the eight most-active neurons (for each case) from the Dense layer learned from the roadmap texture (bottom), in response to different global planar motion patterns (top). Plots are normalized by the maximum trace on the stimuli evaluated: $1.0$ by $i=3$ at $t=3.0$ s for the horizontal motion case. Optical flow visual observables ($\omega_x$, $\omega_y$, $D$) computed from the event sequences with the planar optical flow formulation from \cite{hordijk2017vertical} (see Appendix C).}
		\label{denseroadmap}
	\end{figure*}
	
	Lastly, a Dense layer comprised of sixteen neurons was trained, and the response of its cells is shown in \figref{densedisk}. As expected, the two global motion patterns present in the data are successfully captured: half of the neurons react to clockwise rotation, and the rest to counterclockwise. Besides competition, the different response levels are due to distinct distributions of local motion estimates in the Pooling layer leading to the same global motion pattern.
	
	\subsubsection{Roadmap Texture and Natural Scenes}
	\figref{ssconvs:b} shows the appearance of the SS-Conv kernels from the SNN trained on roadmap recordings. Similarly to those obtained with the rotating disk, these kernels learned edges (and combinations thereof) at several orientations, and of different polarities. However, note that kernel appearance is significantly less smooth due to the unstructured and low-contrast features of this texture, besides the sensor noise.
	
	Regarding the MS-Conv layer, \figref{figcolor2} shows its 64 spatiotemporal kernels in the (normalized) optical flow space (according to Appendices D.5 and D.6). In this figure, we observe that despite the wide variety of vectors learned, these are not as uniformly distributed as for the rotating-disk case. One can see that, first, horizontal motion is the most frequent local image motion type in the roadmap recordings; and second, the unsupervised nature of STDP prioritizes frequent features over others, less frequent, that may be more distant in this two-dimensional space.
	
	\begin{figure*}[!t]
		\centering
		\vspace{-32.5pt}
		\includegraphics[width=\textwidth]{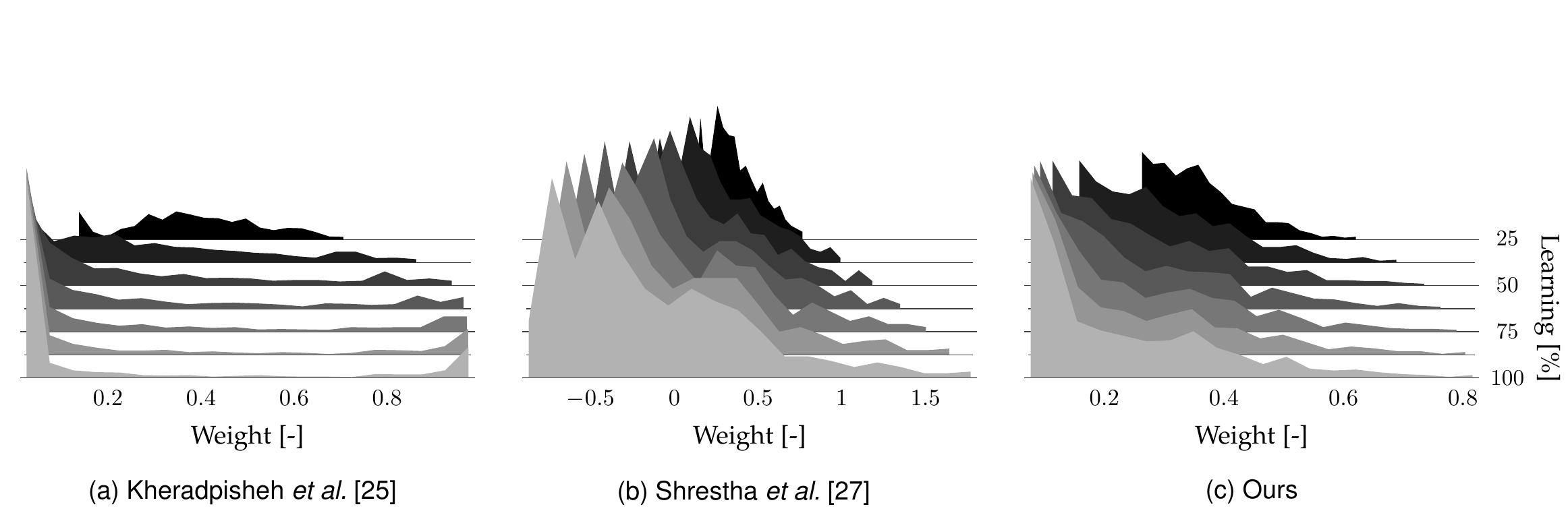}
		\caption{Evolution of the weight distribution of sixteen SS-Conv kernels throughout the learning process, using Kheradpisheh's \cite{kheradpisheh2018stdp}, Shrestha's \cite{shrestha2017stable}, and our STDP formulation. Results obtained with the roadmap texture, the same learning rate, and the same budget of training sequences. Each distribution is normalized by its maximum value in the learning process: $534$ synapses with $W\approx0.025$ for (a), $130$ with $W\approx-0.871$ for (b), and $219$ with $W\approx0.077$ for (c); all from the $100\%$ learning segment.}
		\label{stdpcomparison}
	\end{figure*}
	
	Qualitative results of the network performance up to this layer are shown in \figref{rotating disk} for roadmap and natural scene recordings (last five rows). We draw several conclusions from these results. Firstly, the SS-Conv layer is a key component of the architecture, since it successfully filters out inconsistent local events sequences, which benefits the learning and performance of subsequent layers. Secondly, the optical flow estimation of EV-FlowNet \cite{zhu2018ev} validates our MS-Conv layer, since it estimates highly similar optical flow vectors. However, there is a significant difference between the estimates of these two approaches, besides resolution (i.e. detail level). EV-FlowNet \cite{zhu2018ev} performs best in high texture regions, providing a semi-dense estimate of the local motion. On the other hand, our network only provides local motion estimates whenever and wherever it discerns features whose spatiotemporal trace fits one of the MS-Conv kernels. Due to trace overlap, no estimation is provided for image regions with high feature density. This limitation comes from the working principle of this layer, which takes inspiration from frequency-based optical flow methods \cite{adelson1985spatiotemporal} and bio-inspired motion detectors \cite{reichardt1961autocorrelation, barlow1965mechanism}, and for which these regions are also problematic. Additional qualitative results are provided in the supplementary video (see Appendix D.9).
	
	Lastly, \figref{denseroadmap} shows the temporal activity of some of the 32 neurons comprising the Dense layer of this architecture, in response to several global planar motion patterns. These results confirm the validity of this layer, and hence of the entire SNN, in becoming selective to this motion information through STDP. Moreover, similarly to the rotating-disk case, these results reinforce that, since notably different distributions of local motion estimates may correspond to the same global motion type, multiple Dense neurons can specialize to the same motion pattern without overlapping responses. This is further explained and illustrated in Appendix D.7 for neurons $\smash{i=\{4,\ldots,8\}}$ from \figref{denseroadmapa}.
	
	\subsection{STDP Evaluation}
	The final experiment of this work consists in an evaluation of several STDP formulations in the task of learning the kernels of an SS-Conv layer from the recordings of the roadmap texture. Specifically, we compare our rule, as in \secref{found:2}, to those proposed by Kheradpisheh \textit{et al.} \cite{kheradpisheh2018stdp}, and Shrestha \textit{et al.} \cite{shrestha2017stable}; two of the most recent multiplicative formulations that have successfully been used for image classification with SNNs. \figref{stdpcomparison} shows the weight distribution evolution of the SS-Conv kernels throughout the learning process, using each of the aforementioned formulations. Kernel appearance after learning is shown in Appendix D.8.
	
	The working principle of all STDP formulations is essentially the same. Whenever a neuron fires, the presynaptic connections that transferred the input spikes causing the firing are potentiated, while those that did not are depressed. The differences are in how the relevance of a connection is determined, and in how it is taken into account to compute the weight update $\Delta W$. Both Kheradpisheh's \cite{kheradpisheh2018stdp} and Shrestha's \cite{shrestha2017stable} formulations use temporal windows of fixed length to determine whether an input spike, and so its corresponding synapse, had an influence on the postsynaptic firing. However, this information is only employed to determine whether a synapse is potentiated or depressed, and not in the computation of $\Delta W$. On the one hand, Kheradpisheh's weight update is proportional to the current weight: $\smash{\Delta W \propto W_{i,j,d}(1-W_{i,j,d})}$. Results show that this rule leads to the learning of ambiguous features that fail to capture the spatiotemporal properties of the input, since all the weights become either null or unitary (see \figref{stdpcomparison}a). On the other hand, Shrestha's rule incorporates the weight dependency in an inversely proportional manner: $\smash{\Delta W \propto e^{-W_{i,j,d}}}$ for potentiation, and $\smash{\Delta W \propto -e^{W_{i,j,d}}}$ for depression. As shown, even though the $\Delta W$ for potentiation (depression) diminishes as the weights increase (decrease), weights keep increasing (decreasing) throughout the learning process (see \figref{stdpcomparison}b), and hence constraints to prevent them from exploding (vanishing) are required. The use of these constraints would, in turn, result in a bimodal weight distribution similar to that of Kheradpisheh's rule, with the aforementioned drawbacks.
	
	As explained in \secref{sec:5}, and to the best of the authors' knowledge, our STDP implementation is the first multiplicative formulation in incorporating synaptic relevance in the computation of $\Delta W$, resulting in an update rule whose LTP and LTD processes are not mutually exclusive. We combine (normalized) presynaptic trace information as a measure of synaptic relevance, with the inversely proportional weight dependency from \cite{shrestha2017stable}. Results, and the stability proof included in Appendix A, confirm that with our novel STDP formulation, an equilibrium weight is established for each synapse, towards which the weights converge throughout the learning process (see \figref{stdpcomparison}c). Since the equilibrium state depends on synaptic relevance, the features learned are successful at capturing the spatiotemporal properties of the input.\vfill

	
	\section{Conclusion}\label{sec:8}
	In this paper, we have presented the first SNN in which selectivity to the local and global motion of the visual scene emerges through STDP from event-based stimuli. The success of this emergence depends on three contributions. First, an adaptive spiking neuron model is necessary to handle the rapidly varying input statistics of event-based sensors, and we present a novel suitable formulation for this purpose. Second, we introduce a novel STDP implementation that, contrary to the current state-of-the-art of this learning protocol, is inherently stable. Third, we propose an SNN architecture that learns to perform a hierarchical feature extraction, effectively capturing geometric features, identifying the local motion of these features, and integrating this information into a global ego-motion estimate. We hope that this work, and the framework published alongside it, will provide the first step of many towards highly efficient artificial motion perception.
	
	

	
	\vskip -1.5\baselineskip plus -1fil
	\begin{IEEEbiography}[{\includegraphics[width=1in,height=1.25in,clip,keepaspectratio]{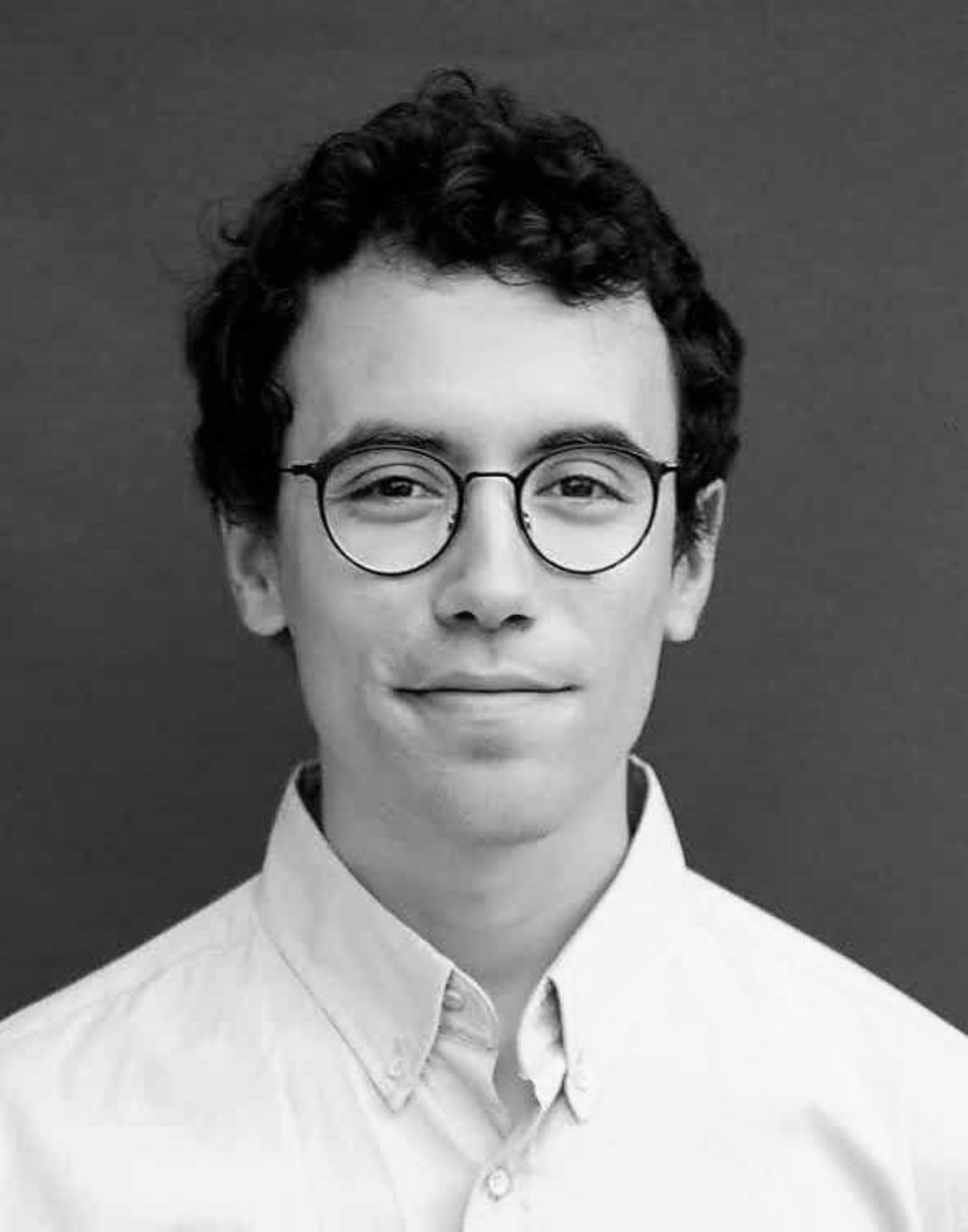}}]{Federico Paredes-Vall\'es}
		received his B.Sc. in Aerospace Engineering from the Polytechnic University of Valencia, Spain, in 2015, and his M.Sc. from Delft University of Technology, the Netherlands, in 2018. He is currently a Ph.D. candidate in the Micro Air Vehicle Laboratory at the latter university. His research interest is the intersection of machine learning, neuroscience, computer vision, and robotics.
	\end{IEEEbiography}
	\vskip -1.5\baselineskip plus -1fil
	\begin{IEEEbiography}[{\includegraphics[width=1in,height=1.25in,clip,keepaspectratio]{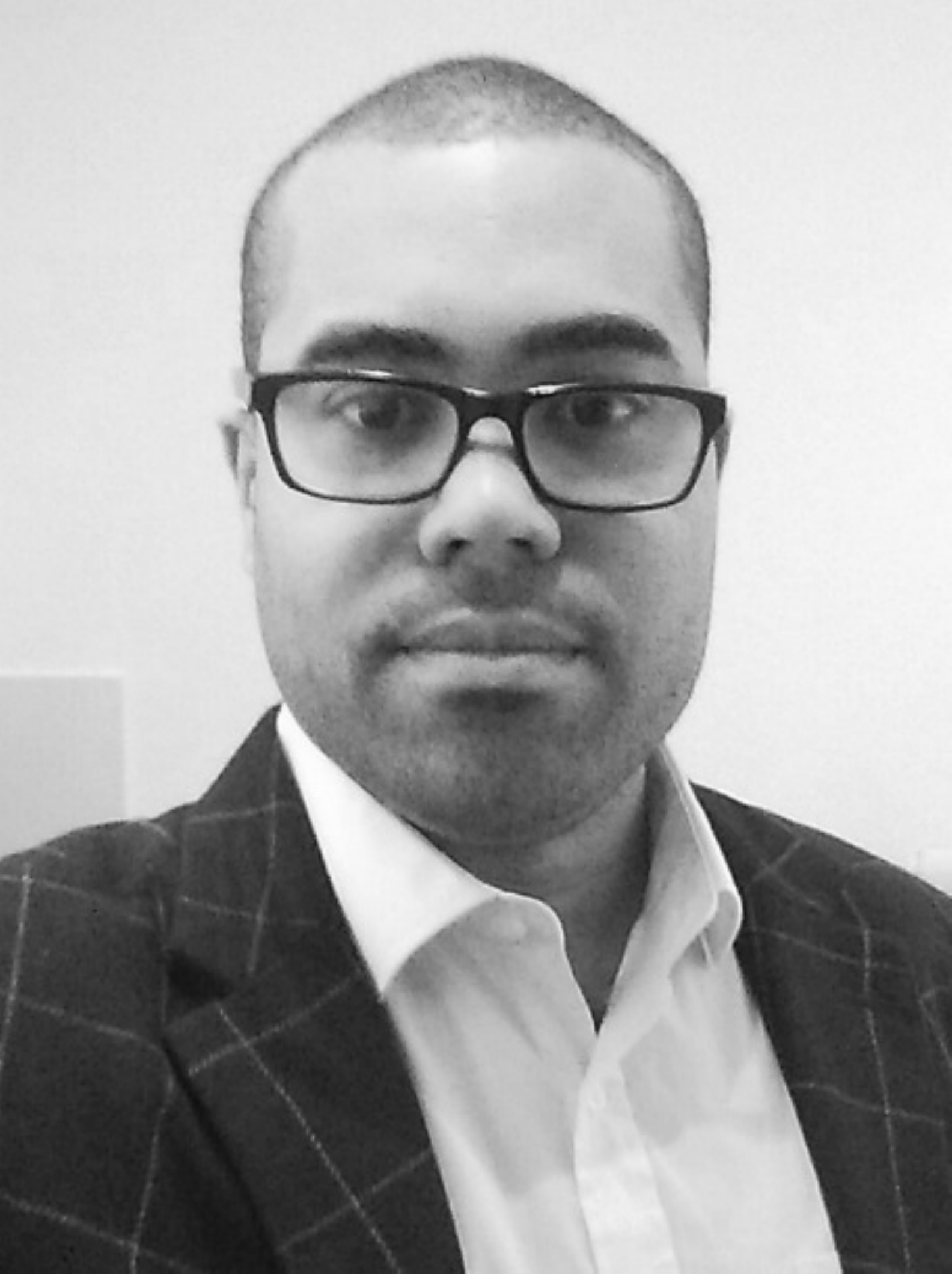}}]{Kirk Y. W. Scheper}
		received his M.Sc. from the Faculty of Aerospace Engineering at Delft University of Technology, the Netherlands, in 2014. Since then, he has been a Ph.D. candidate in the Micro Air Vehicle Laboratory at the same university. His research focuses on the development of embedded software which facilitates high level autonomy of micro air vehicles. His work is mainly in the fields of evolutionary robotics, embodied cognition, and vision-based navigation.
	\end{IEEEbiography}
	\vskip -1.5\baselineskip plus -1fil
	\begin{IEEEbiography}[{\includegraphics[width=1in,height=1.25in,clip,keepaspectratio]{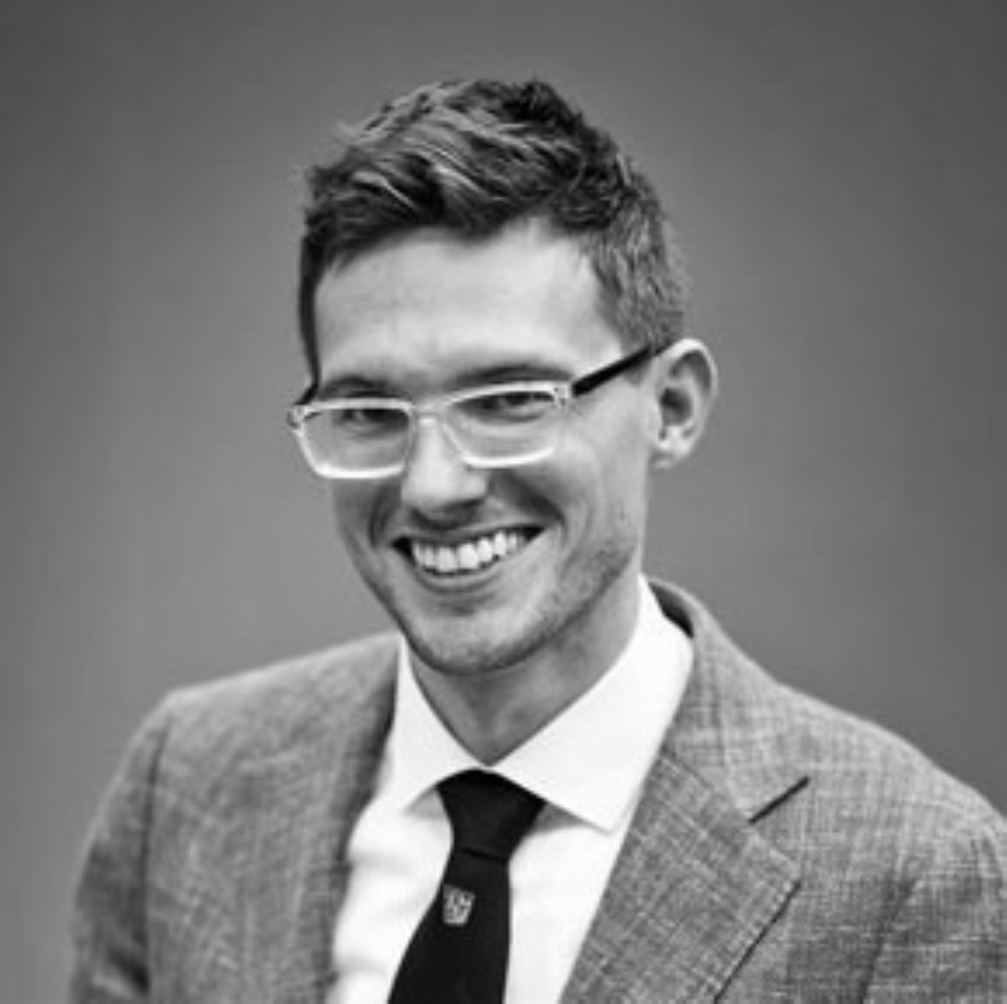}}]{Guido C. H. E. de Croon}
		received his M.Sc. and Ph.D. in the field of Artificial Intelligence at Maastricht University, the Netherlands. His research interest lies with computationally efficient algorithms for robot autonomy, with an emphasis on computer vision. Since 2008 he has worked on algorithms for achieving autonomous flight with small and light-weight flying robots, such as the DelFly flapping wing MAV. In 2011-2012, he was a research fellow in the Advanced Concepts Team of the European Space Agency, where he studied topics such as optical flow based control algorithms for extraterrestrial landing scenarios. Currently, he is associate professor at Delft University of Technology, the Netherlands, where he is the scientific lead of the Micro Air Vehicle Laboratory.
	\end{IEEEbiography}


\clearpage
\pagestyle{empty}
\appendices

\setcounter{figure}{14}
\setcounter{equation}{9}

\section{Proof of Stability of Proposed STDP} \label{STDP_proof}
We use Lyapunov theorem to investigate the global stability of the STDP implementation that we propose in this work. To simplify the proof, we assume that neural connectivity is single-synaptic with no transmission delay. The STDP rule defined in Eqs. (4) and (5) can then be rewritten as:
\begin{equation}\label{eq:stdpeq}
\begin{aligned}
\Delta W_{i,j} = &\eta\Big(e^{-(W_{i,j}-w_{\text{init}})}\big(e^{\hat{X}_{i,j}}-a\big)\\
&-e^{(W_{i,j}-w_{\text{init}})}\big(e^{(1-\hat{X}_{i,j})}-a\big)\Big)
\end{aligned}
\end{equation}

Equilibrium weights $\smash{\bar{W}_{i,j}}$ are given by $\smash{\Delta W_{i,j}=0}$:
\begin{equation} \label{eq:wbar}
\begin{aligned}
\bar{W}_{i,j} = \frac{1}{2}\ln\left(\frac{e^{\hat{X}_{i,j}}-a}{e^{(1-\hat{X}_{i,j})}-a}\right)+w_{\text{init}}
\end{aligned}
\end{equation}

If we let $\smash{z=W_{i,j}-\bar{W}_{i,j}}$, \eqnref{eq:stdpeq} becomes:
\begin{equation}
\begin{aligned}
\Delta W_{i,j}=&\eta \big(e^{\hat{X}_{i,j}}-a\big)^{\frac{1}{2}} \big(e^{(1-\hat{X}_{i,j})}-a\big)^{\frac{1}{2}}\big(e^{-z}-e^{z}\big)\\
\Delta W_{i,j}=&A\big(e^{-z}-e^{z}\big)
\end{aligned}
\end{equation}
\noindent{where $\smash{A(\hat{X}_{i,j}, a)}$ is a convenience function containing all components that are not a function of $z$.}

Then we define the positive definite energy function $\smash{V(z)=\frac{1}{2}z^2}$. As such, $\smash{\dot{V}(z)}$ can be solved as follows:
\begin{equation}
\begin{aligned}
\dot{V}(z)=z\dot{z}=z\big(\Delta W_{i,j} - \Delta \bar{W}_{i,j}\big)
\end{aligned}
\end{equation}
\noindent{where $\smash{\Delta \bar{W}_{i,j}}$ can be computed from the time derivate of \eqnref{eq:wbar} as:}
\begin{equation} \label{eq:delta_wbar}
\begin{aligned}
\Delta \bar{W}_{i,j} =& \frac{1}{2}\Delta \hat{X}_{i,j}\left(\frac{e^{\hat{X}_{i,j}}}{e^{\hat{X}_{i,j}}-a} + \frac{e^{(1-\hat{X}_{i,j})}}{e^{(1-\hat{X}_{i,j})}-a}\right)
\end{aligned}
\end{equation}
\noindent{and $\smash{\Delta \hat{X}_{i,j}}$ can be determined using Eq. (1):}
\begin{equation}\label{eq:delta_xhat}
\begin{aligned}
\Delta \hat{X}_{i,j}=\frac{\alpha}{\lambda_{X} \hat{X}_{i,m}}\big(s_{j}^{l-1}-\hat{X}_{i,j}s_{m}^{l-1}\big)
\end{aligned}
\end{equation}
\noindent{where, here, the subscript $m$ denotes the index of the neuron with the maximum presynaptic trace.}

Combining Eqs. (14) and (15) we are left with:
\begin{equation}
\begin{aligned}
\Delta \bar{W}_{i,j} =& \frac{1}{2}\frac{\alpha}{\lambda_{X} \hat{X}_{i,m}}\left(\frac{e^{\hat{X}_{i,j}}}{e^{\hat{X}_{i,j}}-a} + \frac{e^{(1-\hat{X}_{i,j})}}{e^{(1-\hat{X}_{i,j})}-a}\right)\\
&\cdot \big(s_{j}^{l-1}-\hat{X}_{i,j}s_{m}^{l-1}\big) \\
=& B\big(s_{j}^{l-1}-\hat{X}_{i,j}s_{m}^{l-1}\big)
\end{aligned}
\end{equation}
\noindent{where $\smash{B(\hat{X}_{i,j}, a)}$ is a convenience expression containing all elements not a function of $\smash{s_{j}^{l-1}}$ and $\smash{s_{m}^{l-1}}$.}

Now the energy derivative can be expressed simply as:
\begin{equation}
\begin{aligned}
\dot{V}(z)=A z\big(e^{-z}-e^{z}\big) - Bz\big(s_{j}^{l-1}-\hat{X}_{i,j}s_{m}^{l-1}\big)
\end{aligned}
\end{equation}

Using the Taylor expansion of $\smash{e^z}$ and $\smash{e^{-z}}$, we are left with:
\begin{equation} \label{eq:vdot_final}
\begin{aligned}
\dot{V}(z)=&-2A\big(z^{2} + \frac{z^4}{3!} + \cdots\big) - Bz\big(s_{j}^{l-1}-\hat{X}_{i,j}s_{m}^{l-1}\big)
\end{aligned}
\end{equation}

Now if we look at the case where there is no external input to the neurons (i.e. the normalized presynaptic trace is constant, $\smash{s_{j}^{l-1}=s_{m}^{l-1}=0}$), we have that global asymptotic stability is guaranteed for $A>0$, which can be ensured by setting $\eta>0$ and $a<1$.

When considering the input, we can define bounded error $z$ with input-state stability inequality for a bounded input $\smash{u=\hat{X}_{i,j}s_{m}^{l-1}-s_{j}^{l-1}}$:
\begin{equation}
\|z(t)\|\leq \beta (\|z(t)\|, t-t_0) + \gamma\left(\sup_{\tau \geq t_0} \|u(\tau)\|\right), \forall t \geq t_0
\end{equation}
\noindent{where $\gamma$ is the so-called Lyapunov gain, which will lead to input-state stability if positive definite.}

Now, using the first order approximation for the Taylor expansion from \eqnref{eq:vdot_final}, we can show:
\begin{equation}
\begin{aligned}
\dot{V}(z) \leq -Az^2, 
\forall |z| \geq \frac{B\left(\hat{X}_{i,j}s_{m}^{l-1}-s_{j}^{l-1}\right)}{2A}
\end{aligned}
\end{equation}
\noindent{with Lyapunov gain $\gamma = \frac{B}{2A}$.}

As $A$ must be positive for global asymptotic stability, for $\gamma$ to be positive definite, $B$ must also be positive. As such, $\lambda_X$ and $\alpha$ must have the same sign. Additionally, the values of the constants in $A$ and $B$ can be used to control the bounds of the error $z$.

To give some physical meaning to these parameters we can see that adjusting $a$ will change the sensitivity of the STDP update to the presynaptic trace. The larger the difference $\smash{|1-a|}$, the less sensitive the update will be to the input. The time constant $\lambda_{X}$ will adjust the rate at which the presynaptic trace is updated from the inputs $\smash{s_{j}^{l-1}}$ and $\smash{s_{m}^{l-1}}$. The larger the time constant the slower the presynaptic trace will change and the more bounded the error will become. The scaling factor $\alpha$ changes the magnitude of the presynaptic trace and therefore the magnitude of the rate of change of the presynaptic trace.

One thing to note here is the discontinuity as $\smash{X_{i,m} \to 0}$. This shows that the bound of the error can become large if the maximum presynaptic trace is small and the current neuron being updated is not the neuron with the maximum presynaptic trace. Physically, this would mean that the network cannot accurately learn when the input is infinitely sparse. For the case where the input is measurably sparse, the learning can be improved by compensating with a larger time constant $\lambda_{X}$.

\section{Implementation Details} \label{Implementation_details}

\makeatletter
\def\Url@twoslashes{\mathchar`\/\@ifnextchar/{\kern-.2em}{}}
\g@addto@macro\UrlSpecials{\do\/{\Url@twoslashes}}
\makeatother

In \tabref{tabappB:1}, the parameters of each layer comprising the SNNs employed in this work are specified. All the experiments are conducted using our open-source CUDA-based \emph{cuSNN}\footnote{Source code available at \url{https://github.com/tudelft/cuSNN}} library, with a simulation timestep of $\Delta t_{\text{sim}} = 1$ ms. 

Concerning learning, these architectures are trained in a \textit{layer-by-layer} fashion using the unsupervised STDP rule presented in Section 4. Regardless of the layer type, the parameter $a$ from Eq. (5) is set to $0$, the initialization weight to $w_{\text{init}}=0.5$, the learning rate $\eta$ to $1\e{-4}$, and the convergence threshold to $\mathcal{L}_{\text{th}}=5\e{-2}$. As shown in Figs. 4 (right) and 14c, these parameters lead to weight distributions that, after convergence, are naturally constrained in the range $\boldsymbol{W}\in[0,1]$ for excitatory synapses, and $\boldsymbol{W}\in[-1,0]$ for inhibitory. Throughout the learning phase, short event sequences are presented sequentially at random following a uniform distribution.

\begin{table}[!t]
	\centering
	\begin{threeparttable}  	
		\setlength{\tabcolsep}{4.5pt}
		\renewcommand{\arraystretch}{1.3}
		\caption{SNNs Architectures and Parameters}
		\label{tabappB:1}
		\begin{tabular}{ccccccccc}
			\multicolumn{9}{l}{{\fontfamily{phv}\selectfont (a) Checkerboard Texture}}\\
			\hline
			Layer & $r$ [-] & $s$ [-] & $f$ [-] & $m$ [-] & $\tau$ [ms] & $v_{\text{th}}$ [-] & $\lambda$ [ms] & $\alpha$ [-]\\\hline
			SS-Conv & $7$ & $1$ & $4$ & $1$ & $1$ & $0.5$ & $5$ & $0.4$\\
			Merge & $1$ & $1$ & $1$ & $1$ & $1$ & $0.001$ & $5$ & $-$\\
			MS-Conv & $7$ & $2$ & $16$ & $10$ & $[1, 50]$ & $0.5$ & $5$ & $0.25$\\
			Pooling & $8$ & $8$ & $16$ & $1$ & $1$ & $0.001$ & $5$ & $-$\\
			Dense & $-$ & $-$ & $16$ & $1$ & $1$ & $0.5$ & $5$ & $0.25$\\
			\hline
		\end{tabular}\vspace{15pt}
		\begin{tabular}{ccccccccc}
			\multicolumn{9}{l}{{\fontfamily{phv}\selectfont (b) Rotating-Disk {[83]}}}\\
			\hline
			Layer & $r$ [-] & $s$ [-] & $f$ [-] & $m$ [-] & $\tau$ [ms] & $v_{\text{th}}$ [-] & $\lambda$ [ms] & $\alpha$ [-]\\\hline
			SS-Conv & $5$ & $2$ & $16$ & $1$ & $1$ & $0.3$ & $5$ & $0.1$\\
			Merge & $1$ & $1$ & $1$ & $1$ & $1$ & $0.001$ & $5$ & $-$\\
			MS-Conv & $5$ & $2$ & $64$ & $10$ & $[1, 200]$ & $0.3$ & $30$ & $0.1$\\
			Pooling & $6$ & $6$ & $64$ & $1$ & $1$ & $0.001$ & $5$ & $-$\\
			Dense & $-$ & $-$ & $16$ & $1$ & $1$ & $0.3$ & $30$ & $0.1$\\
			\hline
		\end{tabular}\vspace{15pt}
		\begin{tabular}{ccccccccc}
			\multicolumn{9}{l}{{\fontfamily{phv}\selectfont (c) Roadmap Texture and Natural Scenes {[82]}}}\\
			\hline
			Layer & $r$ [-] & $s$ [-] & $f$ [-] & $m$ [-] & $\tau$ [ms] & $v_{\text{th}}$ [-] & $\lambda$ [ms] & $\alpha$ [-]\\\hline
			SS-Conv & $5$ & $2$ & $16$ & $1$ & $1$ & $0.4$ & $5$ & $0.25$\\
			Merge & $1$ & $1$ & $1$ & $1$ & $1$ & $0.001$ & $5$ & $-$\\
			MS-Conv & $5$ & $2$ & $64$ & $10$ & $[1, 25]$ & $0.4$ & $15$ & $0.25$\\
			Pooling & $8$ & $8$ & $64$ & $1$ & $1$ & $0.001$ & $5$ & $-$\\
			Dense & $-$ & $-$ & $32$ & $1$ & $1$ & $0.4$ & $15$ & $0.25$\\
			\hline
		\end{tabular}
		\begin{tablenotes}
			\item \textit{The resting potential $v_{\text{rest}}$ is considered null, the refractory period $\smash{\Delta_{\text{refr}}}$ is set to 3 ms for (a), and to 1 ms for (b) and (c). For MS-Conv, $\beta$ is set to $0.5$, and the temporal delays are linearly spaced within the specified range. Note that $s$ denotes the convolutional stride of each layer.}
		\end{tablenotes}
	\end{threeparttable}
\end{table}

\begin{figure}[!t]
	\centering
	\subfloat[Checkerboard\label{figarch:2a}]{%
		\includegraphics[width=0.15\textwidth]{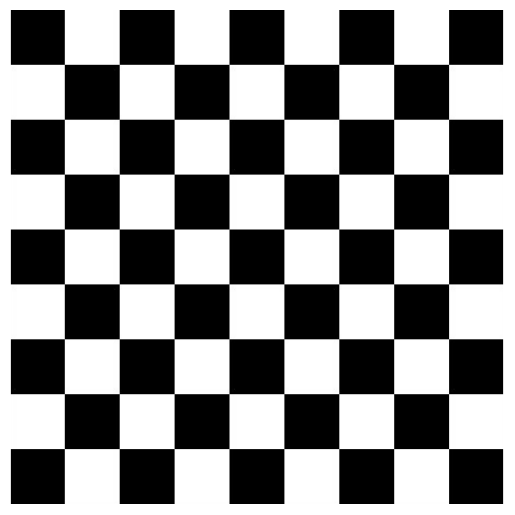}}
	\subfloat[Roadmap\label{figarch:2c}]{%
		\includegraphics[width=0.15\textwidth]{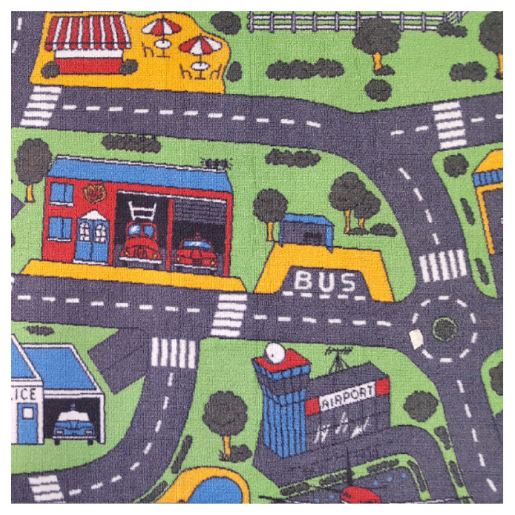}}
	\subfloat[Rotating disk {[83]}\label{figarch:2d}]{%
		\includegraphics[width=0.15\textwidth]{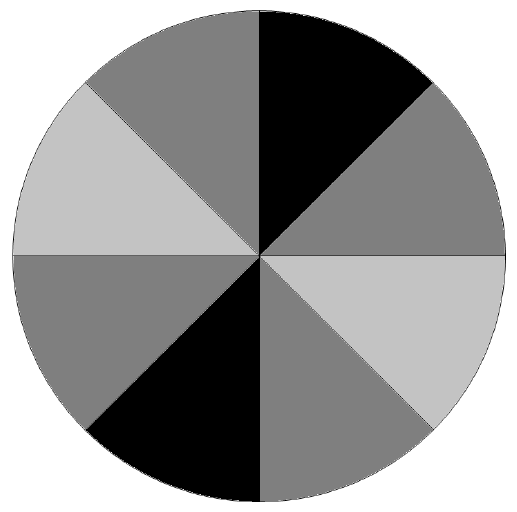}}\\
	\subfloat[Boxes {[82]}\label{figarch:2e}]{%
		\includegraphics[width=0.225\textwidth]{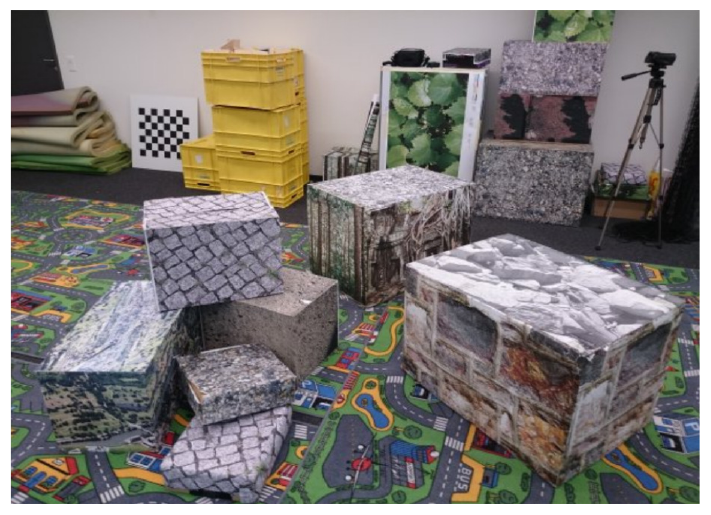}}
	\subfloat[Dynamic {[82]}\label{figarch:2f}]{%
		\includegraphics[width=0.225\textwidth]{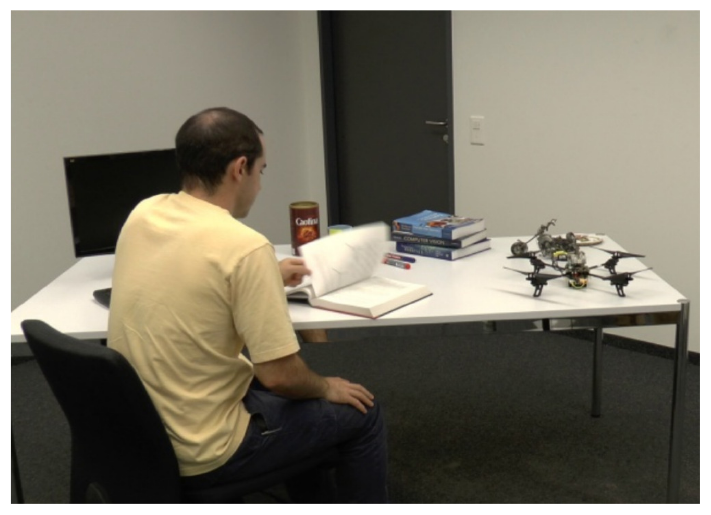}}
	\caption{Texture patterns and scenes employed for generating the synthetic (a) and real (b--$\hspace{0.25pt}$e) event sequences employed in this work.}
	\label{figarch:2}
\end{figure}

We evaluate the performance of our learning rule and hierarchical SNN architecture on several synthetic and real event sequences. For the generation of the former, we employ the DVS simulator {[82]} ($128\times 128$ pixel array) and the checkerboard pattern shown in \figref{figarch:2a}. This simulator renders intensity images from a three-dimensional virtual scene at a high rate (1000 Hz), and estimates events by linearly interpolating logarithmic brightness levels between frames, thus leading to noise-free event sequences. The motion of the simulated DVS was restricted to straight trajectories at different constant speeds, and at a fixed distance with respect to the (virtual) textured planar surface. This, together with the fact that the checkerboard provides high contrast and clear edges, facilitates the understanding of the behavior and main properties of the proposed network.

On the other hand, the real event sequences come from different sources. Firstly, we conduct experiments on the \textit{rotating-disk} event sequence from {[83]}, which corresponds to a circular disk with eight compartments of varying gray-levels turning at a constant speed in front of a DAVIS DVS sensor {[11]} ($240\times 180$ pixel array). The disk is shown in \figref{figarch:2d}. Secondly, we generated several recordings by moving the SEES1 DVS sensor {[12]} ($320\times 264$ pixel array) by hand in front of the roadmap pattern shown in \figref{figarch:2c}. This texture is largely characterized by unstructured and low-contrast visual features, thus leading to noisy event sequences. Note that, in this case, the degrees of freedom of the sensor are not constrained, and rotations and depth variations are present in the data along with translational motion at different speeds. Lastly, we also employ several of the recordings comprising the Event Camera Dataset {[82]} for validation; more specifically, the \textit{boxes} and \textit{dynamic} sequences shown in \figreftwo{figarch:2e}{figarch:2f}, respectively. Similarly to the roadmap sequences, these recordings of natural scenes contain motion in all six degrees of freedom.

Throughout the learning phase, and regardless of the data type and sensor employed, we perform random (with $50\%$ chance) spatial (i.e. horizontal and vertical) and polarity flips to the event sequences as data augmentation mechanisms. Moreover, in both the learning and inference phases, the spatial resolution of each sequence is downsampled to half its original size for computational efficiency purposes.

\section{Planar Optical Flow Formulation} \label{Planar}

The planar optical flow formulation from {[55]} relates the ego-motion of a vision sensor facing a static planar scene to the perceived optical flow and its corresponding visual observables. The derivation of this model is made in the camera-fixed reference frame centered at the focal point of the sensor, denoted by the subscript $\mathcal{C}$. Position is defined by the Cartesian coordinates $(X_\mathcal{C}, Y_\mathcal{C}, Z_\mathcal{C})$, with $(U_\mathcal{C}, V_\mathcal{C}, W_\mathcal{C})$ as the corresponding velocity components. Using the pinhole camera model {[84]}, the \textit{ventral flow} components ($\omega_{x}$, $\omega_{y}$) and the \textit{flow field divergence} $D$, which are the so-called optical flow visual observables, are defined as:
\begin{equation}\label{eqobs:5}
\begin{aligned}
\omega_{x} = -\frac{U_{\mathcal{C}}}{Z_{0}}, \quad  \omega_{x} = -\frac{V_{\mathcal{C}}}{Z_{0}}, \quad D = 2\frac{W_{\mathcal{C}}}{Z_{0}}
\end{aligned}
\end{equation}
\noindent{where $Z_{0}$ is defined as the distance to the surface along the optical axis of the sensor. According to {[55]}, ($\omega_{x}$, $\omega_{y}$) are a quantification of the average flows in the $X$- and $Y$-axis of $\mathcal{C}$, respectively. On the other hand, $D$ is a measure of the divergence of the optical flow field, thus measuring depth variations with respect to the planar surface.}

\section{Supplementary Material} \label{Supplementary}

\subsection{Effect of the Max-Based Homeostasis Formulation} \label{Homeo}
\begin{figure}[!b]
	\centering
	\subfloat[Checkerboard]{%
		\includegraphics[width=0.275\textwidth]{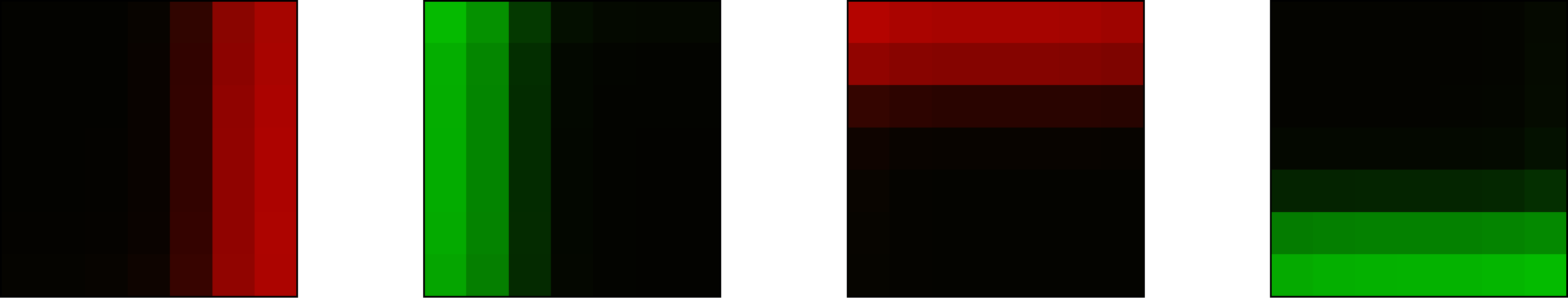}}\\
	\subfloat[Roadmap]{%
		\includegraphics[width=0.475\textwidth]{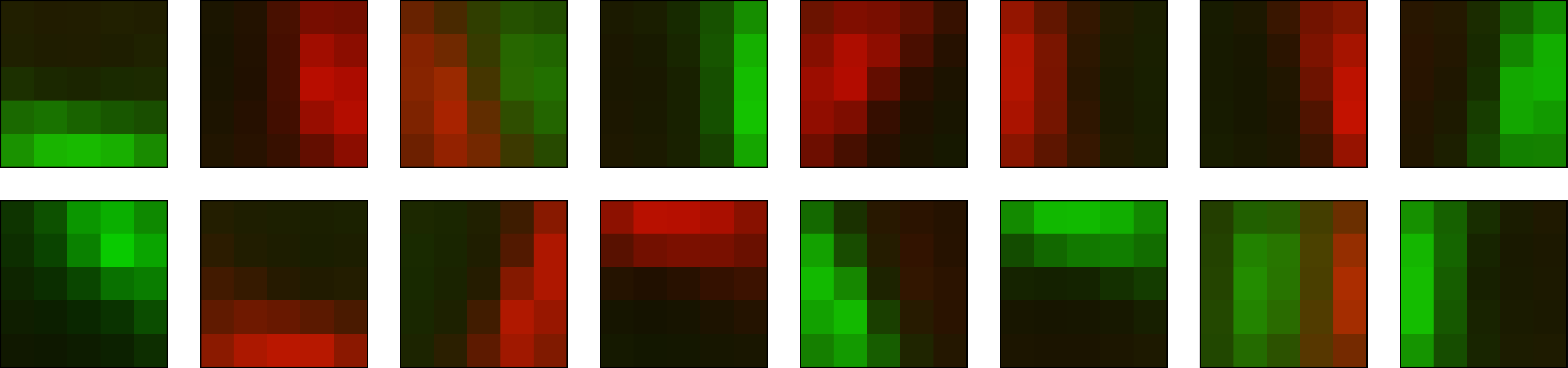}}
	\caption{SS-Conv kernels learned from synthetic (top) and real event sequences (bottom) with the neuron-specific homeostasis formulation. Synaptic strength is encoded in color brightness.}
	\label{homeo:1}
\end{figure}

\begin{figure}[!b]
	\centering
	\subfloat[$x$-\hspace{0.5pt}$\tau$ representation]{%
		\includegraphics[width=0.475\textwidth]{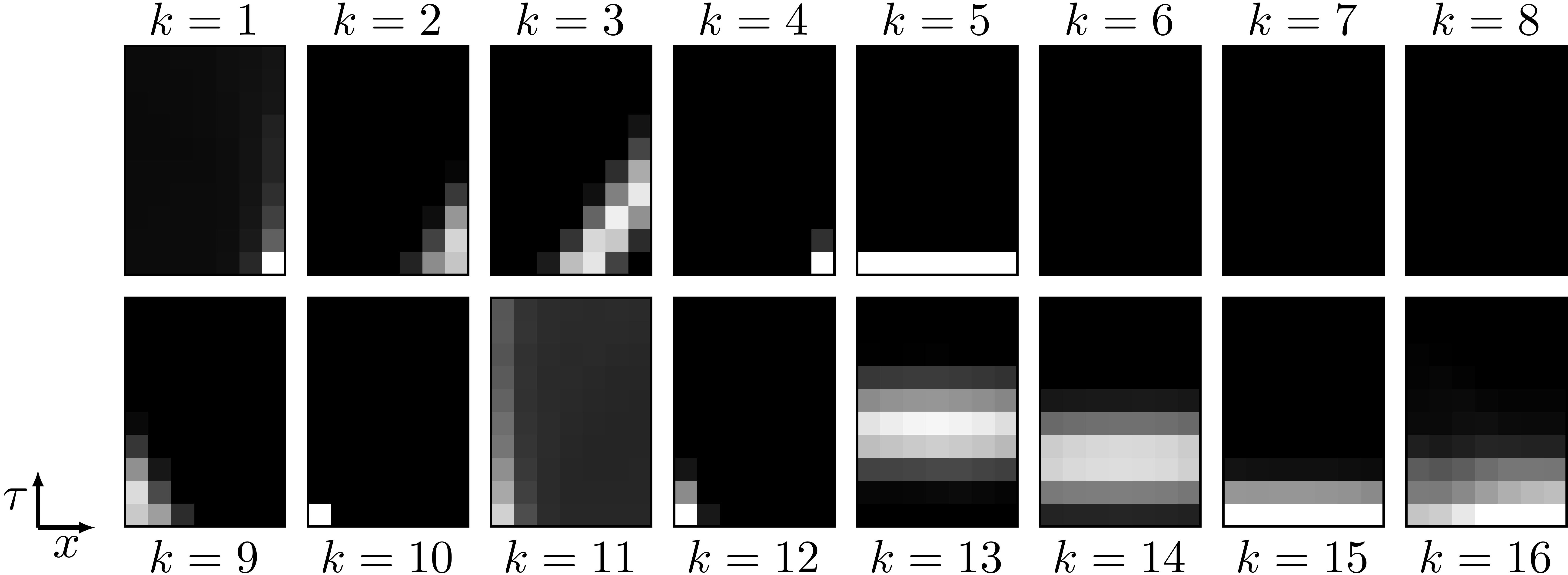}}\\
	\subfloat[$y$-\hspace{0.5pt}$\tau$ representation]{%
		\includegraphics[width=0.475\textwidth]{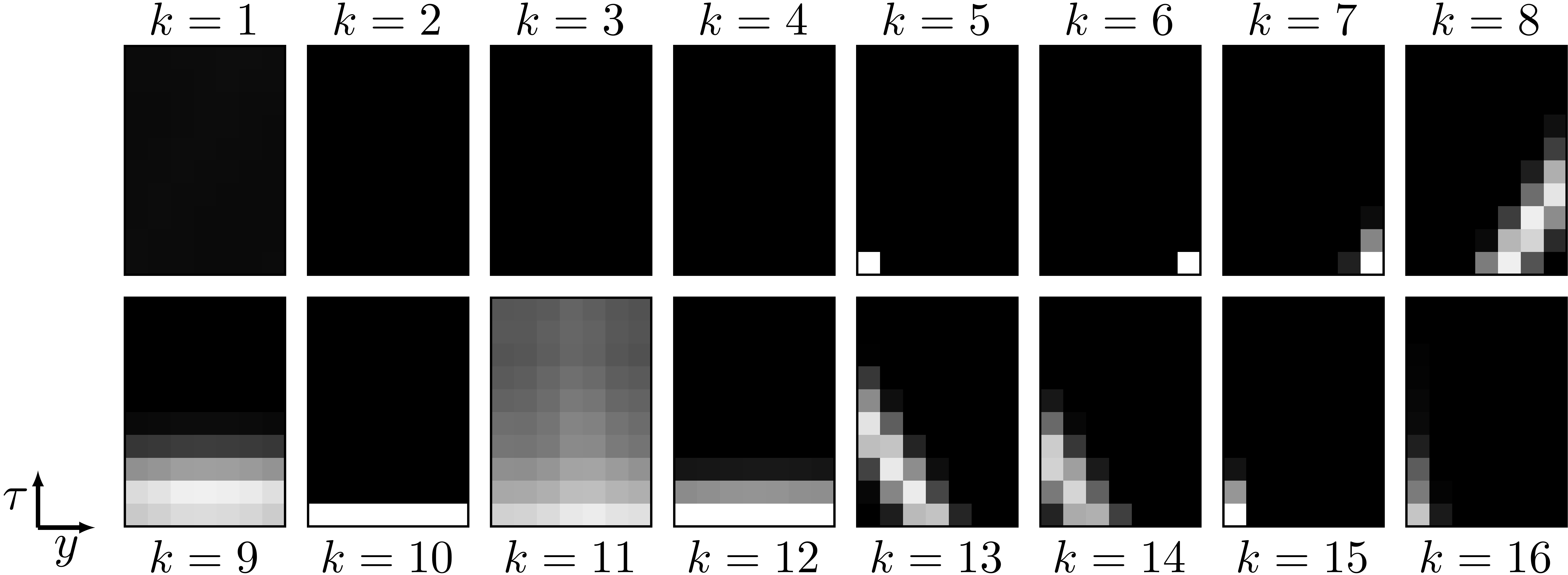}}
	\caption{MS-Conv kernels learned from the checkerboard texture with the neuron-specific homeostasis formulation. Synaptic strength is encoded with brightness.}
	\label{homeo:2}
\end{figure}

\figreftwo{homeo:1}{homeo:2} illustrate the need for the homeostasis parameter as detailed in Eqs. (7) and (8), which considers the maximum presynaptic trace of the direct spatial neighborhood $\boldsymbol{N}_{i,k}$ of the neuron under analysis, when dealing with layers characterized by retinotopically-arranged cells. For a better understanding, \figref{homeo:1} should be compared to Figs. 6 and 9a, and \figref{homeo:2} to Figs. 7a and 7b.

As shown, when the neuron-specific presynaptic trace is employed instead of the full homeostasis formulation, convolutional kernels specialize to the leading edge of moving features, and hence most of these kernels are characterized by more ambiguous synaptic configurations in which the strong synapses are mainly located on the receptive field borders. The effect of using this incomplete model on the performance of the SS-Conv layer is that a greater number of kernels is required to extract the same number of spatial features. However, the impact of this formulation is more visible in the MS-Conv layer. As shown in \figref{homeo:2}, the vast majority of MS-Conv kernels lose their velocity-selective properties, simply because the spatiotemporally-oriented traces of input features are no longer captured. The leading-edge specialization also makes the learning process more complex, since kernel overlap increases. This, in turn, leads to some of these kernels being always prevented from firing, i.e. prevented from triggering STDP (e.g. $k=11$).

\begin{figure}[!b]
	\centering
	\subfloat{%
		\includegraphics[width=0.2\textwidth]{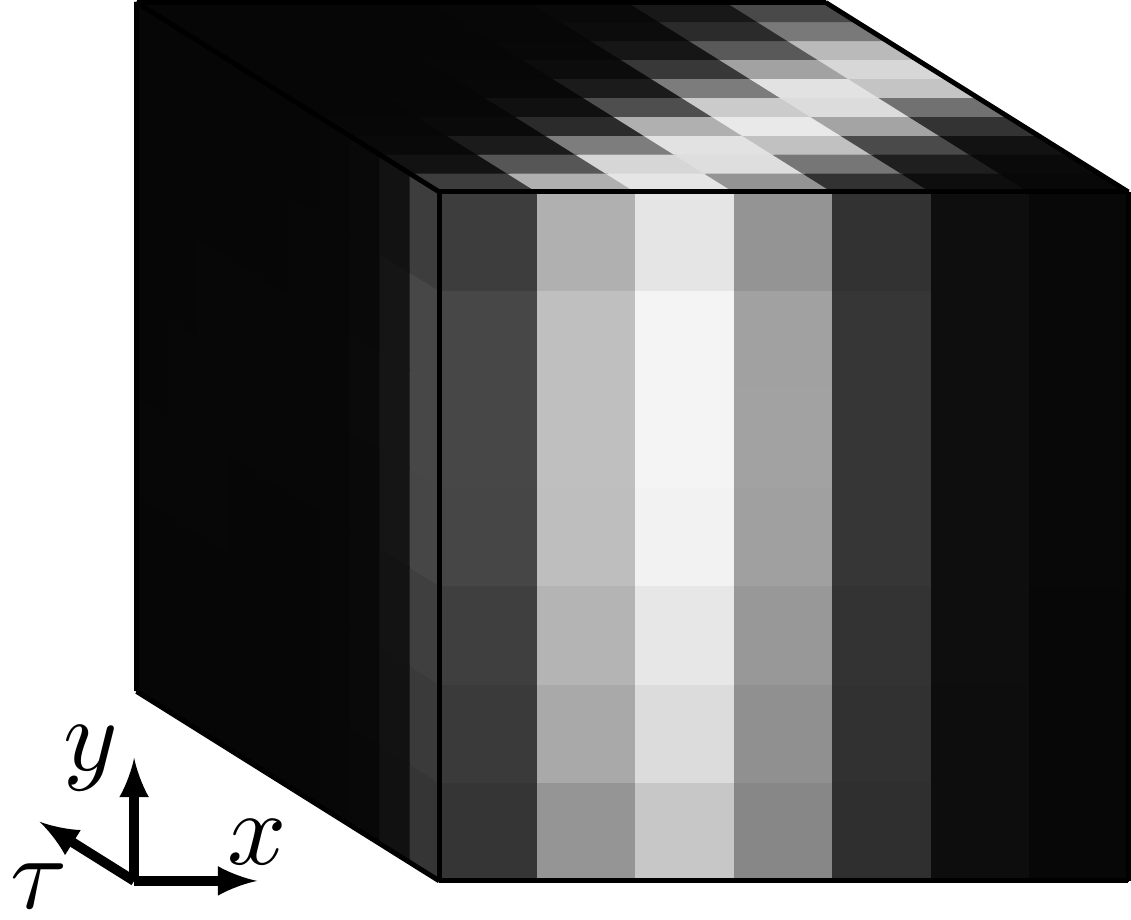}}
	\hspace{15pt}
	\subfloat{%
		\includegraphics[width=0.2\textwidth]{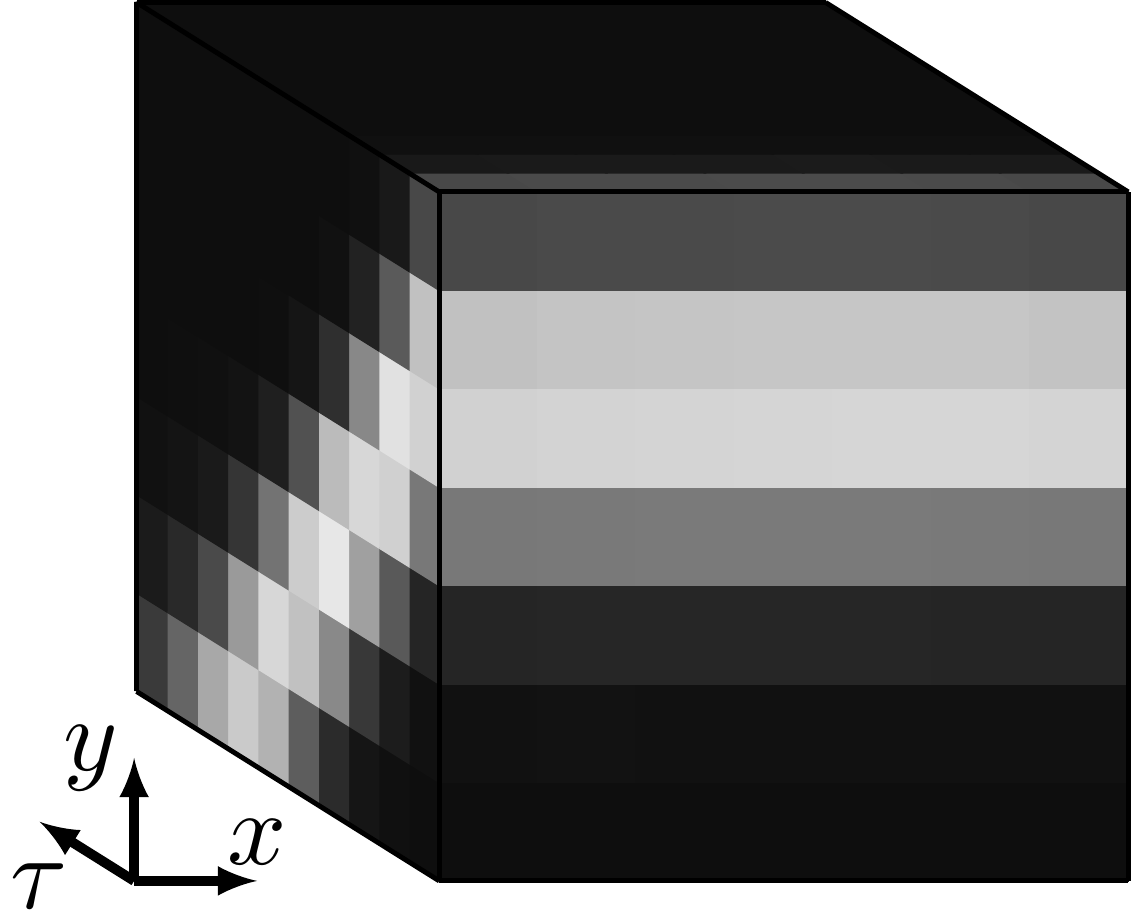}}
	\caption{Three-dimensional illustration of two of the sixteen MS-Conv kernels learned from the checkerboard texture. Synaptic strength is encoded with brightness.}
	\label{figarch:8}
\end{figure}

\begin{figure}[!b]
	\centering
	\includegraphics[width=0.495\textwidth]{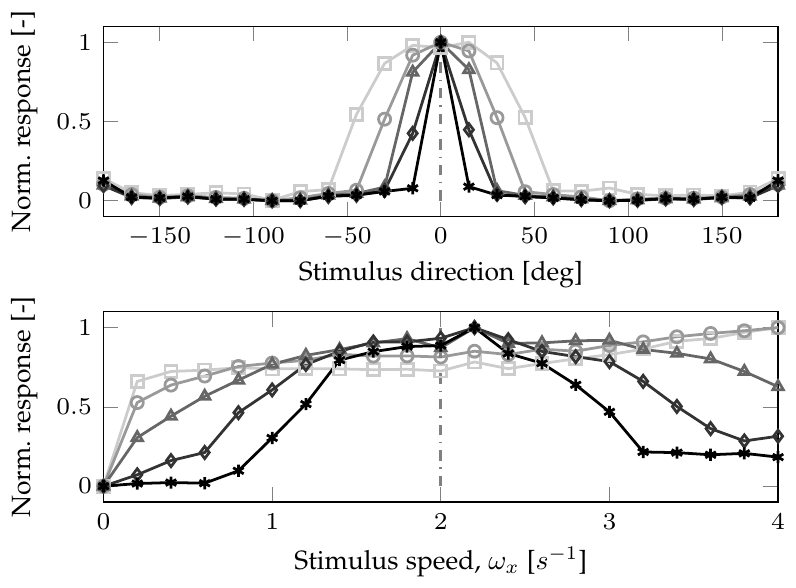}
	\includegraphics[width=0.495\textwidth]{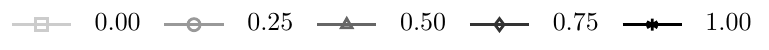}
	\caption{Direction and speed selectivity of neurons in the MS-Conv layer as a function of $\beta$. The dashed lines indicate the training configuration, and each response plot is normalized by its maximum value. Results obtained with the checkerboard texture. $\beta = 0$ means no inhibition, while $\beta = 1$ that inhibitory and excitatory weights contribute equally to the response of this layer.}
	\label{figarch:5}
\end{figure}

\subsection{Spatiotemporal Structure of MS-Conv Kernels}\label{msconv_sup2}

\figref{figarch:8} shows the spatiotemporal appearance of two MS-Conv kernels learned from the checkerboard texture.

\subsection{Velocity Selectivity of MS-Conv Neurons}\label{msconv_sup1}

\figref{figarch:5} shows the velocity selectivity of MS-Conv neurons as a function of $\beta$. These results confirm that, while selectivity to motion direction emerges regardless of the value of 
$\beta$, the inhibitory component of MS-Conv kernels is crucial for the development of speed selectivity. A more extensive sensitivity analysis of these properties can be found in {[77]}.

\begin{figure*}[!t]
	\centering
	\includegraphics[width=\textwidth]{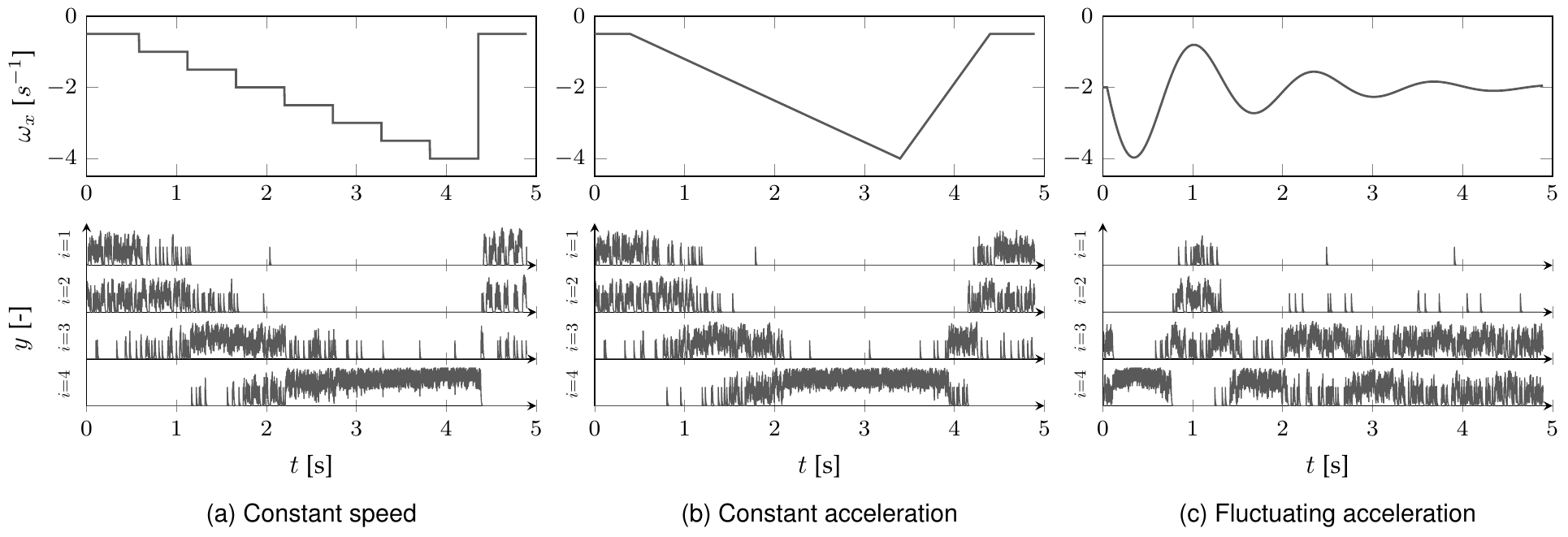}
	\caption{Temporal course of the postsynaptic trace of neurons $i=1$--\hspace{1pt}$4$ from the Dense layer learned from the checkerboard texture (bottom, see Fig. 8), in response to leftward input stimuli with different speed profiles (top). Plots are normalized by the maximum trace on the stimuli evaluated: $0.4098$ by $i = 4$ at $t = 0.36$ s for the fluctuating acceleration case.}
	\label{figarch:10}
\end{figure*}

\subsection{Temporal Response of Dense Neurons}\label{dense1}

\figref{figarch:10} is shown to assess the temporal activity of neurons from the Dense layer learned using the checkerboard texture in response to speed profiles that differ from the constant-speed sequences employed for learning. Due to the pure leftward motion of the stimuli used for this evaluation, only the activity of neurons specialized to this motion direction is shown. Neural activity is measured through the \textit{postsynaptic trace} $y_{i}(t)$ of these cells, which, similarly to Eq. (1), keeps track of the recent history of postsynaptic spikes emitted by a particular neuron, and is given by:

\begin{equation}\label{eqnred:1}
\begin{aligned}
\lambda_{y}\frac{dy_{i}(t)}{dt} = -y_{i}(t) + s_{i}^{l}(t)
\end{aligned}
\end{equation}

\begin{figure}[!b]
	\centering
	\subfloat[Computation of spatial histograms\label{edgeflow1}]{%
		\includegraphics[width=0.45\textwidth]{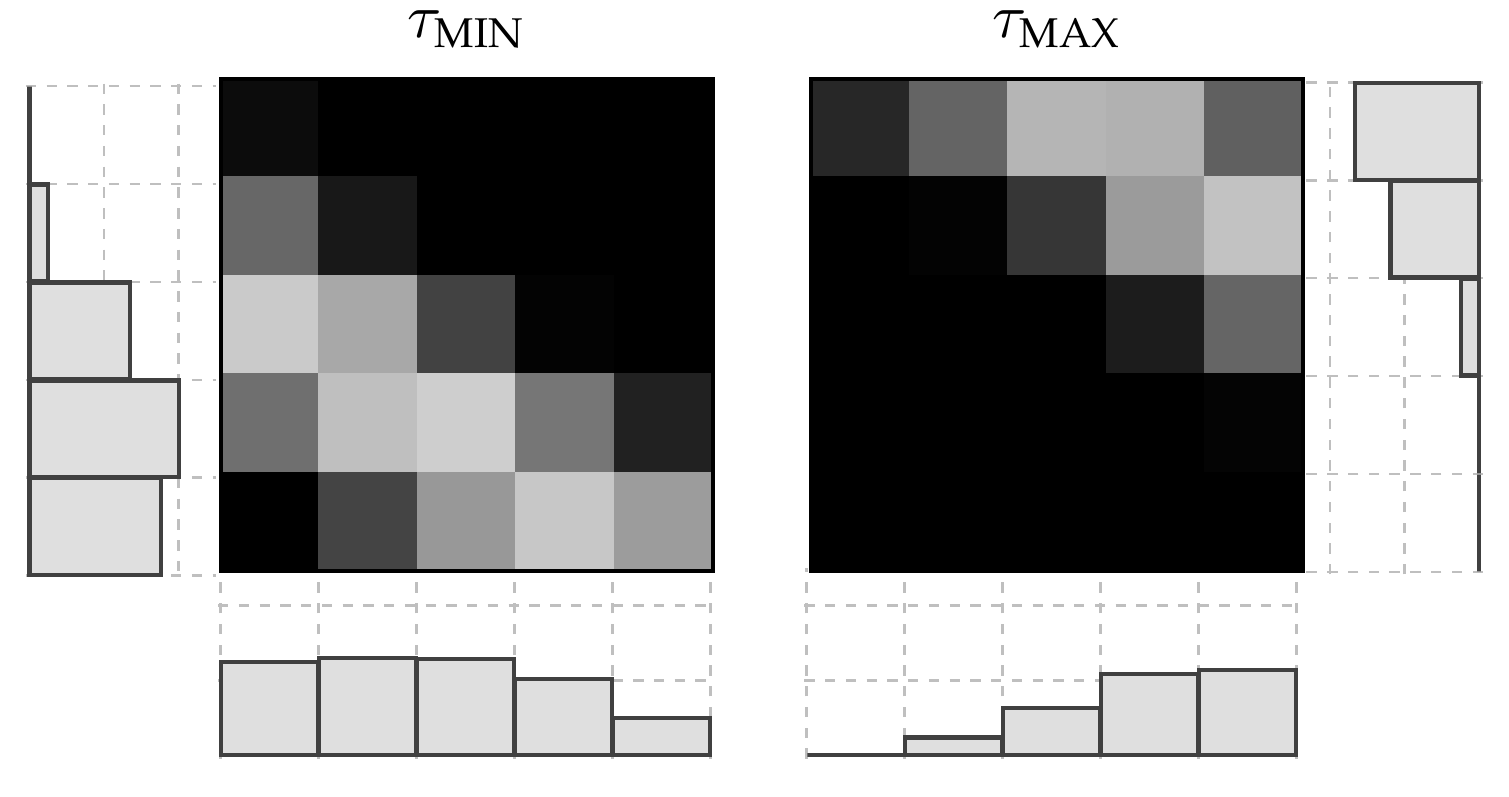}
	}
	\\
	\subfloat[Line fitting of the difference of histograms\label{edgeflow2}]{%
		\includegraphics[width=0.45\textwidth]{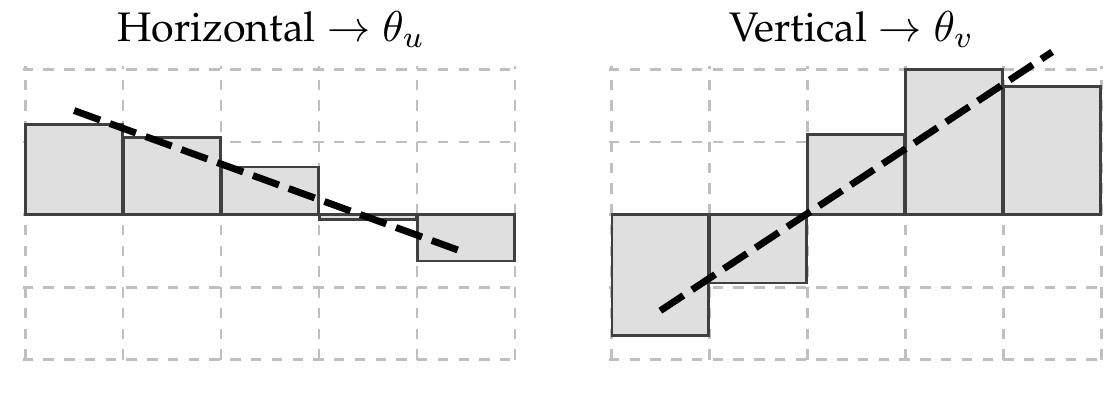}
	}
	\caption{Schematic of the histogram-matching algorithm employed to associate MS-Conv kernels with optical flow vectors based on their spatiotemporal weight distribution.}
	\label{edgeflow}
\end{figure}

Response aspects, such as the overlap of neural activity for some ventral flow ranges, or the dominance of $i=4$ for fast motion, are in line with the results shown in Fig. 8a.

\subsection{From MS-Conv Kernels to Optical Flow Vectors}\label{OF}

To assess the performance of an MS-Conv layer, the optical flow vector identified by each of its spatiotemporal kernels needs to be computed. For this purpose, we employ a variation of the efficient histogram-matching method referred to as EdgeFlow {[85]} on the spatial appearance of these kernels over time (i.e. over the different synaptic delays). It is important to remark that the application of this algorithm does not have any impact on the learning or performance of subsequent layers in the neural architecture. 

The working principle of this approach is illustrated in \figref{edgeflow} for an MS-Conv kernel selective to the local motion of a feature towards the bottom-left corner of the receptive field. Firstly, after convergence, two of the $m$ groups of presynaptic connections of the kernel under analysis are selected. Since optical flow is to be computed from the difference between their spatial weight distributions, the moving feature has to be clearly discernible in both cases. Accordingly, we select the two more distant groups of synapses, denoted by $(\tau_{\text{MIN}},\tau_{\text{MAX}})$, whose cumulative sum of weights $\smash{\Sigma w}$ is greater than the maximum sum $\smash{\Sigma w_{\text{MAX}}}$ scaled by $\gamma\in[0,1]$. Secondly, for each synaptic group, the histogram of each spatial dimension is computed by summing up the weights along the opposite dimension, as shown in \figref{edgeflow1}. Lastly, the resulting weight histograms are subtracted, and the difference for each spatial dimension is fitted to a linear model through least-squares fitting, as shown in \figref{edgeflow2}. The slopes of these lines, denoted by $(\theta_u,\theta_v)$ for the horizontal and vertical dimension respectively, are then used as metrics for the local motion of the identified feature. The sign of these slopes encodes motion direction, while speed is encoded in their absolute value. Consequently, the optical flow components of an MS-Conv kernel are obtained as follows:
\begin{equation}
\begin{aligned}
u = \frac{\theta_u}{\tau_{\text{MAX}}-\tau_{\text{MIN}}}, \quad v = \frac{\theta_v}{\tau_{\text{MAX}}-\tau_{\text{MIN}}}
\end{aligned}
\end{equation}
\noindent{where the inverse of the difference between temporal delays is used as a scaling factor to consider the fact that each kernel uses a specific pair of synaptic groups based on its unique spatiotemporal weight distribution.}

\subsection{Optical Flow Field Color-Coding}\label{OF1}

\begin{figure}[!t]
	\centering
	\includegraphics[width=0.425\textwidth]{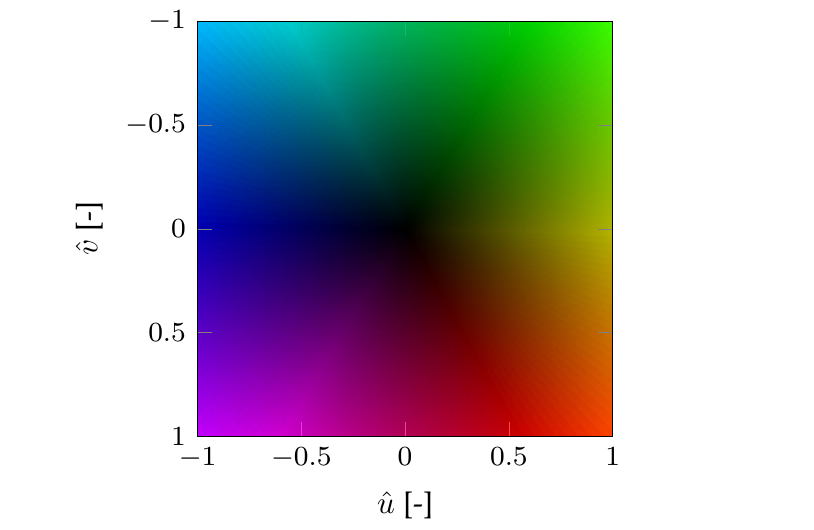}
	\caption{Optical flow field color-coding scheme. Direction is encoded in color hue, and speed in color brightness.}
	\label{color}
\end{figure}

To visualize local motion estimates, we use the color-coding scheme shown in \figref{color}. Direction of motion is encoded in color (i.e. in hue), while speed is encoded in color brightness. Note that, when assigning colors to the kernels of an MS-Conv layer, the corresponding optical flow components (obtained as in Appendix D.5), are normalized to the maximum value of the layer under analysis.

\subsection{Texture Effect in Global Motion Perception}\label{aperture}
\begin{figure}[!b]
	\centering
	\subfloat{%
		\includegraphics[width=0.475\textwidth]{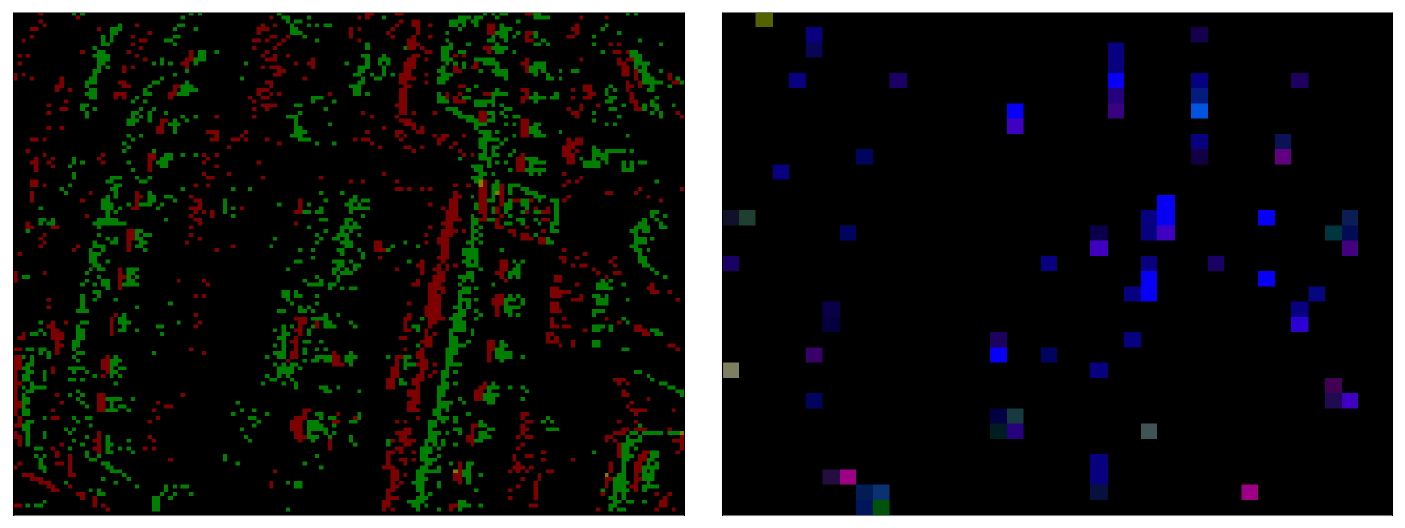}}\\\vspace{-8pt}
	\subfloat{%
		\includegraphics[width=0.475\textwidth]{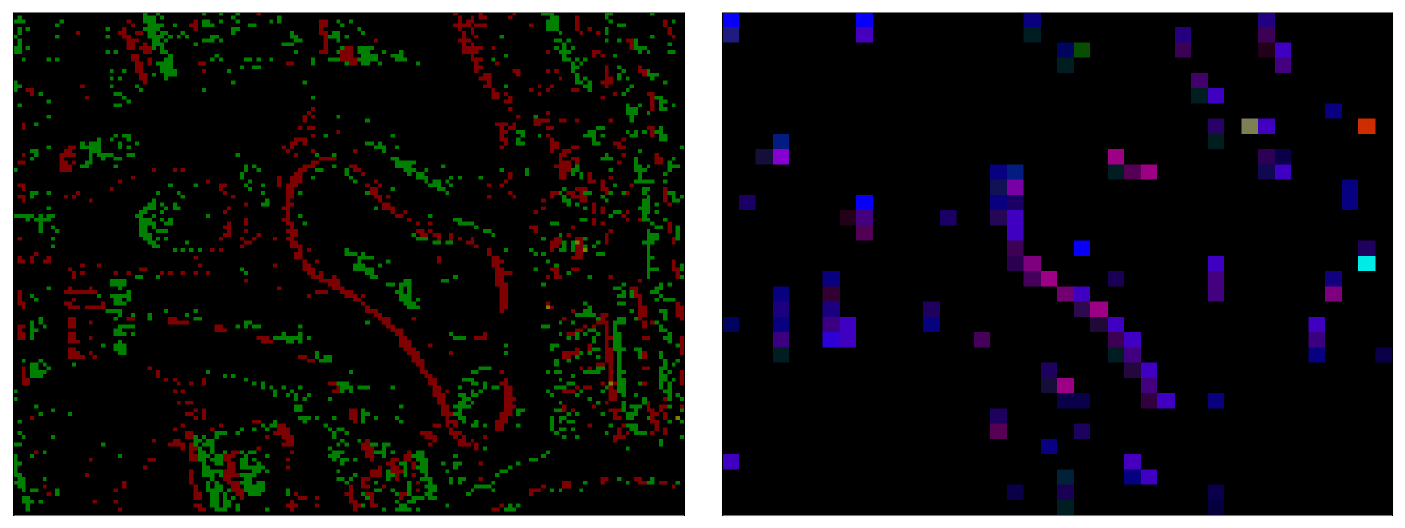}}
	\caption{Input events (left) and local motion perceived by the MS-Conv layer (right) in a sequence from the roadmap texture. Both snapshots correspond to the same global motion according to the planar optical flow formulation from {[55]} (see Appendix \ref{Planar}): $\omega_x\approx -0.4$, $\omega_y\approx 0$, and $D\approx 0$. Specifically, they correspond to $t=0.75$ s (top) and $t=1.25$ s (bottom), from Fig. 13a. MS-Conv color reference shown in Fig. 10, and computed as in Appendices D.5 and D.6}
	\label{apertureture}
\end{figure}

Dense neurons learn from the activity of the Pooling layer, which is a low-dimensional representation of the local motion estimates of the MS-Conv layer. If two activity patterns from the Pooling layer are clearly distinct, Dense neurons will learn them individually, assuming they are equally frequent in the learning process. On the other hand, due to the aperture problem {[71]}, the MS-Conv layer can only represent the optical flow vectors that are normal to the features in sight. Therefore, because of input texture, two activity patterns from the Pooling layer could be sufficiently distinct to have different Dense neurons learning them separately, even though they correspond to the same global motion pattern. This is the case for neurons $\smash{i=\{4,5\}}$ and $\smash{i=\{6,7,8\}}$ from the Dense layer trained on the roadmap texture. According to Fig. 13, these neurons are reactive to pure leftward image motion. However, the former pair of neurons is reactive to the leftward motion of almost pure vertical features (see \figref{apertureture}, top), while the latter group is selective to the same global motion pattern, but perceived through features with different spatial configuration (see \figref{apertureture}, bottom).

\subsection{STDP Evaluation: SS-Conv Kernel Appearance}\label{stdpeval}
\begin{figure}[!h]
	\centering
	\subfloat[Kheradpisheh \textit{et al.} {[25]}]{%
		\includegraphics[width=0.475\textwidth]{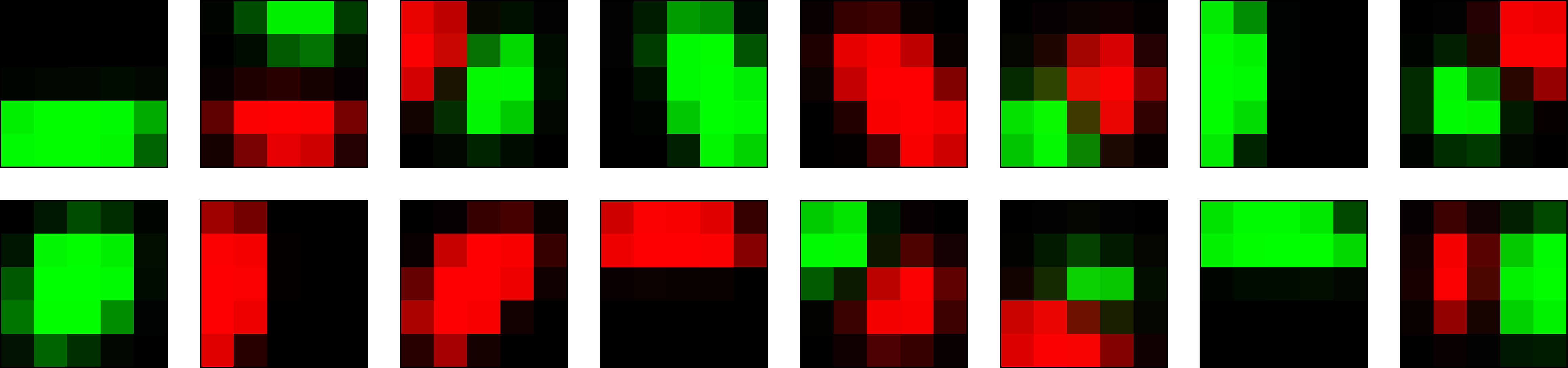}}\\
	\subfloat[Shrestha \textit{et al.} {[27]}]{%
		\includegraphics[width=0.475\textwidth]{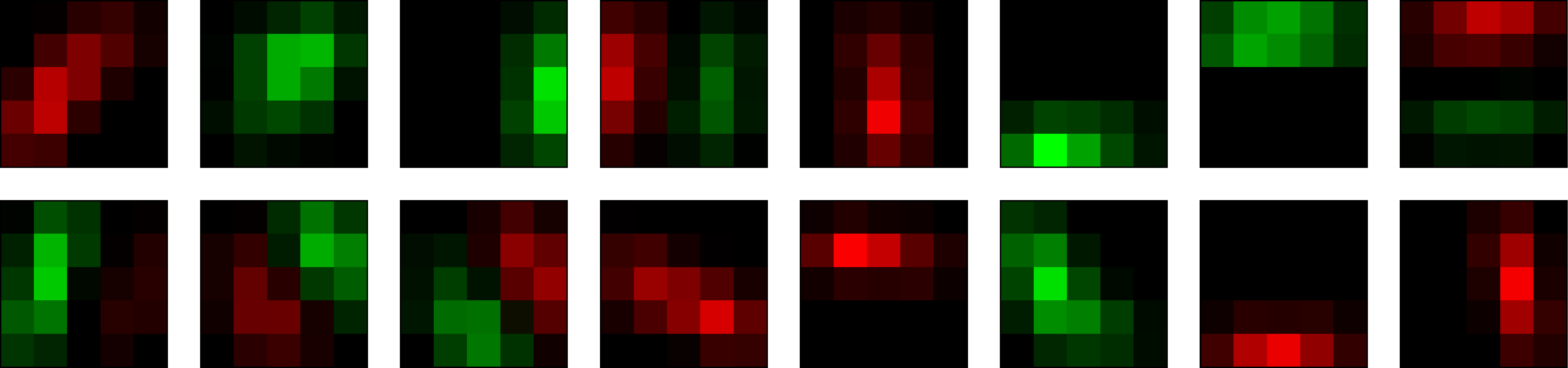}}\\
	\subfloat[Ours]{%
		\includegraphics[width=0.475\textwidth]{09b_ssconv_road.pdf}
	}
	\caption{Appearance of sixteen SS-Conv kernels after the learning process, using Kheradpisheh's {[25]}, Shrestha's {[27]}, and our STDP formulation. Results obtained with the roadmap texture, the same learning rate, and the same budget of training sequences. Synaptic strength is encoded in color brightness.}
	\label{stdpcompappearance}
\end{figure}

\figref{stdpcompappearance} shows the appearance of a set of SS-Conv kernels learned from the roadmap texture using our STDP formulation, and those proposed by Kheradpisheh \textit{et al.} {[25]}, and Shrestha \textit{et al.} {[27]}. Synaptic weights are clipped to the range $W_{i,j,d} \in [0,1]$ for display purposes only.

\subsection{Video}\label{video}

In the supplementary video, we show additional qualitative results of the network performance up to the MS-Conv layer on the real event sequences employed for evaluation in this work. The video can be found at \url{https://goo.gl/MRTeUv}.
\vspace{12.5pt}

\begin{thebibliography}{10}
\providecommand{\url}[1]{#1}
\csname url@samestyle\endcsname
\providecommand{\newblock}{\relax}
\providecommand{\bibinfo}[2]{#2}
\providecommand{\BIBentrySTDinterwordspacing}{\spaceskip=0pt\relax}
\providecommand{\BIBentryALTinterwordstretchfactor}{4}
\providecommand{\BIBentryALTinterwordspacing}{\spaceskip=\fontdimen2\font plus
\BIBentryALTinterwordstretchfactor\fontdimen3\font minus
  \fontdimen4\font\relax}
\providecommand{\BIBforeignlanguage}[2]{{%
\expandafter\ifx\csname l@#1\endcsname\relax
\typeout{** WARNING: IEEEtran.bst: No hyphenation pattern has been}%
\typeout{** loaded for the language `#1'. Using the pattern for}%
\typeout{** the default language instead.}%
\else
\language=\csname l@#1\endcsname
\fi
#2}}
\providecommand{\BIBdecl}{\relax}
\BIBdecl

\bibitem{borst2015common}
A.~Borst and M.~Helmstaedter, ``Common circuit design in fly and mammalian
  motion vision,'' \emph{Nature {N}euroscience}, vol.~18, no.~8, pp.
  1067--1076, 2015.

\bibitem{gibson1950perception}
J.~J. Gibson, \emph{The perception of the visual world}, L.~Carmichael,
  Ed.\hskip 1em plus 0.5em minus 0.4em\relax Houghton Mifflin, 1950.

\bibitem{borst2010fly}
A.~Borst, J.~Haag, and D.~F. Reiff, ``Fly motion vision,'' \emph{Annual
  {R}eview of {N}euroscience}, vol.~33, pp. 49--70, 2010.

\bibitem{srinivasan1996honeybee}
M.~V. Srinivasan, S.~Zhang, M.~Lehrer, and T.~Collett, ``Honeybee navigation en
  route to the goal: {V}isual flight control and odometry,'' \emph{Journal of
  {E}xperimental {B}iology}, vol. 199, no.~1, pp. 237--244, 1996.

\bibitem{de2014autonomous}
C.~De~Wagter, S.~Tijmons, B.~D.~W. Remes, and G.~C. H.~E. {de Croon},
  ``Autonomous flight of a 20-gram flapping wing {MAV} with a 4-gram onboard
  stereo vision system,'' in \emph{Proceedings of the 2014 {IEEE} International
  Conference on Robotics and Automation}, 2014, pp. 4982--4987.

\bibitem{karasek2018tailless}
M.~Kar{\'a}sek, F.~T. Muijres, C.~De~Wagter, B.~D.~W. Remes, and G.~C. H.~E.
  de~Croon, ``A tailless aerial robotic flapper reveals that flies use torque
  coupling in rapid banked turns,'' \emph{Science}, vol. 361, no. 6407, pp.
  1089--1094, 2018.

\bibitem{kirkwood1994hebbian}
A.~Kirkwood and M.~F. Bear, ``Hebbian synapses in visual cortex,''
  \emph{Journal of {N}euroscience}, vol.~14, no.~3, pp. 1634--1645, 1994.

\bibitem{katz1996synaptic}
L.~C. Katz and C.~J. Shatz, ``Synaptic activity and the construction of
  cortical circuits,'' \emph{Science}, vol. 274, no. 5290, pp. 1133--1138,
  1996.

\bibitem{lichtsteiner2008128}
P.~Lichtsteiner, C.~Posch, and T.~Delbruck, ``A 128x128 120 d{B} 15 $\upmu$s
  latency asynchronous temporal contrast vision sensor,'' \emph{IEEE {J}ournal
  of {S}olid-{S}tate {C}ircuits}, vol.~43, no.~2, pp. 566--576, 2008.

\bibitem{posch2011qvga}
C.~Posch, D.~Matolin, and R.~Wohlgenannt, ``A {QVGA} 143 d{B} dynamic range
  frame-free {PWM} image sensor with lossless pixel-level video compression and
  time-domain {CDS},'' \emph{IEEE Journal of Solid-State Circuits}, vol.~46,
  no.~1, pp. 259--275, 2011.

\bibitem{brandli2014240}
C.~Brandli, R.~Berner, M.~Yang, S.~Liu, and T.~Delbruck, ``A 240x180 130 d{B} 3
  $\upmu$s latency global shutter spatiotemporal vision sensor,'' \emph{IEEE
  Journal of Solid-State Circuits}, vol.~49, no.~10, pp. 2333--2341, 2014.

\bibitem{sees1}
\BIBentryALTinterwordspacing
C.~Brandli, R.~Berner, M.~Osswald, and N.~Baumli, ``Silicon eye event sensor
  {SEES1},'' \emph{Insightness AG}, 2018. [Online]. Available:
  \url{https://www.insightness.com}
\BIBentrySTDinterwordspacing

\bibitem{benosman2014event}
R.~Benosman, C.~Clercq, X.~Lagorce, S.~Ieng, and C.~Bartolozzi, ``Event-based
  visual flow,'' \emph{{IEEE} {T}ransactions on {N}eural {N}etworks and
  {L}earning {S}ystems}, vol.~25, no.~2, pp. 407--417, 2014.

\bibitem{fortun2015optical}
D.~Fortun, P.~Bouthemy, and C.~Kervrann, ``Optical flow modeling and
  computation: {A} survey,'' \emph{Computer {V}ision and {I}mage
  {U}nderstanding}, vol. 134, pp. 1--21, 2015.

\bibitem{ilg2016flownet}
E.~Ilg, N.~Mayer, T.~Saikia, M.~Keuper, A.~Dosovitskiy, and T.~Brox, ``Flownet
  2.0: {E}volution of optical flow estimation with deep networks,'' in
  \emph{Proceedings of the 2017 {IEEE} {C}onference on {C}omputer {V}ision and
  {P}attern {R}ecognition}, vol.~2, 2017, pp. 1647--1655.

\bibitem{zhu2018ev}
A.~Z. Zhu, L.~Yuan, K.~Chaney, and K.~Daniilidis, ``{EV}-{F}low{N}et:
  {S}elf-supervised optical flow estimation for event-based cameras,''
  \emph{Proc. Robotics: Science and Systems}, 2018.

\bibitem{ye2018unsupervised}
\BIBentryALTinterwordspacing
C.~Ye, A.~Mitrokhin, C.~Parameshwara, C.~Ferm{\"u}ller, J.~A. Yorke, and
  Y.~Aloimonos, ``Unsupervised learning of dense optical flow and depth from
  sparse event data,'' 2018. [Online]. Available:
  \url{https://arxiv.org/abs/1809.08625}
\BIBentrySTDinterwordspacing

\bibitem{maass1997networks}
W.~Maass, ``Networks of spiking neurons: {T}he third generation of neural
  network models,'' \emph{Neural {N}etworks}, vol.~10, no.~9, pp. 1659--1671,
  1997.

\bibitem{orchard2014bioinspired}
G.~Orchard and R.~{Etienne-Cummings}, ``Bioinspired visual motion estimation,''
  \emph{Proceedings of the {IEEE}}, vol. 102, no.~10, pp. 1520--1536, 2014.

\bibitem{merolla2014million}
P.~A. Merolla, J.~V. Arthur, R.~Alvarez-Icaza, A.~S. Cassidy, J.~Sawada,
  F.~Akopyan, B.~L. Jackson, N.~Imam, C.~Guo, Y.~Nakamura \emph{et~al.}, ``A
  million spiking-neuron integrated circuit with a scalable communication
  network and interface,'' \emph{Science}, vol. 345, no. 6197, pp. 668--673,
  2014.

\bibitem{davies2018loihi}
M.~Davies, N.~Srinivasa, T.-H. Lin, G.~Chinya, Y.~Cao, S.~H. Choday, G.~Dimou,
  P.~Joshi, N.~Imam, S.~Jain \emph{et~al.}, ``Loihi: {A} neuromorphic manycore
  processor with on-chip learning,'' \emph{{IEEE} {M}icro}, vol.~38, no.~1, pp.
  82--99, 2018.

\bibitem{caporale2008spike}
N.~Caporale and Y.~Dan, ``Spike timing--dependent plasticity: {A} {H}ebbian
  learning rule,'' \emph{Annual Review of Neuroscience}, vol.~31, pp. 25--46,
  2008.

\bibitem{masquelier2007unsupervised}
T.~Masquelier and S.~J. Thorpe, ``Unsupervised learning of visual features
  through spike timing dependent plasticity,'' \emph{Public {L}ibrary of
  {S}cience: {C}omputational {B}iology}, vol.~3, no.~2, pp. 247--257, 2007.

\bibitem{diehl2015unsupervised}
P.~U. Diehl and M.~Cook, ``Unsupervised learning of digit recognition using
  spike-timing-dependent plasticity,'' \emph{Frontiers in {C}omputational
  {N}euroscience}, vol.~9, pp. 1--9, 2015.

\bibitem{kheradpisheh2018stdp}
S.~R. Kheradpisheh, M.~Ganjtabesh, S.~J. Thorpe, and T.~Masquelier,
  ``{STDP}-based spiking deep convolutional neural networks for object
  recognition,'' \emph{Neural Networks}, vol.~99, pp. 56--67, 2018.

\bibitem{tavanaei2017multi}
A.~Tavanaei and A.~S. Maida, ``Multi-layer unsupervised learning in a spiking
  convolutional neural network,'' in \emph{Proceedings of the 2017 {IEEE}
  {I}nternational {J}oint {C}onference on {N}eural {N}etworks}, 2017, pp.
  2023--2030.

\bibitem{shrestha2017stable}
A.~Shrestha, K.~Ahmed, Y.~Wang, and Q.~Qiu, ``Stable spike-timing dependent
  plasticity rule for multilayer unsupervised and supervised learning,'' in
  \emph{Proceedings of the 2017 {I}nternational {J}oint {C}onference on
  {N}eural {N}etworks}.\hskip 1em plus 0.5em minus 0.4em\relax IEEE, 2017, pp.
  1999--2006.

\bibitem{stein1965theoretical}
R.~B. Stein, ``A theoretical analysis of neuronal variability,''
  \emph{Biophysical {J}ournal}, vol.~5, no.~2, pp. 173--194, 1965.

\bibitem{hodgkin1952quantitative}
A.~L. Hodgkin and A.~F. Huxley, ``A quantitative description of membrane
  current and its application to conduction and excitation in nerve,''
  \emph{The {J}ournal of {P}hysiology}, vol. 117, no.~4, pp. 25--71, 1952.

\bibitem{izhikevich2003simple}
E.~M. Izhikevich, ``Simple model of spiking neurons,'' \emph{IEEE
  {T}ransactions on {N}eural {N}etworks}, vol.~14, no.~6, pp. 1569--1572, 2003.

\bibitem{kistler1997reduction}
W.~M. Kistler, W.~Gerstner, and J.~L. {van Hemmen}, ``Reduction of the
  {H}odgkin-{H}uxley equations to a single-variable threshold model,''
  \emph{Neural {C}omputation}, vol.~9, no.~5, pp. 1015--1045, 1997.

\bibitem{baudry1998synaptic}
M.~Baudry, ``Synaptic plasticity and learning and memory: 15 years of
  progress,'' \emph{Neurobiology of {L}earning and {M}emory}, vol.~70, no.~1,
  pp. 113--118, 1998.

\bibitem{doya1999computations}
K.~Doya, ``What are the computations of the cerebellum, the basal ganglia and
  the cerebral cortex?'' \emph{Neural {N}etworks}, vol.~12, no. 7-8, pp.
  961--974, 1999.

\bibitem{hebb1952organisation}
D.~O. Hebb, \emph{The organisation of behaviour: {A} neuropsychological
  theory}.\hskip 1em plus 0.5em minus 0.4em\relax Wiley, 1952.

\bibitem{gerstner2002spiking}
W.~Gerstner and W.~M. Kistler, \emph{Spiking neuron models: {S}ingle neurons,
  populations, plasticity}.\hskip 1em plus 0.5em minus 0.4em\relax Cambridge
  {U}niversity {P}ress, 2002.

\bibitem{rumelhart1988learning}
D.~E. Rumelhart, G.~E. Hinton, and R.~J. Williams, ``Learning representations
  by back-propagating errors,'' \emph{Cognitive {M}odeling}, vol.~5, no.~3, pp.
  533--536, 1988.

\bibitem{lee2016training}
J.~H. Lee, T.~Delbruck, and M.~Pfeiffer, ``Training deep spiking neural
  networks using backpropagation,'' \emph{Frontiers in {N}euroscience},
  vol.~10, pp. 1--13, 2016.

\bibitem{wu2018spatio}
Y.~Wu, L.~Deng, G.~Li, J.~Zhu, and L.~Shi, ``Spatio-temporal backpropagation
  for training high-performance spiking neural networks,'' \emph{Frontiers in
  Neuroscience}, vol.~12, pp. 1--12, 2018.

\bibitem{taherkhani2018supervised}
A.~Taherkhani, A.~Belatreche, Y.~Li, and L.~P. Maguire, ``A supervised learning
  algorithm for learning precise timing of multiple spikes in multilayer
  spiking neural networks,'' \emph{{IEEE} {T}ransactions on {N}eural {N}etworks
  and {L}earning {S}ystems}, pp. 1--14, 2018.

\bibitem{shrestha2018slayer}
S.~B. Shrestha and G.~Orchard, ``Slayer: Spike layer error reassignment in
  time,'' in \emph{Advances in Neural Information Processing Systems}, 2018,
  pp. 1417--1426.

\bibitem{perez2013mapping}
J.~A. P{\'e}rez-Carrasco, B.~Zhao, C.~Serrano, B.~Acha,
  T.~{Serrano-Gotarredona}, S.~Chen, and B.~{Linares-Barranco}, ``Mapping from
  frame-driven to frame-free event-driven vision systems by low-rate rate
  coding and coincidence processing--application to feedforward convnets,''
  \emph{IEEE {T}ransactions on {P}attern {A}nalysis and {M}achine
  {I}ntelligence}, vol.~35, no.~11, pp. 2706--2719, 2013.

\bibitem{zambrano2017efficient}
\BIBentryALTinterwordspacing
D.~Zambrano, R.~Nusselder, H.~S. Scholte, and S.~M. Bohte, ``Efficient
  computation in adaptive artificial spiking neural networks,'' 2017. [Online].
  Available: \url{https://arxiv.org/abs/1710.04838}
\BIBentrySTDinterwordspacing

\bibitem{rueckauer2017conversion}
B.~Rueckauer, I.-A. Lungu, Y.~Hu, M.~Pfeiffer, and S.-C. Liu, ``Conversion of
  continuous-valued deep networks to efficient event-driven networks for image
  classification,'' \emph{Frontiers in Neuroscience}, vol.~11, pp. 1--12, 2017.

\bibitem{florian2007reinforcement}
R.~V. Florian, ``Reinforcement learning through modulation of
  spike-timing-dependent synaptic plasticity,'' \emph{Neural {C}omputation},
  vol.~19, no.~6, pp. 1468--1502, 2007.

\bibitem{izhikevich2007solving}
E.~M. Izhikevich, ``Solving the distal reward problem through linkage of {STDP}
  and dopamine signaling,'' \emph{Cerebral {C}ortex}, vol.~17, no.~10, pp.
  2443--2452, 2007.

\bibitem{rombouts2012neurally}
J.~O. Rombouts, P.~R. Roelfsema, and S.~M. Bohte, ``Neurally plausible
  reinforcement learning of working memory tasks,'' in \emph{Advances in
  {N}eural {I}nformation {P}rocessing {S}ystems}, 2012, pp. 1871--1879.

\bibitem{rombouts2012biologically}
J.~O. Rombouts, A.~{van Ooyen}, P.~R. Roelfsema, and S.~M. Bohte,
  ``Biologically plausible multi-dimensional reinforcement learning in neural
  networks,'' in \emph{International {C}onference on {A}rtificial {N}eural
  {N}etworks}, 2012, pp. 443--450.

\bibitem{friedrich2016goal}
J.~Friedrich and M.~Lengyel, ``Goal-directed decision making with spiking
  neurons,'' \emph{Journal of {N}euroscience}, vol.~36, no.~5, pp. 1529--1546,
  2016.

\bibitem{bing2018end}
Z.~Bing, C.~Meschede, K.~Huang, G.~Chen, F.~Rohrbein, M.~Akl, and A.~Knoll,
  ``End-to-end learning of spiking neural network based on {R-STDP} for a lane
  keeping vehicle,'' in \emph{2018 IEEE International Conference on Robotics
  and Automation}.\hskip 1em plus 0.5em minus 0.4em\relax IEEE, 2018, pp. 1--8.

\bibitem{mozafari2018combining}
\BIBentryALTinterwordspacing
M.~Mozafari, M.~Ganjtabesh, A.~{Nowzari-Dalini}, S.~J. Thorpe, and
  T.~Masquelier, ``Combining {STDP} and reward-modulated {STDP} in deep
  convolutional spiking neural networks for digit recognition,'' 2018.
  [Online]. Available: \url{https://arxiv.org/abs/1804.00227}
\BIBentrySTDinterwordspacing

\bibitem{Lucas1981a}
B.~D. Lucas and T.~Kanade, ``An iterative technique of image registration and
  its application to stereo,'' in \emph{Proceedings of the 7th {I}nternational
  {J}oint {C}onference on {A}rtificial {I}ntelligence}, vol.~2, 1981, pp.
  674--679.

\bibitem{benosman2012asynchronous}
R.~Benosman, S.~Ieng, C.~Clercq, C.~Bartolozzi, and M.~Srinivasan,
  ``Asynchronous frameless event-based optical flow,'' \emph{Neural
  {N}etworks}, vol.~27, pp. 32--37, 2012.

\bibitem{brosch2015event}
T.~Brosch, S.~Tschechne, and H.~Neumann, ``On event-based optical flow
  detection,'' \emph{Frontiers in {N}euroscience}, vol.~9, pp. 1--15, 2015.

\bibitem{aung2018event}
M.~T. Aung, R.~Teo, and G.~Orchard, ``Event-based plane-fitting optical flow
  for dynamic vision sensors in {FPGA},'' in \emph{Proceedings of the 2018 IEEE
  International Symposium on Circuits and Systems}.\hskip 1em plus 0.5em minus
  0.4em\relax IEEE, 2018, pp. 1--5.

\bibitem{hordijk2017vertical}
B.~J.~P. Hordijk, K.~Y.~W. Scheper, and G.~C. H.~E. {de Croon}, ``Vertical
  landing for micro air vehicles using event-based optical flow,''
  \emph{Journal of Field Robotics}, vol.~35, no.~1, pp. 69--90, 2018.

\bibitem{tschechne2014bio}
S.~Tschechne, R.~Sailer, and H.~Neumann, ``Bio-inspired optic flow from
  event-based neuromorphic sensor input.'' in \emph{Proceedings of the 6th
  {IAPR} {W}orkshop on {A}rtificial {N}eural {N}etworks in {P}attern
  {R}ecogniftion}.\hskip 1em plus 0.5em minus 0.4em\relax Springer, 2014, pp.
  171--182.

\bibitem{barranco2015bio}
F.~Barranco, C.~Fermuller, and Y.~Aloimonos, ``Bio-inspired motion estimation
  with event-driven sensors,'' in \emph{International Work-Conference on
  Artificial Neural Networks}.\hskip 1em plus 0.5em minus 0.4em\relax Springer,
  2015, pp. 309--321.

\bibitem{brosch2016event}
T.~Brosch and H.~Neumann, ``Event-based optical flow on neuromorphic
  hardware,'' in \emph{Proceedings of the 9th {EAI} {I}nternational
  {C}onference on {B}io-inspired {I}nformation and {C}ommunications
  {T}echnologies}, 2016, pp. 551--558.

\bibitem{zhu2017event}
A.~Z. Zhu, N.~Atanasov, and K.~Daniilidis, ``Event-based feature tracking with
  probabilistic data association,'' in \emph{Proceedings of the 2017 IEEE
  International Conference on Robotics and Automation}.\hskip 1em plus 0.5em
  minus 0.4em\relax IEEE, 2017, pp. 4465--4470.

\bibitem{gallego2017accurate}
G.~Gallego and D.~Scaramuzza, ``Accurate angular velocity estimation with an
  event camera,'' \emph{IEEE Robotics and Automation Letters}, vol.~2, no.~2,
  pp. 632--639, 2017.

\bibitem{mitrokhin2018event}
\BIBentryALTinterwordspacing
A.~Mitrokhin, C.~Fermuller, C.~Parameshwara, and Y.~Aloimonos, ``Event-based
  moving object detection and tracking,'' 2018. [Online]. Available:
  \url{https://arxiv.org/abs/1803.04523}
\BIBentrySTDinterwordspacing

\bibitem{gallego2018unifying}
G.~Gallego, H.~Rebecq, and D.~Scaramuzza, ``A unifying contrast maximization
  framework for event cameras, with applications to motion, depth, and optical
  flow estimation,'' in \emph{Proceedings of the 2018 IEEE International
  Conference on Computer Vision and Pattern Recognition}, vol.~1, 2018.

\bibitem{liu2018abmof}
\BIBentryALTinterwordspacing
M.~Liu and T.~Delbruck, ``{ABMOF}: {A} novel optical flow algorithm for dynamic
  vision sensors,'' 2018. [Online]. Available:
  \url{https://arxiv.org/abs/1805.03988}
\BIBentrySTDinterwordspacing

\bibitem{lagorce2015spatiotemporal}
X.~Lagorce, S.-H. Ieng, X.~Clady, M.~Pfeiffer, and R.~B. Benosman,
  ``Spatiotemporal features for asynchronous event-based data,''
  \emph{Frontiers in Neuroscience}, vol.~9, pp. 1--13, 2015.

\bibitem{giulioni2016event}
M.~Giulioni, X.~Lagorce, F.~Galluppi, and R.~B. Benosman, ``Event-based
  computation of motion flow on a neuromorphic analog neural platform,''
  \emph{Frontiers in {N}euroscience}, vol.~10, pp. 1--13, 2016.

\bibitem{haessig2018spiking}
G.~Haessig, A.~Cassidy, R.~Alvarez, R.~Benosman, and G.~Orchard, ``Spiking
  optical flow for event-based sensors using {IBM}'s {T}rue{N}orth
  neurosynaptic system,'' \emph{IEEE Transactions on Biomedical Circuits and
  Systems}, vol.~12, no.~4, pp. 860--870, 2018.

\bibitem{richter2014bio}
C.~Richter, F.~R{\"o}hrbein, and J.~Conradt, ``Bio-inspired optic flow
  detection using neuromorphic hardware,'' \emph{Bernstein {C}onference on
  {C}omputational {N}euroscience}, poster, 2014.

\bibitem{orchard2013spiking}
G.~Orchard, R.~Benosman, R.~Etienne-Cummings, and N.~V. Thakor, ``A spiking
  neural network architecture for visual motion estimation,'' in
  \emph{Proceedings of the 2013 {IEEE} {B}iomedical {C}ircuits and {S}ystems
  {C}onference}, 2013, pp. 298--301.

\bibitem{reichardt1961autocorrelation}
W.~Reichardt, ``Autocorrelation, a principle for the evaluation of sensory
  information by the central nervous system,'' \emph{Sensory {C}ommunication},
  pp. 303--317, 1961.

\bibitem{adelson1985spatiotemporal}
E.~H. Adelson and J.~R. Bergen, ``Spatiotemporal energy models for the
  perception of motion,'' \emph{Journal of the {O}ptical {S}ociety of
  {A}merica}, vol.~2, no.~2, pp. 284--299, 1985.

\bibitem{ullman1979interpretation}
S.~Ullman, \emph{The interpretation of visual motion.}\hskip 1em plus 0.5em
  minus 0.4em\relax MIT Press, 1979.

\bibitem{shon2004motion}
A.~P. Shon, R.~P. Rao, and T.~J. Sejnowski, ``Motion detection and prediction
  through spike-timing dependent plasticity,'' \emph{Network: {C}omputation in
  Neural Systems}, vol.~15, no.~3, pp. 179--198, 2004.

\bibitem{wenisch2005spontaneously}
O.~G. Wenisch, J.~Noll, and J.~L. {Van Hemmen}, ``Spontaneously emerging
  direction selectivity maps in visual cortex through {STDP},''
  \emph{Biological {C}ybernetics}, vol.~93, no.~4, pp. 239--247, 2005.

\bibitem{adams2015computational}
S.~V. Adams and C.~M. Harris, ``A computational model of innate directional
  selectivity refined by visual experience,'' \emph{Scientific {R}eports},
  vol.~5, pp. 1--13, 2015.

\bibitem{morrison2007spike}
A.~Morrison, A.~Aertsen, and M.~Diesmann, ``Spike-timing-dependent plasticity
  in balanced random networks,'' \emph{Neural {C}omputation}, vol.~19, no.~6,
  pp. 1437--1467, 2007.

\bibitem{abercrombie2017dictionary}
M.~Abercrombie, C.~J. Hickman, and M.~L. Johnson, \emph{A dictionary of
  biology}.\hskip 1em plus 0.5em minus 0.4em\relax Routledge, 2017.

\bibitem{neuroFede}
F.~Paredes-Vall\'es, \emph{Neuromorphic Computing of Event-Based Data for
  High-Speed Vision-Based Navigation}.\hskip 1em plus 0.5em minus 0.4em\relax
  \hspace{-5pt}M.Sc. Thesis, Faculty of Aerospace Engineering, Delft University
  of Technology, 2018. [Online]. Available:
  \href{https://repository.tudelft.nl/islandora/object/uuid%3Aaa13959b-79b9-4dfc-b5e0-7c501d9d3e2f}{TU
  Delft Education Repository}.

\bibitem{bohte2012efficient}
S.~M. Bohte, ``Efficient spike-coding with multiplicative adaptation in a spike
  response model,'' in \emph{Advances in {N}eural {I}nformation {P}rocessing
  {S}ystems}, 2012, pp. 1835--1843.

\bibitem{van2000stable}
M.~C.~W. Van~Rossum, G.~Q. Bi, and G.~G. Turrigiano, ``Stable {H}ebbian
  learning from spike timing-dependent plasticity,'' \emph{Journal of
  {N}euroscience}, vol.~20, no.~23, pp. 8812--8821, 2000.

\bibitem{thorpe1990spike}
S.~J. Thorpe, ``Spike arrival times: {A} highly efficient coding scheme for
  neural networks,'' \emph{Parallel {P}rocessing in {N}eural {S}ystems}, pp.
  91--94, 1990.

\bibitem{barlow1965mechanism}
H.~B. Barlow and W.~R. Levick, ``The mechanism of directionally selective units
  in rabbit's retina.'' \emph{The {J}ournal of {P}hysiology}, vol. 178, no.~3,
  pp. 477--504, 1965.

\bibitem{mueggler2017event}
E.~Mueggler, H.~Rebecq, G.~Gallego, T.~Delbruck, and D.~Scaramuzza, ``The
  event-camera dataset and simulator: {E}vent-based data for pose estimation,
  visual odometry, and {SLAM},'' \emph{The {I}nternational {J}ournal of
  {R}obotics {R}esearch}, vol.~36, no.~2, pp. 142--149, 2017.

\bibitem{rueckauer2016evaluation}
B.~Rueckauer and T.~Delbruck, ``Evaluation of event-based algorithms for
  optical flow with ground-truth from inertial measurement sensor,''
  \emph{Frontiers in Neuroscience}, vol.~10, pp. 1--17, 2016.

\bibitem{Longuet-Higgins1980}
H.~C. Longuet-Higgins and K.~Prazdny, ``The interpretation of a moving retinal
  image,'' \emph{Proceedings of the Royal Society of London B: Biological
  Sciences}, vol. 208, no. 1173, pp. 385--397, 1980.

\bibitem{mcguire2016local}
K.~McGuire, G.~C. H.~E. {de Croon}, C.~De~Wagter, B.~D.~W. Remes, K.~Tuyls, and
  H.~Kappen, ``Local histogram matching for efficient optical flow computation
  applied to velocity estimation on pocket drones,'' in \emph{2016 {IEEE}
  {I}nternational {C}onference on {R}obotics and {A}utomation}.\hskip 1em plus
  0.5em minus 0.4em\relax IEEE, 2016, pp. 3255--3260.

\end{thebibliography}
\end{document}